%% file: 00_main.tex
\documentclass[twocolumn]{svjour3}          
\smartqed  
\usepackage{graphicx}
\usepackage{boldline,multirow}
\usepackage{xcolor}
\usepackage{comment}
\usepackage{amsmath}
\usepackage{amssymb}
\usepackage{subcaption}
\newcommand{\norm}[1]{\left\lVert#1\right\rVert}
\DeclareMathOperator*{\argmax}{argmax}
\begin{document}

\title{Compositional Convolutional Neural Networks: \\
A Robust and Interpretable Model for Object Recognition under Occlusion}
\titlerunning{Compositional Convolutional Neural Networks}        

\author{Adam Kortylewski \and
        Qing Liu \and
        Angtian Wang \and
        Yihong Sun \and\\
        Alan Yuille
}


\institute{Adam Kortylewski, Qing Liu, Angtian Wang,
        Yihong Sun, 
        Alan Yuille \at
           Johns Hopkins University, Baltimore, MD, USA \\
          \email{\{akortyl1,qingliu,angtianwang,ysun86,ayuille1\}@jhu.edu}           
}

\date{Received: date / Accepted: date}

\maketitle
\input{00_abstract}
\input{01_introduction}
\input{02_related_work}
\input{03_0_model_classification}

\input{03_1_model_detection}

\input{03_2_model_inference}

\input{04_0_exp_class}
\input{04_1_exp_detect}
\input{04_2_exp_interpret}

\input{05_conclusion}


\bibliographystyle{spmpsci}      
\bibliography{egbib}   

\end{document}

%% file: 00_abstract.tex
\begin{abstract}
Computer vision systems in real-world applications need to be robust to partial occlusion while also being explainable. 
In this work, we show that black-box deep convolutional neural networks (DCNNs) have only limited robustness to partial occlusion.
We overcome these limitations by unifying DCNNs with part-based models into Compositional Convolutional Neural Networks (CompositionalNets) - an interpretable deep architecture with innate robustness to partial occlusion.
Specifically, we propose to replace the fully connected classification head of DCNNs with a differentiable compositional model that can be trained end-to-end. The structure of the compositional model enables CompositionalNets to decompose images into objects and context, as well as to further decompose object representations in terms of individual parts and the objects' pose.
The generative nature of our compositional model enables it to localize occluders and to recognize objects based on their non-occluded parts. 
We conduct extensive experiments in terms of image classification and object detection on images of artificially occluded objects from the PASCAL3D+ and ImageNet dataset, and real images of partially occluded vehicles from the MS-COCO dataset.
Our experiments show that CompositionalNets made from several popular DCNN backbones (VGG-16, ResNet50, ResNext) improve by a large margin over their non-compositional counterparts at classifying and detecting partially occluded objects.
Furthermore, they can localize occluders accurately despite being trained with class-level supervision only. 
Finally, we demonstrate that CompositionalNets provide human interpretable predictions as their individual components can be understood as detecting parts and estimating an objects' viewpoint. 

\keywords{Compositional models \and Robustness to partial occlusion \and Image classification \and Object detection \and Out-of-distribution generalization}
\end{abstract}

%% file: 01_introduction.tex
\begin{figure*}
    \begin{subfigure}{.6\linewidth}
        \centering
        \includegraphics[width=0.8\linewidth]{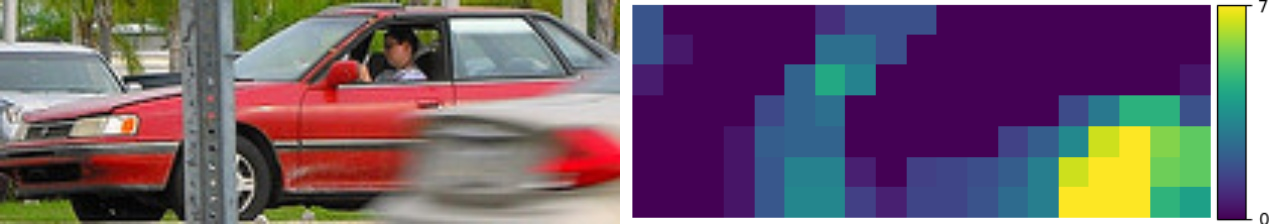}
        \includegraphics[width=0.8\linewidth]{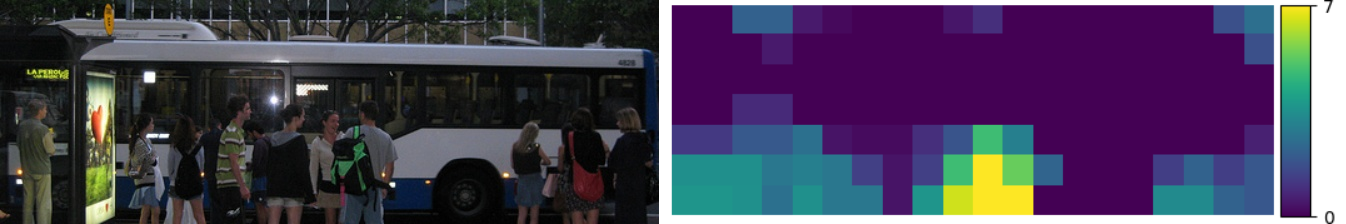}
        \includegraphics[width=0.8\linewidth]{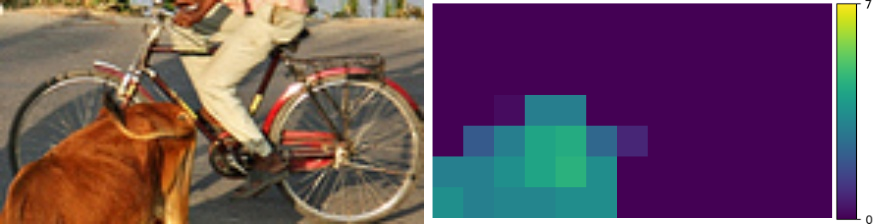}
        \caption{}
    \end{subfigure}
    \begin{subfigure}{.35\linewidth}
        \centering
        \includegraphics[height=5.1cm]{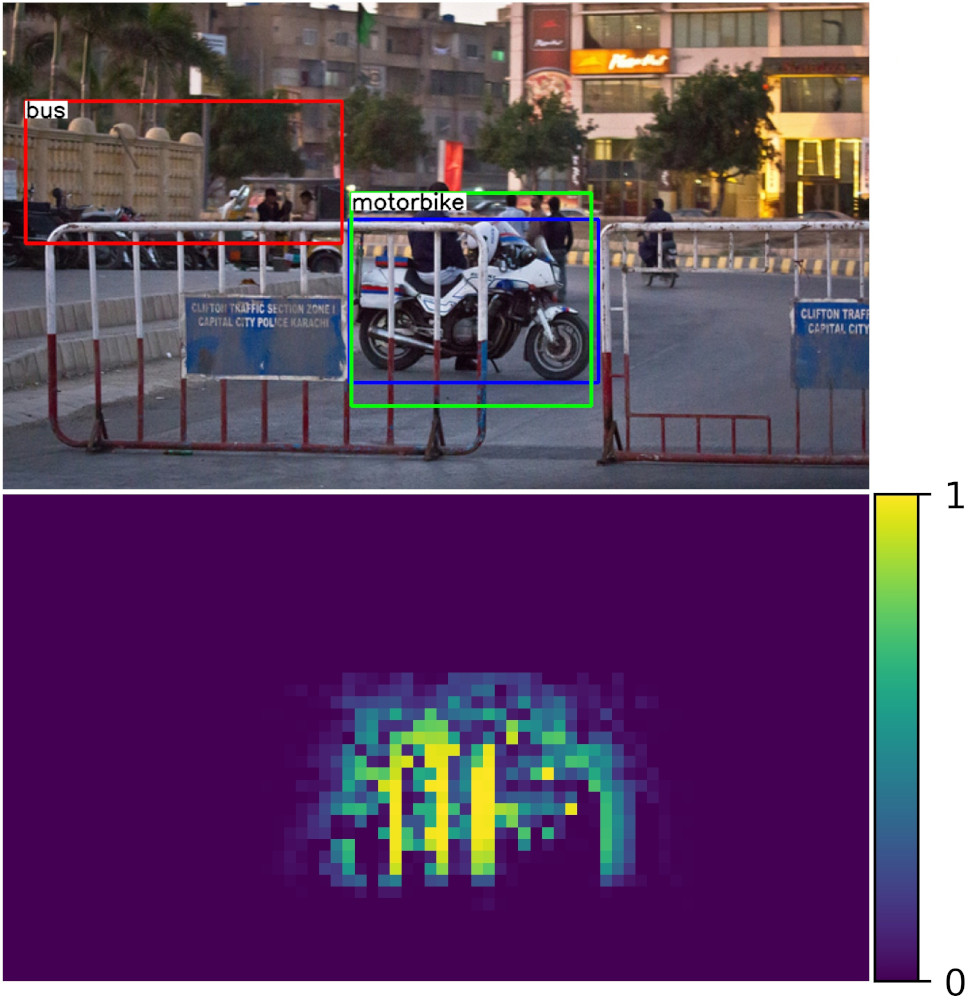}
        \caption{}
    \end{subfigure}    
    \caption{Example of images for the classification (a) and detection (b) of partially occluded objects from the MS-COCO dataset \cite{lin2014microsoft}. 
    A standard DCNN misclassifies the images in (a) and does not detect the motorbike in (b), while also having a false-positive detection of a bus in the background. In contrast, CompositionalNets provide correct predictions in all cases. 
    Intuitively, a CompositionalNet can localize the occluders (see visualization of occlusion scores) and subsequently focus on the non-occluded parts of the object to make a robust prediction.    
    }
    \label{fig:intro}
\end{figure*}

\section{Introduction}
Advances in the architecture design of deep convolutional neural networks (DCNNs) \cite{krizhevsky2012imagenet,simonyan2014very,he2016deep} increased the performance of computer vision systems at object recognition enormously.
This led to the deployment of computer vision models in safety-critical real-world applications, such as self-driving cars and security systems. 
In these application areas, we expect models to reliably generalize to previously unseen visual stimuli.
However, in practice we observe that deep models do not generalize as well as humans in scenarios that are different from what has been observed during training, e.g., unseen partial occlusion, rare object pose, changes in the environment, etc..
This lack of generalization may lead to fatal consequences in real-world applications, e.g. when driver-assistant systems fail to detect partially occluded pedestrians \cite{economist2017uber}.

In particular, a key problem for computer vision systems is how to deal with partial occlusion. 
In natural environments, objects are often surrounded and partially occluded by each other. 
The large variability of occluders in terms of their shape, appearance and position introduces an exponential complexity in the data distribution \cite{yuille2018deep} that is unfeasible to be exhaustively represented in finite training data.
Recent works \cite{hongru,kortylewski2019compositional} have shown that deep vision systems are not as robust as humans at recognizing partially occluded objects.
Moreover, our experiments show that this limitation persists even when deep networks have been exposed to large amounts of partial occlusion during training. 
Hence, this reveals a fundamental limitation of current approaches to computer vision that needs to be addressed.

While robustness to partial occlusion is crucial, safety-critical applications also require AI systems to provide human interpretable explanations of their prediction. Such explanations can help to understand failures and enable the further advancement of the performance of the models, while potentially also supporting the scientific understanding of the vision process. 
This insight motivated recent work to focus on developing interpretable vision models \cite{ross2017right,hu2016harnessing,stone2017teaching,zhang2018interpretable,zhang2018visual}. 
However, most often interpretable models do not perform as well as black-box DCNNs and can only be applied in a very specific domain.

In this work, we propose a general deep architecture that recognizes partially occluded objects robustly even when it has not been exposed to partial occlusion during training, while also being able to provide human interpretable explanations of its prediction. 
Here, we refer to interpretability in terms of the definition provided by Montavon et al. \cite{montavon2018methods} as \textit{the mapping of an abstract concept (e.g. a predicted class) into a domain that the human can make sense of} (e.g. in the image space instead of an abstract neural feature space) and an explanation as \textit{a collection of features from the interpretable domain that have contributed to the decision} (e.g. object part detections and occluder locations in the image space).

Our key contribution is that we \textit{unify} compositional models and DCNNs into an architecture that we term \textit{Compositional Convolutional Neural Network}.
Our model is inspired by a number of works that demonstrated how the modularity of compositional representations enables efficient learning \cite{jin2006context}, interpretability \cite{george2017generative} and strong generalization at classifying partially occluded 2D patterns \cite{george2017generative,kortylewski2017model,wang2017detecting,zhang2018deepvoting} and 3D objects \cite{kortylewski2019compositional}.
In particular, we propose to replace the fully-connected classification head of a DCNN with a differentiable compositional model that can be trained end-to-end.
The compositional model represents objects as a set of parts that are composed spatially. This enables a robust classification based on the spatial configuration of a few visible parts. 
The compositional model is regularized to be fully generative in terms of the neural feature activations of the last convolutional layer.
The generative nature of the model enables the network to localize occluders in an image and subsequently focus on the non-occluded parts of the object for recognition. 
In addition, the structure of our compositional model enforces the decomposition of the image representation as a mixture of the context and object representation. 
This context-aware image representation enables us to control the influence of the context on the models' prediction, which turns out to be important for the detection of partially occluded objects.
Figure \ref{fig:intro} illustrates the robustness of CompositionalNets at classifying and detecting partially occluded objects, while also being able to localize occluders in an image.
In particular, it shows several images of partially occluded objects from the MS-COCO dataset \cite{lin2014microsoft}.
Next to these images, we show occlusion scores that illustrate the position of occluders 
as estimated by the CompositionalNets.
Note how the occluders are accurately localized and provide a human interpretable explanation of the models' perception of the image.

Our work on CompositionalNets includes several important contributions:
\begin{enumerate}
	\item We propose \textbf{a differentiable compositional model} that can be trained end-to-end and that is regularized to be generative in terms of the feature activations of a DCNN.
	This enables us to integrate compositional models with popular deep network architectures into \textbf{compositional convolutional neural networks}, a unified deep model with innate robustness to partial occlusion.
	\item We evaluate the robustness to partial occlusion on images of artificially occluded objects from the PASCAL3D+ and ImageNet datasets, as well as real images of partially occluded vehicles from the MS-COCO dataset.
	Our extensive experiments with popular DCNN backbones (VGG-16 \cite{simonyan2014very}, ResNet50 \cite{he2016deep}, ResNext \cite{xie2017aggregated}) demonstrate that \textbf{CompositionalNets consistently outperform their non-com- positional counterparts by a large margin} at the classification and detection of partially occluded objects.
    \item We propose to decompose the image representation in CompositionalNets as a mixture model of context and object representations. We demonstrate that such \textbf{context-aware CompositionalNets} allow for fine-grained control of the influence of the object context on the model prediction, which increases the robustness when detecting strongly occluded objects.	
	\item We show that \textbf{CompositionalNets are human interpretable}, because their predictions can be understood in terms of object part detection, occluder localization and object viewpoint estimation. We perform qualitative and quantitative experiments that demonstrate the ability of CompositionalNets to localize occluders accurately, despite being trained with class labels only.
\end{enumerate}

Finally, we note that this article extends the conference papers \cite{kortylewski2020compositional,wang2020compositional} in multiple ways: (1) We present the models of both papers coherently in a common theoretical framework and perform a number additional experiments, including ablation studies. (2) We show that a bad generalization to out-of-distribution examples in terms of partial occlusion is not just a limitation of the VGG-16 network. Instead, our experiments show that it is a general challenge for a variety of advanced deep architectures. (3) We find that CompositionalNets learned from residual backbones can use two fundamentally different approaches to achieve robustness to partial occlusion: Invariance to occlusion and localization of occluders. We observe that the combination of both approaches enables these models to achieve the highest robustness. (4) We study the interpretability of CompositionalNets quantitatively and show that the predictions of CompositionalNets are highly interpretable in terms of part detection, occluder localization and pose estimation. (5) We study large-scale classification of non-vehicle objects with CompositionalNets and achieve very promising results. 

In summary, this article shows that the recognition of partially occluded objects poses a general and fundamental challenge to deep models. We give important insights into how this limitation can be overcome by unifying deep models with compositional models, and we show that the resulting CompositionalNets are not just more robust but also much more interpretable compared to their non-compositional counterparts.

%% file: 02_related_work.tex
\section{Related Work}
\subsection{Object Recognition Under Occlusion}
Many recent studies \cite{hongru,kortylewski2019compositional} have shown that, current deep neural networks are much less robust to partial occlusion compared with humans at object classification. Fawzi et al. \cite{fawzi2016measuring} show that DCNNs are vulnerable to partial occlusions simulated by masking small patches of the input image. Several following works \cite{devries2017improved,yun2019cutmix} propose to augment the training data by masking out patches from the image. Our experimental results in Section \ref{sec:exp_class} show that such data augmentation approaches only have limited beneficial effects. 
Moreover, data augmentation increases the amount of training data and thus the training time and cost. Therefore, it is desirable to develop novel neural network architectures that are inherently robust to partial occlusion.
Xiao et al. \cite{xiao2019tdapnet} propose TDAPNet, a deep network with an attention mechanism that masks out occluded features in lower layers to increase the robustness of the model against occlusion. Though it can work reliably on artificial occlusion, our results show that this model does not perform well on images with real occlusion.

Compared to image classification, object detection additionally involves the estimation of the object location and bounding box. While a search over the image can be implemented efficiently, e.g. using a scanning window \cite{lampert2008beyond}, the number of potential bounding boxes is combinatorial in the number of pixels.
Currently, the most widely used approach for solving this problem is to use region proposal networks (RPNs) \cite{girshick2014rich} which enable the training of fast approaches for object detection 
\cite{girshick2015fast,ren2015faster,cai2018cascade}. 
However, our experiments in Section \ref{sec:exp_detect} demonstrate that RPNs do not estimate the location and bounding box of an object correctly under occlusion, which consequentially deteriorates the performance of these approaches.

To resolve this problem, a boosted cascade-based method for detecting partially visible objects has been proposed by \cite{InferenOccDetect}. However, this approach is based on hand-crafted features and can only be applied to images which are artificially occluded by cutting out patches.
A number of deep learning based approaches have also been proposed for detecting occluded objects \cite{OR-CNN,OccNet}, but they require detailed part level annotation to reconstruct the occluded objects.
The work of \cite{3DAspectlets} proposes to use 3D models and treat occlusion as a multi-label classification problem. However, in real-world scenarios, the classes of the occluders can be difficult to model and are often not known a-priori. 
Other approaches focus on videos or stereo images \cite{SymmNet,Symmetricstereo}. In this work, we consider the problem of partial occlusion in still images.

In  contrast  to  deep  learning  approaches, generative compositional models \cite{jin2006context,zhu2008,fidler2014,dai2014unsupervised,kortylewski2017greedy} have been shown to be inherently robust to partial occlusion. Such models have been successfully applied for detecting object parts \cite{wang2017detecting,zhang2018deepvoting} and recognizing 2D patterns \cite{george2017generative,kortylewski2016probabilistic} under partial occlusion. Such part-based voting approaches \cite{wang2017detecting,zhang2018deepvoting} perform reliably for semantic part detection under occlusion, but they assume a fixed size bounding box and are viewpoint specific, which limits their applicability in the context of object detection.

\subsection{Relation of CompositionalNets to And-Or Graphs}
And-Or graphs (AOG) \cite{nilsson1980principles,jin2006context,dechter2007and} have been investigated e.g. to build hierarchical part-based models for human parsing and for object detection. 
Intuitively, the or-nodes allow the model to learn different object/part configurations, while the and-node decomposes them into smaller components. 
Early works in this direction \cite{chen2008rapid,zhu2008max,li2013vehicle} relied on pre-defined parts and pre-defined graph structures. 
To learn the AOG model with less supervision, Zhu et al. \cite{zhu2008} use recursive compositional clustering. However, this method may lead to unexplainable parts and structures. 
Song et al. \cite{song2013discriminatively} used an over-complete set of shape primitives to quantize the image lattices and then organized them into an AOG by exploiting their compositional relations through iterative cutting. 
Xia et al. \cite{xia2016pose} explicitly defined parts and part compositions, which correspond to the leaf node and non-leaf node in the AOG respectively. Then they used a score function with pre-defined adjacent part pairs to learn the structure of the AOG, which still required considerable amount of human input. 
Wu et al. \cite{wu2015learning} made use of a large number of synthetic images generated by CAD simulations, on which 17 semantic parts were manually labeled. They enumerated all configurations observed from the synthetic data and then used a graph compression algorithm to get the refined AOG structure. All these models learned the AOG in two steps: after the graph structure was decided, variants of latent structural SVM was used to learn model parameters. In a different approach, Lin et al. \cite{lin2014discriminatively} learned AOG by joint optimization of both model structures and parameters. The resulting model worked on object shape detection with moderate performance and may not be easily applied to general object detection.\\

Most of the works on AOGs used low-level features (e.g., HOG features) to model the part appearance, which may limit their capacity and discriminative power. 
Furthermore, none of these works modeled occluders explicitly or tested their method on images with different level of occlusions, therefore it is unclear how those models can be made robust to occlusion.
CompositionalNets can be considered to be complementary to and-or-graphs. In fact, our mixture model can be interpreted as simple two-layered and-or-graph, where each mixture components combines parts (vMF kernels) for a certain object pose, and the final class score is an “or”-combination over the different object poses (mixture components). While our model could be generalized to have multiple layers with and-or-nodes to introduce more flexibility in the representation, the focus of this work is robustness to partial occlusion. Furthermore, our experiments show that a two-layered and-or graph seems to be good enough to achieve high performance at image classification and detection on popular datasets such as PASCAL3D+, MS-COCO and ImageNet. Moreover, and-or-graphs often require considerable amounts of supervision for the graph structure \cite{chen2008rapid,zhu2008max,li2013vehicle} or are not well interpretable \cite{zhu2008,song2013discriminatively}, whereas our graph is learned from class supervision only and still learns a meaningful human-interpretable representation. 

\subsection{Deep Compositional Models in Computer Vision}
An early work from Liao et al. \cite{liao2016learning} proposes to integrate compositionality into DCNNs by regularizing the feature representations of DCNNs to cluster during learning. Their qualitative results show that the resulting feature clusters resemble part-like detectors.
Zhang et al. \cite{zhang2018interpretable} also demonstrate that part detectors emerge in DCNNs by restricting the activations in feature maps to have a localized distribution.
However, these approaches have not been shown to enhance the robustness of deep models to partial occlusion. 
Other related works propose to regularize the convolution kernels to be sparse \cite{tabernik2016towards}, or to force feature activations to be disentangled for different objects \cite{stone2017teaching}. 
As the compositional model is not explicitly incorporated but rather implicitly encoded within the parameters of the DCNNs, the resulting models remain black-box and not expedcted to be robust to partial occlusion. 
A number of works \cite{li2019aognets,tang2018deeply,tang2017towards} use differentiable graphical models to integrate part-whole compositions into DCNNs. However, these models are purely discriminative and thus are also deep networks with no internal mechanism to account for partial occlusion.
Girshick et al. \cite{girshick2015deformable} discussed that compositional deformable part models can be formulated as neural networks. However, they do not evaluate their models' robustness to partial occlusion nor its explainability. 
Kortylewski et al. \cite{kortylewski2019compositional} propose to learn a generative dictionary-based compositional model using the features of a DCNN. Instead of merging the compositional model into the DCNN, they use it as a ``backup" for an independently trained DCNN. Only when the DCNN classification score falls below a certain threshold, the prediction will be substituted by the output of the compositional model.

\subsection{Explainable Computer Vision Models}
Many post-hoc methods have been proposed to explain what has been encoded in the intermediate layers of DCNNs.
Several works \cite{le2013building,zhou2015object} visualize a real or generated input that activates a filter most to study the roles of individual units inside neural networks. Similarly, Nguyen et al. \cite{nguyen2016synthesizing} synthesize prototypical images for individual units by learning a feature inversion mapping, while Bau et al. \cite{bau2017network} visualize segmentation masks extracted from filter activations and assign concept labels automatically. On the other hand, the works of \cite{mahendran2015understanding,simonyan2013deep,zeiler2014visualizing} use variants of back-propagation to identify or generate salient image features.
Moving beyond studying individual hidden units, Wang et al. \cite{wang2015unsupervised} use clusters of activations from all units in a layer and shows that the cluster centers yield better part detectors. Alain and Bengio \cite{alain2016understanding} probe mid-layer filters by training linear classifiers on the intermediate activations. They also analyze the information dynamics among layers and its effect on the final prediction. The work of Fong et al. \cite{fong2018net2vec} shows that filter embeddings better characterize the meaning of a representation and its relationship to other concepts. Most of these works evaluate their results using human judgments.

Unlike the post-hoc methods that focus on visualizing/analyzing pre-trained DCNNs, other approaches aim to learn more meaningful representations during the network training stage. The work of \cite{ross2017right} explains and regularizes differentiable models by examining and selectively penalizing their input gradients, but this method requires extra annotation from human experts. Hu et al. \cite{hu2016harnessing} regularize the learning process by introducing an iterative distillation method that transfers the structured information of logic rules into the weights of neural networks. Stone et al. \cite{stone2017teaching} encourage networks to form representations that disentangle objects from their surroundings and from each other, but they do not obtain part-level semantics explicitly. Sabour et al. \cite{sabour2017dynamic} propose a capsule model, which used a dynamic routing mechanism to parse the entire object into a parsing tree of capsules, and each capsule may encode a specific meaning. The work of Zhang et al. \cite{zhang2018interpretable} invents a generic loss to regularize the representation of a filter to improve its interpretability.
\\
\\
In this work, we unify generative compositional models and deep convolutional neural networks into a joint architecture with innate robustness to partial occlusion. 
The generative nature of the model enables it to localize occluders and to recognize objects based on the spatial configuration of visible object parts. 
CompositionalNets are naturally interpretable as their predictions can be understood in terms of part detection, occluder localization and viewpoint estimation.

%% file: 03_0_model_classification.tex
\section{CompositionalNets for Image Classification}
\label{sec:compnet}

In this section, we introduce CompositionalNets, a neural architecture design that replaces the fully-connected classification head of deep networks with a differentiable generative compositional model. We extend CompositionalNets to object detection in Section \ref{sec:detect} and discuss how CompositionalNets can be trained in an end-to-end manner in Section \ref{sec:e2e}.

\begin{figure}
    \centering
    \begin{subfigure}{\linewidth}
		\centering
		\includegraphics[width=.24\linewidth]{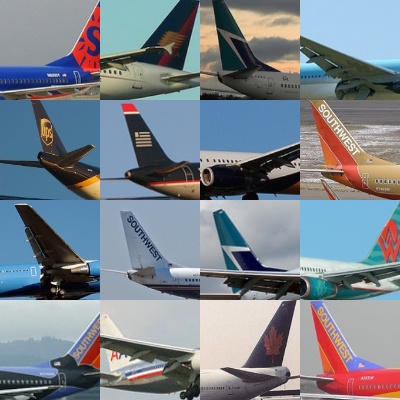}
		\includegraphics[width=.24\linewidth]{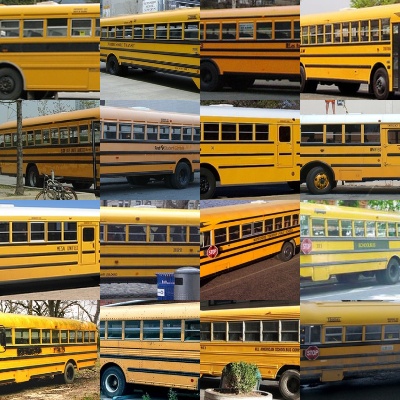}
		\includegraphics[width=.24\linewidth]{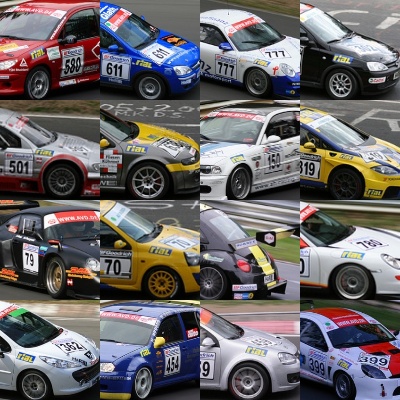}
        \includegraphics[width=.24\linewidth]{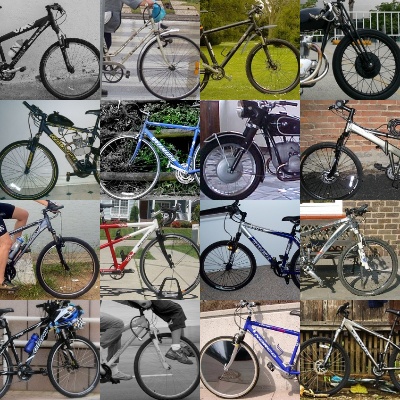}		
    \caption{}	
	\end{subfigure}
    \begin{subfigure}{\linewidth}
		\centering
		\includegraphics[width=.24\linewidth]{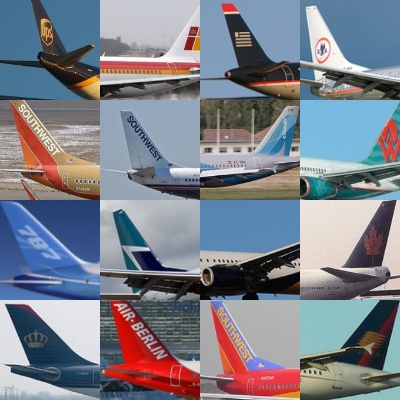}
		\includegraphics[width=.24\linewidth]{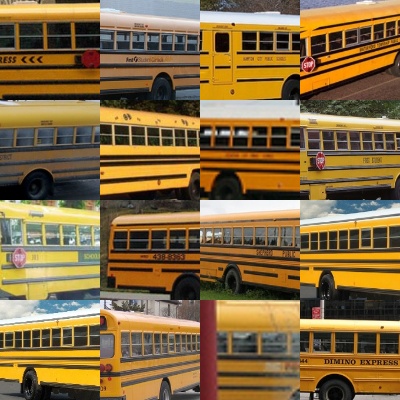}
		\includegraphics[width=.24\linewidth]{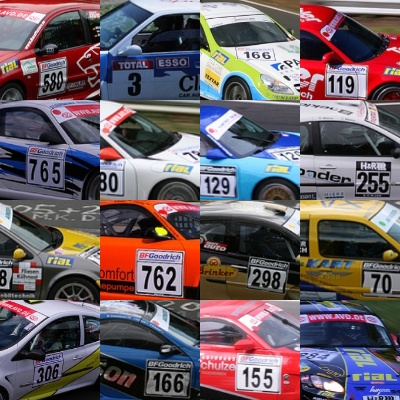}
        \includegraphics[width=.24\linewidth]{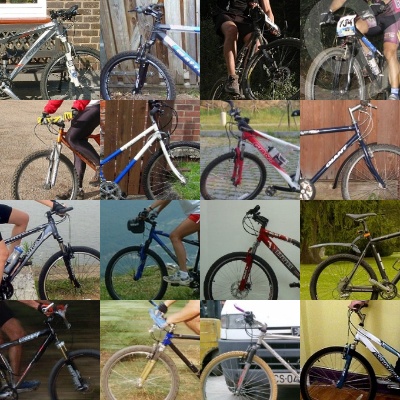}				
    \caption{}	
	\end{subfigure}	
    \begin{subfigure}{\linewidth}
		\centering
		\includegraphics[width=.24\linewidth]{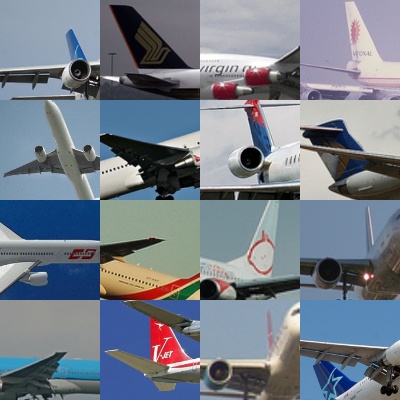}
		\includegraphics[width=.24\linewidth]{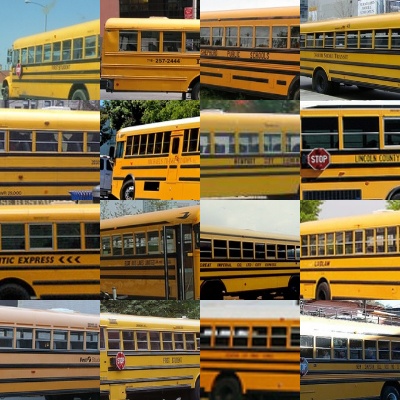}
		\includegraphics[width=.24\linewidth]{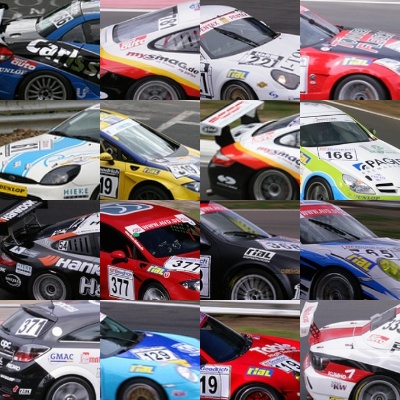}
        \includegraphics[width=.24\linewidth]{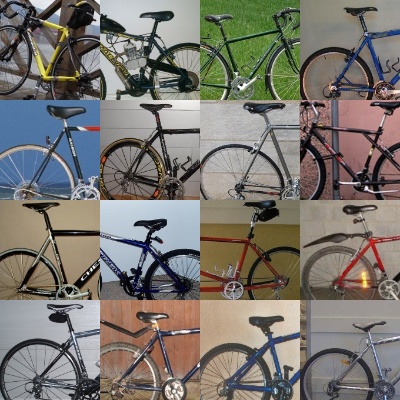}		
    \caption{}	
	\end{subfigure}		
	\caption{Illustration of vMF kernels $\mu_k$ learned from: (a) VGG-16, (b) ResNet50 and (c) ResNext. We visualize image patterns from the training data that activate a vMF kernel the most. Note that feature vectors that are similar to one of the vMF kernels, are often induced by image patches that have similar geometry and appearance. Furthermore, we were able to find vMF kernels of each backbone that represent similar visual concepts (vertical stabilizer, school bus side, number sticker on car, top of bicycle wheel).}
    \label{fig:vc}
\end{figure}

\subsection{A Generative Compositional Model of Neural Feature Activations}
\label{sec:vmf}

We denote a feature map $F^l\in \mathbb{R}^{H\times W \times D}$ 
as the output of a layer $l$ in a DCNN, with $D$ being the number of channels. $f^l_i \in \mathbb{R}^D$ is a feature vector in $F^l$ at position $i$ on the 2D lattice $\mathcal{P}$ of the feature map.
In the remainder of this section we omit the superscript $l$ for notational simplicity because this is fixed a-priori in our experiments.

Our goal is to learn a generative model $p(F|y)$ of the real-valued feature activations $F$ for an object class $y$. 
In the following, we assume the viewpoint of the object to be known. Later, we generalize the model to 3D objects with varying viewpoints. 
We define the probabilistic model $p(F|y)$ to be a mixture of von-Mises-Fisher (vMF) distributions:
\begin{align}
p(F|\theta_y) &=  \prod_{i} p(f_i|\mathcal{A}_{i,y},\Lambda) \label{eq:vmf}\\
p(f_i|\mathcal{A}_{i,y},\Lambda) &= \sum_k \alpha_{i,k,y} p(f_i|\lambda_k),\label{eq:vmf2}
\end{align}
where $\theta_y= \{\mathcal{A}_y,\Lambda\}$ are the model parameters and  $\mathcal{A}_y=\{\mathcal{A}_{i,y}\}$ are the parameters of the mixture models at every position $i \in \mathcal{P}$ on the 2D lattice of the feature map $F$. 
Note that the probabilistic model defined in Equation \ref{eq:vmf} has a tree-like independence structure and therefore enables efficient inference \cite{kortylewski2017greedy}. 
Moreover, 
\begin{equation}
\mathcal{A}_{i,y} = \{\alpha_{i,0,y},\dots,\alpha_{i,K,y}|\sum_{k=0}^K \alpha_{i,k,y} = 1\}
\end{equation}
are the mixture coefficients, 
$K$ is the number of mixture components and $\Lambda = \{\lambda_k = \{\sigma_k,\mu_k \} | k=1,\dots,K \}$ are the variance and mean of the vMF distribution:
\begin{equation}
\label{eq:vmfprob}
p(f_i|\lambda_k) = \frac{e^{\sigma_k \mu_k^T f_i}}{Z(\sigma_k)}, \norm{f_i} = 1, \norm{\mu_k}= 1, 
\end{equation}
where $Z(\sigma_k)$ is the normalization constant. As $Z(\sigma_k)$ is difficult to compute for high-dimensional data \cite{banerjee2005clustering}, we assume $\sigma_k$ to be fixed a-priori. Hence, the normalization constant is the same for each mixture component and does not need to be computed explicitly when optimizing the parameters.
After learning the vMF cluster centers $\{\mu_k\}$ with maximum likelihood optimization, they resemble feature activation patterns that frequently occur in the training data. 
Interestingly, feature vectors that are similar to one of the vMF cluster centers, are often induced by image patches that are similar in appearance and often even share semantic meanings (see Figure \ref{fig:vc}). 
This property was also observed in a number of related works that used clustering in the neural feature space \cite{wang2015unsupervised,liao2016learning,wang2017detecting}. 
Subsequently, we learn the mixture coefficients $\alpha_{i,k,y}$ with maximum likelihood estimation from the training images.
Intuitively, $\alpha_{i,k,y}$ describes the expected activation of a cluster center $\mu_k$ at a position $i$ in a feature map $F$ for a class $y$. 

\subsection{Viewpoint Modeling}
An important property of convolutional networks is that the spatial information from the image is preserved in the feature maps. 
Hence, the set of mixture coefficients $\mathcal{A}_{y}$ intuitively can be thought of as being a 2D template that describes the expected spatial activation pattern of parts in an image for a given class $y$ - e.g. where the tires of a car are expected to be located in an image.
Therefore, our proposed vMF model intuitively accumulates the part detections spatially into a hypothesis about the objects' presence. Note that this implements a part-based voting mechanism.

As the spatial pattern of parts varies significantly with the 3D pose of the object, we represent 3D objects as a mixture of compositional models:
\begin{equation}
\label{eq:mix}
p(F|\Theta_y) = \sum_m \nu^m p(F|\theta^m_y),
\end{equation}
with  $\mathcal{V}\texttt{=}\{\nu^m \in\{0,1\}, \sum_m \nu^m\texttt{=} 1 \}$ and $\Theta_y = \{\theta^m_y, m=1,\dots,M\}$. Here $M$ is the number  of compositional models in the mixture distribution and $\nu_m$ is a binary assignment variable that indicates which mixture component is active.
Intuitively, each mixture component $m$ will represent a different viewpoint of an object (see Experiments in Section \ref{sec:exp_view}).

The parameters of the mixture components $\{\mathcal{A}^m_y\}$ need to be learned in an EM-type manner by iterating between estimating the assignment variables $\mathcal{V}$ and maximum likelihood estimation of $\{\mathcal{A}^m_y\}$. We discuss how this process can be performed in a neural network in Section \ref{sec:e2e_class}. 

\subsection{Occlusion modeling} 
Following the approach presented in \cite{kortylewski2017model}, compositional models can be augmented with an occlusion model. The intuition behind an occlusion model is that at each position $i$ in the image either the object model $p(f_i|\mathcal{A}^m_{i,y},\Lambda)$ or an occluder model $p(f_i|\beta,\Lambda)$ is active (note that this is closely related to robust statistics \cite{huber2011robust}):
\begin{align}
	&p(F|\theta^m_y,\beta)\hspace{-0.075cm} =\hspace{-0.075cm} \prod_{i} p(f_i,z^m_i \hspace{0.05cm}\texttt{=}\hspace{0.05cm}0)^{1-z^m_i} p(f_i,z^m_i\hspace{0.05cm}\texttt{=}\hspace{0.05cm}1)^{z^m_i},\label{eq:occ}\\
	&p(f_i,z^m_i\hspace{0.05cm}\texttt{=}\hspace{0.05cm}1) = p(f_i|\beta,\Lambda)\hspace{0.1cm}p(z^m_i\texttt{=}1),\label{eq:occ2}\\
	&p(f_i,z^m_i\hspace{0.05cm}\texttt{=}\hspace{0.05cm}0) = p(f_i|\mathcal{A}^m_{i,y},\Lambda)\hspace{0.1cm}(1\texttt{-}p(z^m_i\texttt{=}1))\label{eq:occ3}.
\end{align}
The binary variables $\mathcal{Z}^m=\{z^m_i \in \{0,1\} | i \in \mathcal{P}\}$ indicate if the object is occluded at position $i$ for mixture component $m$. 
The occlusion prior $p(z^m_i\texttt{=}1)$ is fixed a-priori. 
We use a mixture of several occluder models that are learned in an unsupervised manner: 
\begin{align}
p(f_i|\beta,\Lambda) &= \prod_n p(f_i|\beta_n,\Lambda)^{\tau_n} \\
&=\prod_n \Big(\sum_{k} \beta_{n,k} p(f_i|\sigma_k,\mu_k)\Big)^{\tau_n},
\end{align}
where \{$\tau_n \in \{0,1\},\sum_n \tau_n =1\}$ indicates which occluder model explains the data best.
Note that the model parameters $\beta$ are independent of the position $i$ in the feature map and thus the model has no spatial structure. 

\begin{figure}
    \centering
    \begin{subfigure}{\linewidth}
    	\begin{subfigure}{\linewidth}
    		\centering
    		\includegraphics[width=0.32\linewidth]{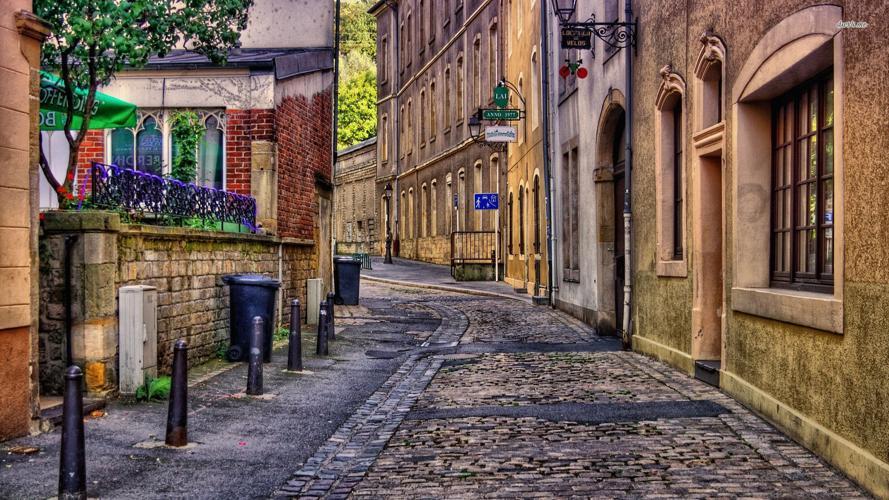}
    		\includegraphics[width=0.32\linewidth]{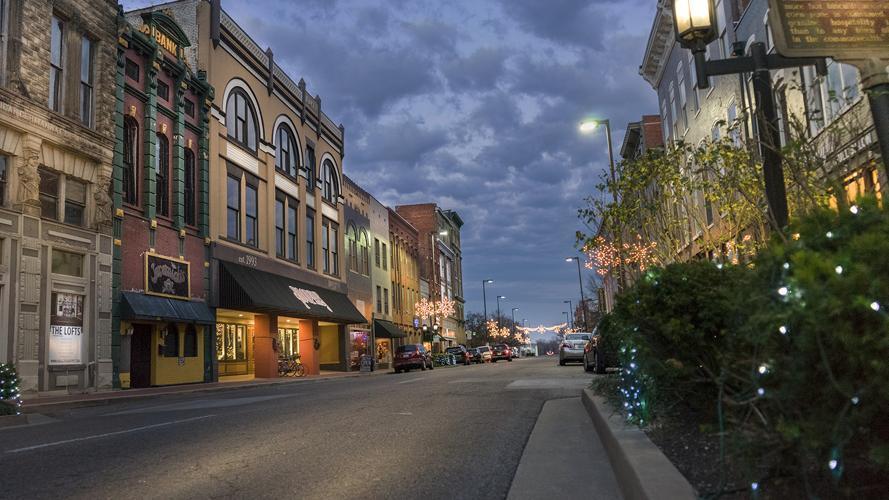}
    		\includegraphics[width=0.32\linewidth]{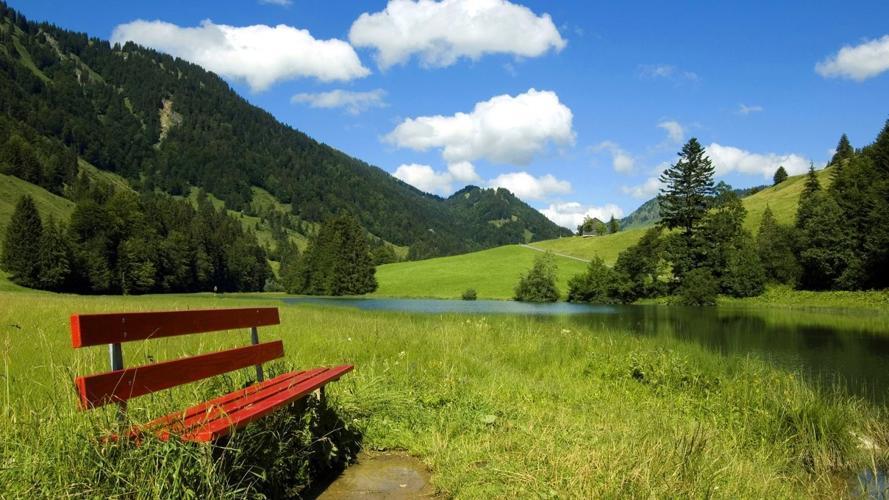}
    	\end{subfigure}\\
    	\begin{subfigure}{\linewidth}
    		\centering
    		\includegraphics[width=0.32\linewidth]{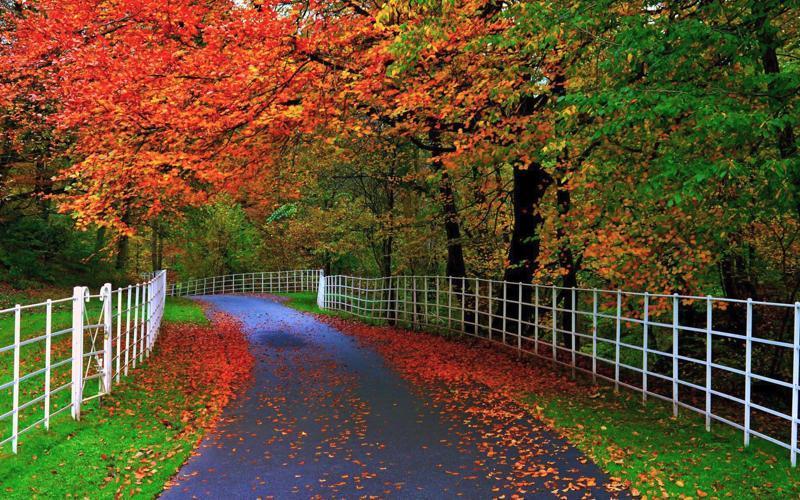}
    		\includegraphics[width=0.32\linewidth]{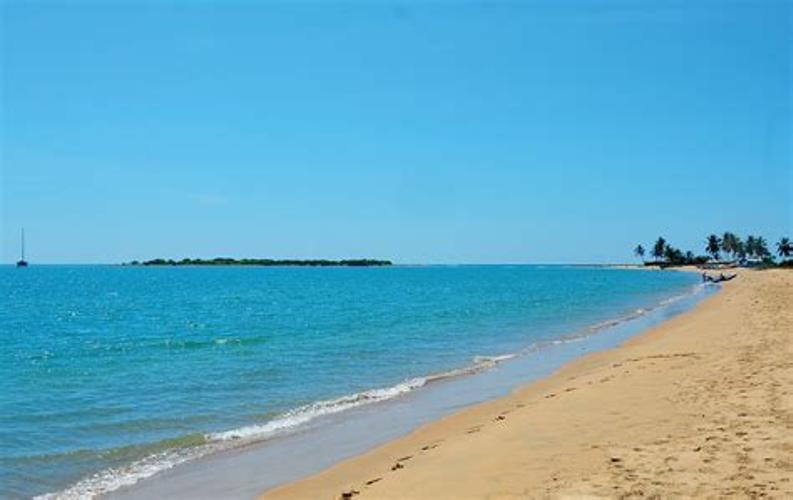}
    	\end{subfigure}%
    	\caption{}
    	\label{fig:nat-a}
	\end{subfigure}
	\begin{subfigure}{\linewidth}
    	\includegraphics[width=0.19\linewidth]{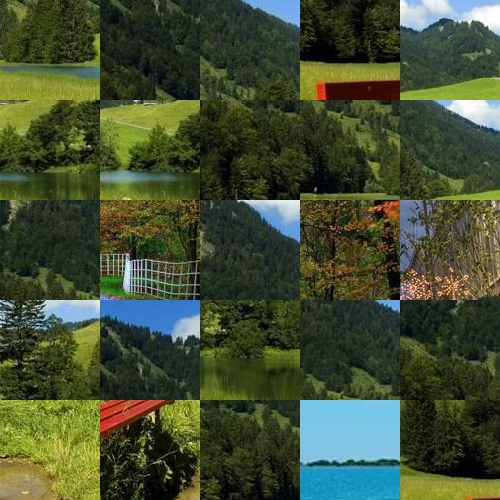}
    	\includegraphics[width=0.19\linewidth]{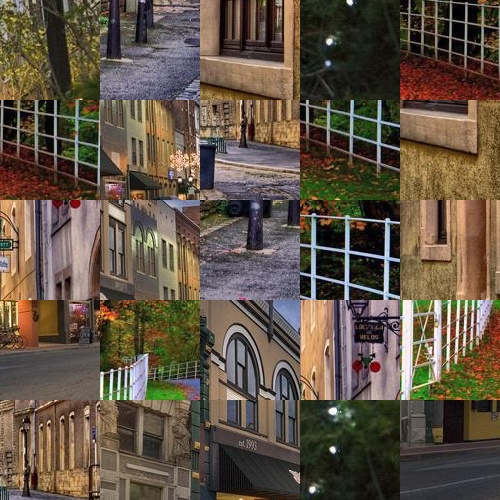}
    	\includegraphics[width=0.19\linewidth]{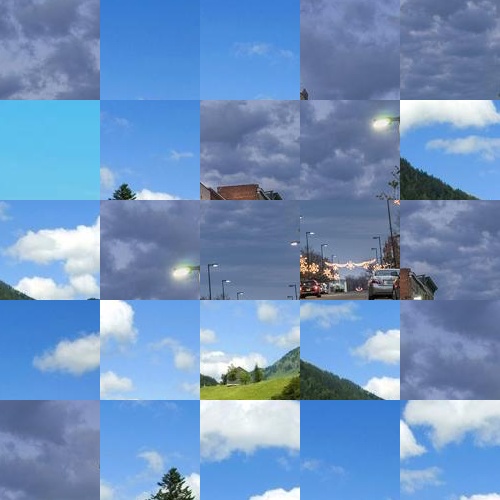}
    	\includegraphics[width=0.19\linewidth]{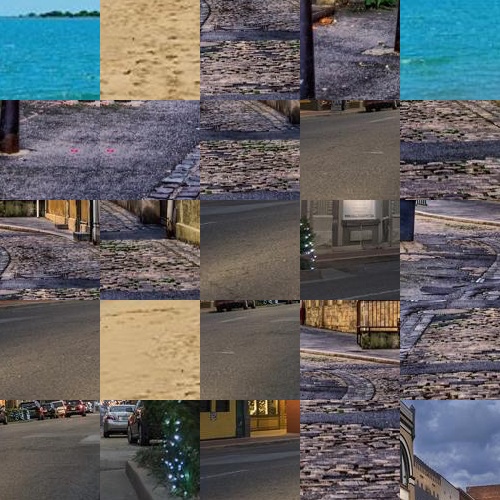}
    	\includegraphics[width=0.19\linewidth]{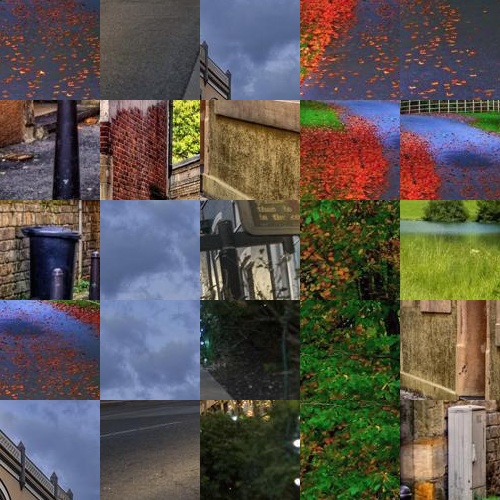}
    	\caption{}
    	\label{fig:nat-b}
	\end{subfigure}
	\caption{The occluder models are learned from natural images (a). Note that no target object is present in any of these images. (b) illustrates patches from the training data in (a) that have the highest likelihood for each of five occluder models. Note how some models focus more on uniform colored patches while others focus more on textured patches.}
    \label{fig:backgrounds}
\end{figure}

The parameters of the occluder models $\beta_n$ are learned from clustered features of random natural images that do not contain any object of interest (see Figure \ref{fig:nat-a}). 
Hence, the mixture coefficients $\beta_{n,k}$ intuitively describe the expected activation of $\mu_k$ in regions of natural images. 
While it is possible to use just a single occluder model \cite{kortylewski2019compositional}, we found that having a mixture model slightly increases the classification performance and the occluder localization performance. 
The reason is that different occluder models can focus at explaining part distributions for different typical regions in images e.g. uniform colored image patches or textured regions. 
We illustrate this in Figure \ref{fig:nat-b} by visualizing patches from the training data in Figure \ref{fig:nat-a} that have the highest likelihood for five different occluder models.
Note that the purpose of the occluder models is not to be purely specific to one particular texture type, they also need to be general enough to explain a variety of local image patterns which the object model (Eq. \ref{eq:vmf}) is not able to explain well. Therefore, there is a trade-off between specificity and generality of the occluder models. We found that using five models balances this trade-off well.

%% file: 03_1_model_detection.tex
\section{Object Detection with CompositionalNets}
\label{sec:detect}
Object detection involves the estimation of the object class, the object center, and the object scale (a bounding box around the object). 
We find that partial occlusion can have significant negative effects on all three tasks.
In the following section, we generalize CompositionalNets to enable robust object detection under partial occlusion. 
We first discuss how CompositionalNets can be generalized to locate objects in images by introducing a detection layer (Section \ref{sec:detect_layer}).
Subsequently, we show how robust bounding box estimates can be obtained with an advanced compositional voting mechanism in Section \ref{sec:bbox}.
Finally, we discuss the importance of separating the representations of objects and their context in object detection, and show how this can be achieved with CompositionalNets in Section \ref{sec:context}.

\subsection{CompositionalNets for Object Localization}
\label{sec:detect_layer}
\begin{figure}
\centering
\includegraphics[height=4cm]{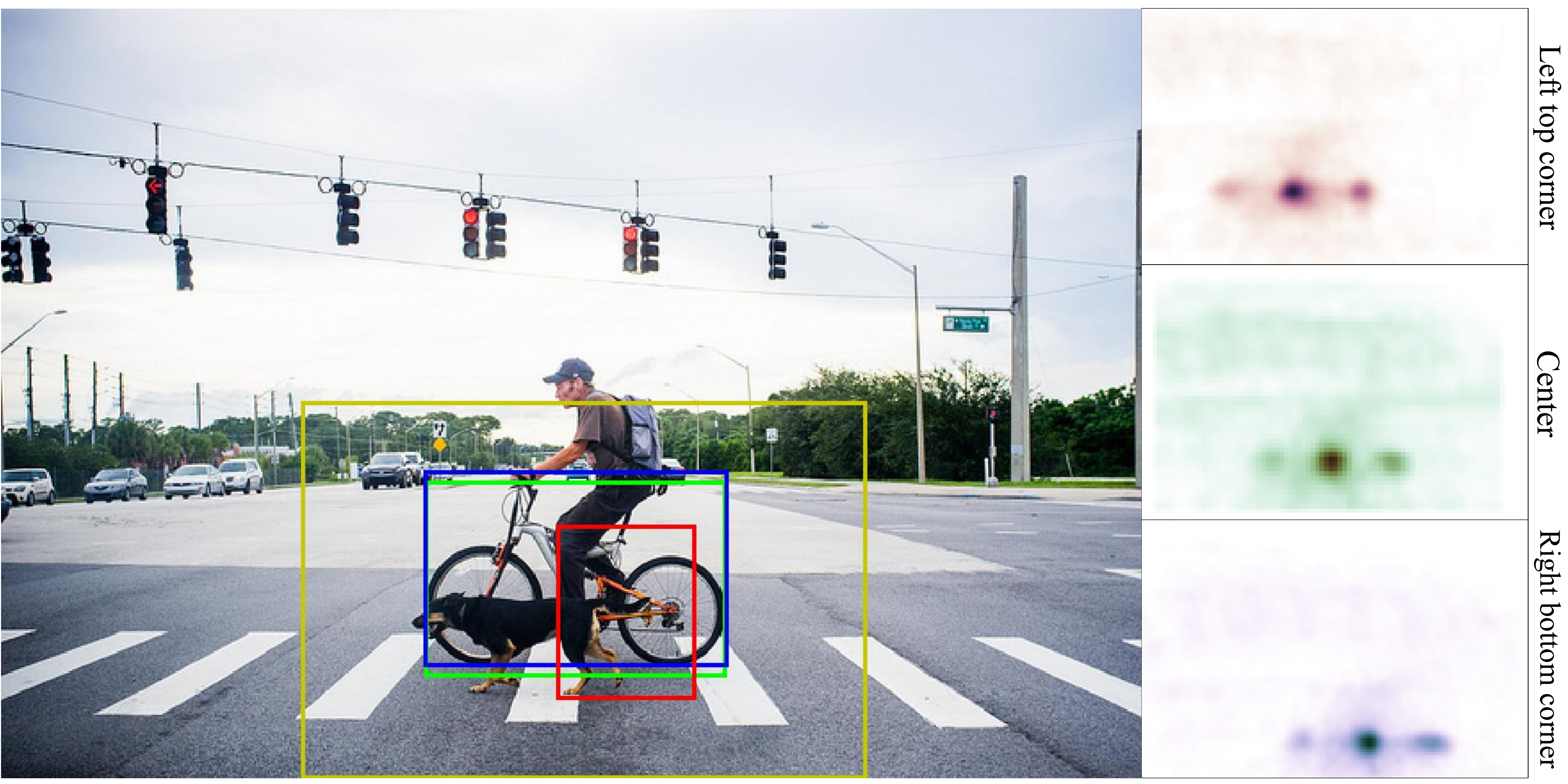}
\caption{Example of robust bounding box voting results. Blue box: Ground truth. Red box: Bounding box by Faster R-CNN. Yellow box: RPN + CompNet. Green box: CompNet + part-based bounding box voting. Our proposed part-based voting mechanism generates probability maps (right) for the object center (cyan point), the top left corner (purple point) and the bottom right corner (yellow point) of the bounding box. The final bounding box estimate is generated by combining the voting results.}
\label{fig:corner}
\end{figure}

A natural way of generalizing CompositionalNets to object detection is to combine them with region proposal networks (RPNs). 
However, RPNs are typically trained end-to-end and therefore cannot predict the bounding box of strongly occluded objects reliably (see our experiments in Section \ref{sec:exp_detect}). 
Figure \ref{fig:corner} illustrates this limitation by depicting the detection results of Faster R-CNN trained with CutOut \cite{devries2017improved} (red box) and a combination of RPN and CompositionalNet (yellow box). 
We address this limitation by generalizing CompositionalNets to object localization. In particular, we introduce a detection layer that accumulates votes for the object center for all mixtures $m$ over the positions $i$ in the feature map $F$.
In order to achieve this, we compute the object likelihood in a scanning window manner.
Thus, we shift the feature map, w.r.t. the object model along all points $i$ from the 2D lattice of the feature map.
This process will generate a spatial likelihood map:
\begin{equation}
\label{eq:detect_map}
    R = \{p(F_ {i}|\Theta_y)| \forall i \in \mathcal{P}\}, 
\end{equation}
where $F_ {i}$ denotes the feature map centered at the position $i$. 
Using this simple generalization we can perform object localization by selecting all maxima in $R$ above a threshold $t$ after non-maximum suppression. Our proposed detection layer can be implemented efficiently on modern hardware using convolution-like operations (see Section \ref{sec:exp_detect} for more details).

\subsection{Robust Compositional Bounding-Box Voting} 
\label{sec:bbox}
While CompositionalNets can be generalized to localize partially occluded objects using our proposed detection layer, estimating the bounding box of an object under occlusion is more difficult because a significant amount of the object might not be visible (Figure \ref{fig:corner}). 
We solve this problem by generalizing the part-based voting mechanism in CompositionalNets to vote for the bounding box corners in addition to voting for the object center.
In this way, we can leverage the viewpoint-specific spatial distribution of part activations for bounding-box estimation. This makes the process highly robust to missing parts.
In particular, we learn additional mixture components that model the expected feature activations around bounding box corners $p(F_{i}|\Theta_y^c)$, where $c=\{ct,br,tl\}$ are the object center $ct$ and two opposite bounding box corners $\{br,tl\}$.
Figure \ref{fig:corner} illustrates the spatial likelihood maps $R^c$ of all three models.
We generate a bounding box using the two points that have maximal likelihood.
Note how in Figure \ref{fig:corner} the bounding box can be localized accurately, despite the partial occlusion of the object.
We discuss how the model can be learned in an end-to-end manner in Section \ref{sec:e2e_detect}.

\subsection{Context-aware CompositionalNets}
\label{sec:context}
\begin{figure}
\centering
\includegraphics[height=4cm]{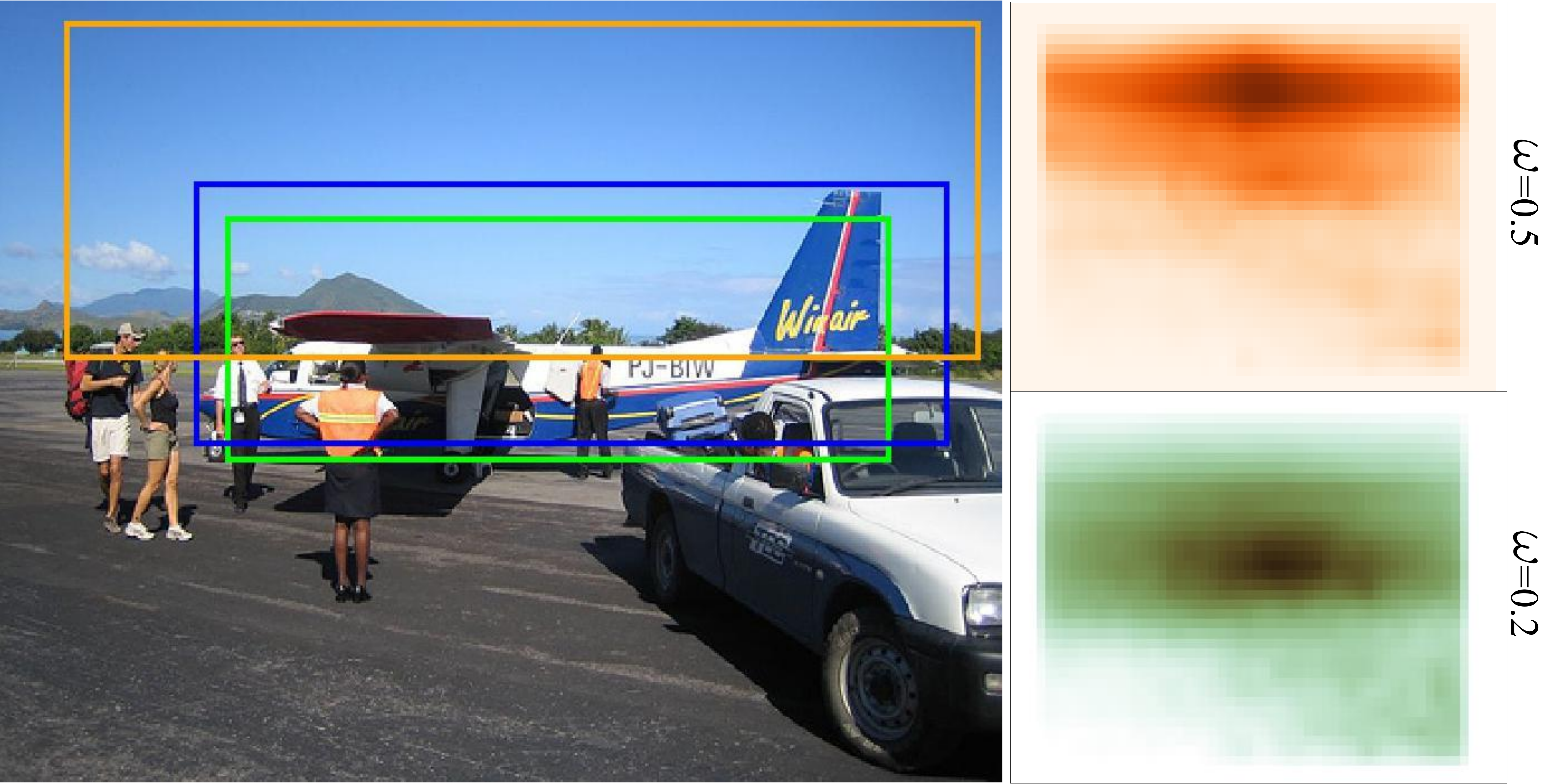}
\caption{Influence of context in airplane detection under occlusion. Blue box: Ground truth. Orange box: Bounding box of CompositionalNets ($\omega=0.5$). Green box: Bounding box of Context-aware CompositionalNets ($\omega=0.2$).
Probability maps of the object center are on the right. 
Note how reducing the influence of the context improves the localization response.
}
\label{fig:context_bad}
\end{figure}

While in image classification, the object of interest often dominates a large part of the image, in object detection the object is embedded in a larger context that is often biased in the training data (e.g. airplanes often have blue background).
This gives rise to a critical problem when aiming to detect partially occluded objects.
Intuitively, the objects' context can be useful for recognizing objects due to biases in the data. 
However, relying too strongly on context can be misleading, because if objects are strongly occluded (Figure \ref{fig:context_bad}) the detection thresholds must be lowered. This, in turn, increases the influence of the objects context and leads to false-positive detections in regions with no object. 
Hence, it is important to have control over the influence of contextual cues on the detection result. 

\begin{figure}
    \centering
    \includegraphics[width=\linewidth]{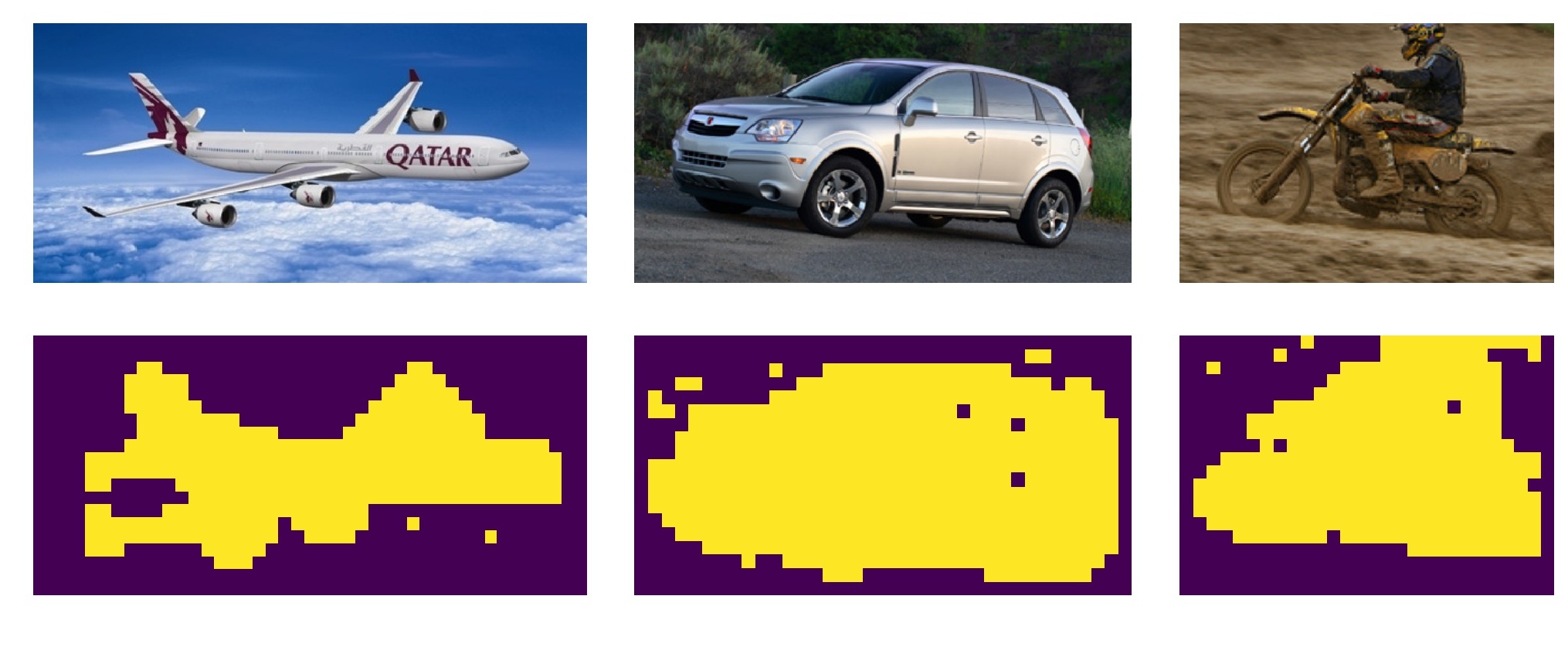}
    \caption{Context segmentation results.
    A standard CompositionalNet learns a joint representation of the image including the context. Our context-aware CompositionalNet will separate the representation of the context from that of the object based on the illustrated segmentation masks.}
    \label{fig:seg}
\end{figure}

We overcome this issue by explicitly separating the representation of the context from that of the object, to control the influence of the contextual information on the detection result. In particular, we represent the feature map $F$ as a mixture of two models:
\begin{align}
\label{eq:context_likely}
    	p(f_i|\mathcal{A}^m_ {i,y},\chi^m_ {i,y},\Lambda)=
    	&\omega \hspace{0.1cm}p(f_i|\chi^m_ {i,y},\Lambda) + \nonumber \\
    	&(1-\omega) p(f_i|\mathcal{A}^m_ {i,y},\Lambda).
\end{align}
Here $\{{A}^m_ {i,y},\chi^m_ {i,y}\}$ are the parameters of the context-aware model that is defined to be a mixture of vMF likelihoods (Equation \ref{eq:vmf2}).
The parameter $\omega$ controls the trade-off between context and object, and is fixed a-priori at test time. 
Note that setting $\omega=0.5$ retains the CompositionalNet as proposed in Section \ref{sec:compnet} as the context and object would be weighted equally.
Figure \ref{fig:context_bad} illustrates the benefits of reducing the influence of the context on the detection result under partial occlusion.
The context parameters $\chi^m_ {i,y}$ and object parameters $\mathcal{A}^m_ {i,y}$ can be learned from the training data using maximum likelihood estimation, assuming a binary assignment $\pi_i$ of the feature vectors $f_i$ in the training data to either the context or the object. 
To obtain this binary assignment, we need to segment the training images into context and object based on the available bounding box annotation. We do so, by assuming that any feature vector with a receptive field outside of the scope of the bounding boxes can be considered to be context. 
Based on this assumption, we cluster the context features using K-means++ \cite{k-means++} to generate a dictionary of context feature centers $\mathcal{C}=\{\mathcal{C}_q \in \mathbb{R}^D |q= 1,\dots,Q \}$. 
We apply a threshold on the cosine similarity $s(\mathcal{C},f_i)=\max_q [(\mathcal{C}_q^T f_i)/(\norm{\mathcal{C}_q}\norm{f_i})]$ to segment the context and the object in any given training image (Figure \ref{fig:seg}). 

%% file: 03_2_model_inference.tex
\begin{figure*}[t]
    \centering
    \includegraphics[width=\linewidth]{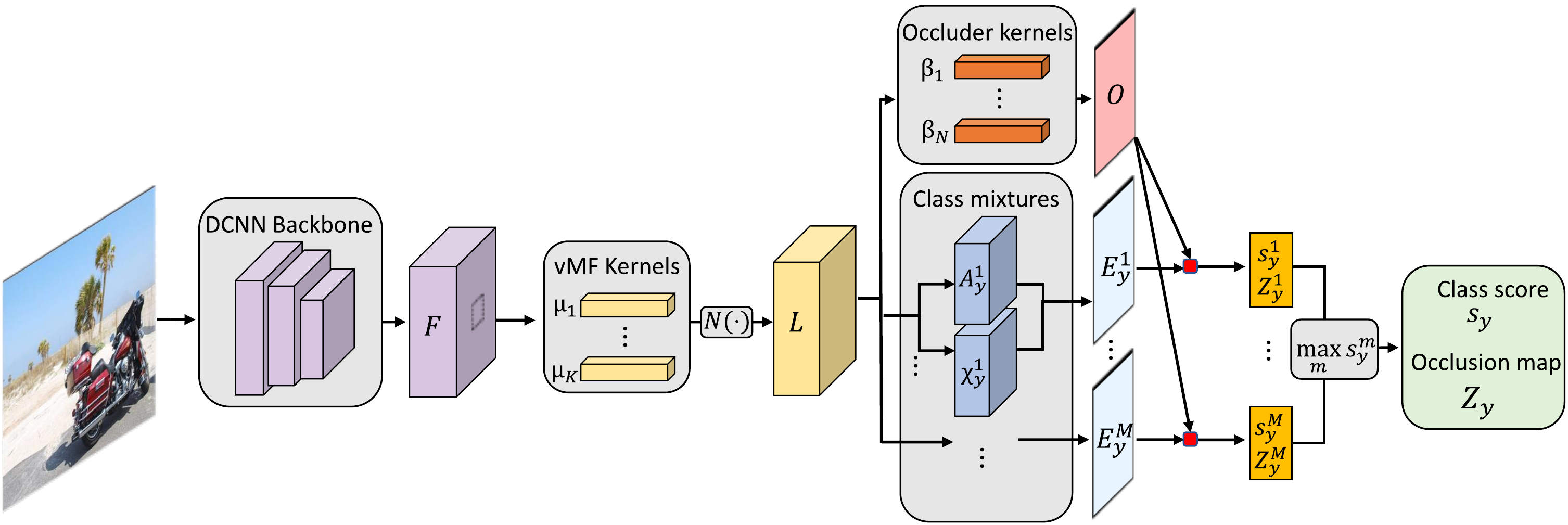}
    \caption{Feed-forward inference with a context-aware CompositionalNet. A DCNN backbone is used to extract the feature map $F$, followed by a convolution with the vMF kernels $\{\mu_k\}$ and a non-linear vMF activation function $\mathcal{N}(\cdot)$. The resulting vMF likelihood $L$ is used to compute the occlusion likelihood $O$ using the occluder kernels $\{\beta_n\}$.
    Furthermore, $L$ is used to compute the context-aware mixture likelihoods $\{E^m_y\}$ using the mixture models of the object $\{A_y^{m}\}$ and the context $\{\chi_y^{m}\}$. 
    $O$ and $\{E^m_y\}$ compete in explaining $L$ (red box) and are combined to compute an occlusion robust score $\{s^m_y\}$. The binary occlusion maps $\{Z^m_y\}$ indicate which positions in $L$ are occluded.
    The final class score $s_y$ is computed as $s_y\texttt{=}\max_m s^m_y$ and the occlusion map $Z_y$ is selected accordingly.}
    \label{fig:model}
\end{figure*}
\section{End-to-End Training of CompositionalNets}
\label{sec:e2e}
In the following, we show how inference in CompositionalNets can be formulated as feed-forward neural network (Section \ref{sec:inference}) and discuss how CompositionalNets can be trained end-to-end for image classification (Section \ref{sec:e2e_class}) and object detection (Section \ref{sec:e2e_detect}).

\subsection{Inference as Feed-Forward Neural Network}
\label{sec:inference}
The computational graph of our proposed fully generative compositional model is acyclic. 
Hence, we can perform inference in a single forward pass as illustrated in Figure \ref{fig:model}. We use a standard DCNN backbone to extract a feature representation $F = \psi(I,\Omega) \in \mathbb{R}^{H \times W \times D}$ from the input image $I$, where $\Omega$ are the parameters of the feature extractor (purple tensor in Figure \ref{fig:model}). The pipeline after the feature extractor illustrates the computation of the model likelihood $p(F_i|\Theta)$ when the model is centered at position $i$ in the feature map (illustrated by the dotted black square on the feature tensor $F$). For image classification, the model will always be positioned at the image center, whereas for object detection, the model will be evaluated at every position in $F$ (as defined in Equation \ref{eq:detect_map}). We omit the subscript in $F_i$ in the following for notational clarity.\\

After feature extraction the model computes the vMF likelihood function $p(f_i|\lambda_k)$ (Equation \ref{eq:vmfprob}). The function is composed of two operations: An inner product $b_ {i,k}=\mu_k^T f_i$ and a non-linear transformation $\mathcal{N}=\exp(\sigma_k b_ {i,k})/Z(\sigma_k)$.
Since $\mu_k$ is independent of the position $i$, computing $b_ {i,k}$ is equivalent to a $1\times1$ convolution of $F$ with $\mu_k$. 
Hence, the vMF likelihood (Figure \ref{fig:model} yellow tensor) can be computed by :
\begin{equation}
L=\{\mathcal{N}(F\ast\mu_k)|k=1,\dots,K\} \in \mathbb{R}^{H\times W \times K}.
\end{equation}
The mixture likelihoods $p(f_i|\mathcal{A}^m_ {i,y},\chi^m_ {i,y},\Lambda)$ (Equation \ref{eq:context_likely}) are computed for every position $i$ as a dot-product between the mixture coefficients $\{\mathcal{A}^m_ {i,y},\chi^m_ {i,y}\}$ and the corresponding vector $l_i\in\mathbb{R}^K$ from the vMF likelihood tensor $L$: 
\begin{equation}
E^m_y = \{(1-\omega) l_i^T \mathcal{A}^m_ {i,y} + \omega l_i^T \chi^m_ {i,y} |\forall i \in\mathcal{P}\} \in \mathbb{R}^{H\times W},    
\end{equation}
(Figure \ref{fig:model} blue planes). 
Similarly, the occlusion likelihood can be computed as $O=\{\max_n l_i^T \beta_{n}| \forall i\in\mathcal{P}\}\in\mathbb{R}^{H \times W}$
(Figure \ref{fig:model} red plane).
Together, the occlusion likelihood $O$ and the mixture likelihoods $\{E^m_y\}$ are used to estimate the overall likelihood of the individual mixtures as $s^m_y=\log p(F|\theta_y^m,\beta)=\sum_i \max(E^m_ {i,y},O_i)$. 
The final model likelihood is computed as $s_y= \log p(F|\Theta_y)=\max_m s^m_y$ and the final occlusion map is selected accordingly as $\mathcal{Z}_y = \mathcal{Z}^{\bar{m}}_y \in \mathbb{R}^{H \times W}$ where $\bar{m}=\argmax_m s^m_y$.

\subsection{End-to-end Training for Image Classification}
\label{sec:e2e_class}
Our model is fully differentiable and can be trained end-to-end using backpropagation.
In our image classification experiments, context-awareness does not have a significant influence as the background is largely cropped out. 
Therefore, we use CompositionalNets as defined in Section \ref{sec:compnet} for image classification.
Hence, the trainable parameters of a CompositionalNet are $T=\{\Lambda,\mathcal{A}_y\}$.
We optimize those parameters jointly using stochastic gradient descent. The loss function is composed of three terms:
\begin{align}
    \mathcal{L}(y,y',F,T) =  & \mathcal{L}_{class}(y,y') + \gamma_1 \mathcal{L}_{vmf}(F,\Lambda) + \\&\gamma_2 \mathcal{L}_{mix}(F,\mathcal{A}_y).
\end{align}
$\mathcal{L}_{class}(y,y')$ is the cross-entropy loss between the network output $y'$ and the true class label $y$. We use a temperature $\mathcal{T}$ in the softmax classifier: $f(y)_i=\frac{e^{y_i\cdot \mathcal{T}}}{\Sigma_{i}e^{y_i\cdot \mathcal{T}}}$, with $\mathcal{T}=2$. 
$\mathcal{L}_{vmf}$ and $\mathcal{L}_{mix}$ regularize the parameters of the compositional model to have maximal likelihood for the features in $F$.
The parameters $\{\gamma_1,\gamma_2\}$ control the trade-off between the loss terms. 

The vMF cluster centers $\mu_k$ are learned by maximizing the vMF-likelihoods (Equation \ref{eq:vmfprob}) for the feature vectors $f_i$ in the training images. We keep the vMF variance $\sigma_k$ constant, which also reduces the normalization term $Z(\sigma_k)$ to a constant. 
We assume a hard assignment of the feature vectors $f_i$ to the vMF clusters during training. Hence, the free energy to be minimized for maximizing the vMF likelihood \cite{wang2017visual} is:
\begin{align}
\label{eq:loss_vmf}
    \mathcal{L}_{vmf}(F,\Lambda)  &= - \sum_i \max_k \log p(f_i|\mu_k)\\
                        &= C \sum_i \max_k \mu_k^T f_i,
\end{align}
where $C$ is a constant. Intuitively, this loss encourages the cluster centers $\mu_k$ to be similar to the feature vectors $f_i$. 

In order to learn the mixture coefficients $\mathcal{A}^m_{y}$ we need to maximize the model likelihood (Equation \ref{eq:mix}). 
We can avoid an iterative EM-type learning procedure by making use of the fact that the the mixture assignment $\nu_m$ and the occlusion variables $z_ {i}$ have been inferred in the forward inference process.
Furthermore, the parameters of the occluder model are learned a-priori and then fixed. Hence the energy to be minimized for learning the mixture coefficients is:
\begin{equation}
\label{eq:loss_mix}
    \mathcal{L}_{mix}(F,\mathcal{A}_y) =\hspace{-.05cm}\texttt{-}\hspace{-.1cm}\sum_i\hspace{-0.05cm}(1\texttt{-}z^\uparrow_ {i}) \log \hspace{-0.05cm}\Big[\hspace{-0.05cm}\sum_k\hspace{-0.1cm}\alpha^{m^\uparrow}_ {i,k,y}p(f_i|\lambda_k)\Big]
\end{equation}
Here, $z^\uparrow_ {i}$ and $m^\uparrow$ denote the variables that were inferred in the forward process (Figure \ref{fig:model}). 

\subsection{Training CompositionalNets for Object Detection}
\label{sec:e2e_detect}
For object detection, we train context-aware CompositionalNets as proposed in Section \ref{sec:detect}.
The trainable parameters of the model are $T^c=\{\Lambda,\{\mathcal{A}^c_y\},\{\chi^c_y\}\}$ where $c\in\{ct,br,tl\}$.
The loss function has three main objectives. The model should explain data with maximal likelihood ($\mathcal{L}_{g}$), while also localizing ($\mathcal{L}_{detect}$) and classifying ($\mathcal{L}_{class}$) the object accurately in the training images. : 
\begin{align}
    \label{eq:loss}
    \mathcal{L} &= \mathcal{L}_{class}(y, y') +
    \epsilon_1 \sum_c \big( \mathcal{L}_g(F^c,T^c) \\
    &+ \epsilon_2 \mathcal{L}_{detect}(\hat{S}^c,\bar{S}^c,F,T^c) \big), 
\end{align}
where $\epsilon_1, \epsilon_2$ control  the  trade-off  between  the  loss terms.
While $\mathcal{L}_g$ is learned from images $\hat{I^c}$ with feature maps $F^c$ that are centered at $c\in\{c,bl,tr\}$, the other losses are learned from unaligned training images $I$ with feature maps $F$.

\textbf{Generative Regularization.} The model is regularized to be generative in terms of the neural feature activations by optimizing:
\begin{align}
    \mathcal{L}_{g}(F^c,T) =  
     &\mathcal{L}_{vmf}(F^c,\Lambda) \\
     &+ \sum_i \mathcal{L}_{con}(f^c_i,\mathcal{A}^c_ {i,y},\chi^c_ {i,y}), 
\end{align}
where $\mathcal{L}_{vmf}(F^c,\Lambda)$ is defined in Equation \ref{eq:loss_vmf} and the parameters of the context-aware model $\mathcal{A}^c_y$ and $\chi^c_y$ are learned by optimizing the context loss:
\begin{align}
    \mathcal{L}_{con}(f^c_i,\mathcal{A}^c_ {i,y},\chi^c_ {i,y}) = &\pi_i \mathcal{L}_{mix}(f^c_i,\mathcal{A}^c_ {i,y})+\\
    &(1-\pi_i)\mathcal{L}_{mix}(f^c_i,\chi^c_ {i,y}).
\end{align}
Here, $\pi_i \in \{0,1\}$ is a context assignment variable that indicates if a feature vector $f_i$ belongs to the context or to the object model and $\mathcal{L}_{mix}$ is defined in Equation \ref{eq:loss_mix}. We estimate the context assignments a-priori using a simple binary segmentation as described in Section \ref{sec:context}.

\begin{figure*}
    \centering
	\begin{subfigure}{0.25\linewidth}
    	\begin{subfigure}{\linewidth}
    		\includegraphics[height=1.1cm,width=2cm]{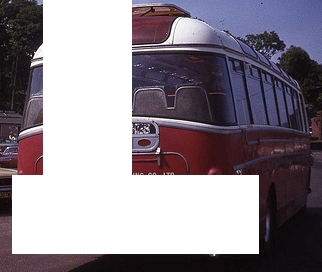}
    		\includegraphics[height=1.1cm,width=2cm]{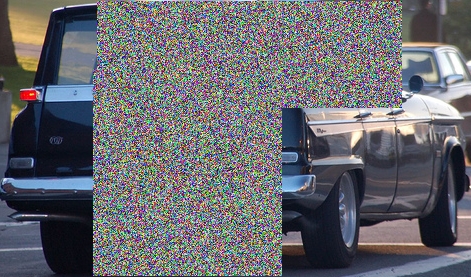}
    	\end{subfigure}%
    	\vspace{.1cm}
    	\\
    	\begin{subfigure}{\linewidth}
    		\includegraphics[height=1.1cm,width=2cm]{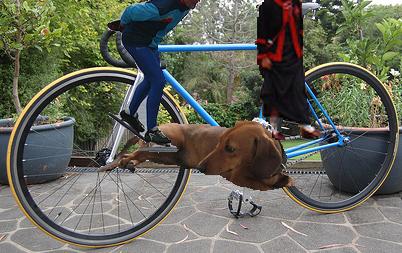}
    		\includegraphics[height=1.1cm,width=2cm]{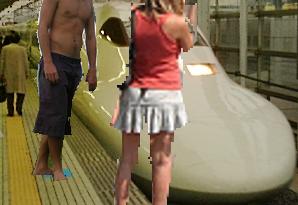}
    	\end{subfigure}%
    	\vspace{.1cm}
    	\\
    	\begin{subfigure}{\linewidth}
    		\includegraphics[height=1.1cm,width=2cm]{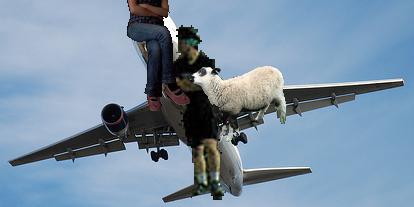}
    		\includegraphics[height=1.1cm,width=2cm]{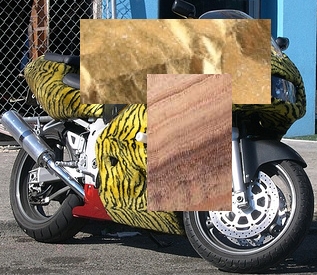}
    	\end{subfigure}%
    	\caption{}
    	\label{fig:occ-synth-class}	
	\end{subfigure}%
	\begin{subfigure}{0.25\linewidth}
    	\begin{subfigure}{\linewidth}
    		\includegraphics[height=1.1cm,width=2cm]{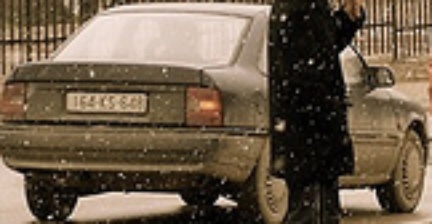}
    		\includegraphics[height=1.1cm,width=2cm]{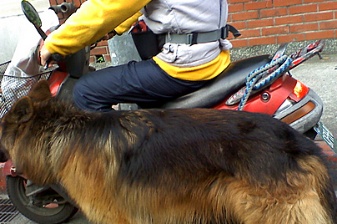}
    	\end{subfigure}%
    	\vspace{.1cm}
    	\\\\\
    	\vspace{.1cm}
    	\begin{subfigure}{\linewidth}
    		\includegraphics[height=1.1cm,width=2cm]{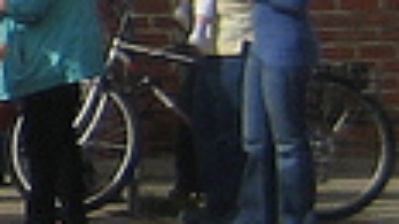}
    		\includegraphics[height=1.1cm,width=2cm]{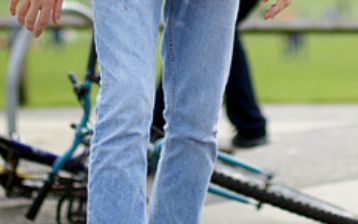}
    	\end{subfigure}%
    	\\\\\
    	\begin{subfigure}{\linewidth}
    		\includegraphics[height=1.1cm,width=2cm]{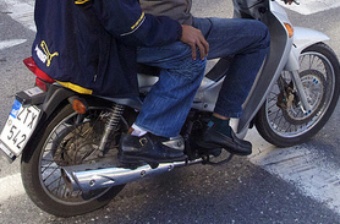}
    		\includegraphics[height=1.1cm,width=2cm]{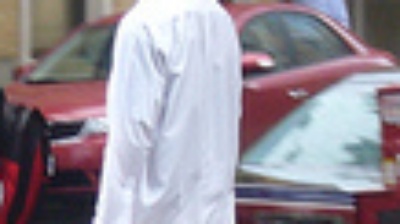}
    	\end{subfigure}%
    	\caption{}
    	\label{fig:occ-real-class}
	\end{subfigure}%
	\begin{subfigure}{0.25\linewidth}
		\centering
    	\begin{subfigure}{0.49\linewidth}
    		\includegraphics[height=1.1cm,width=2cm]{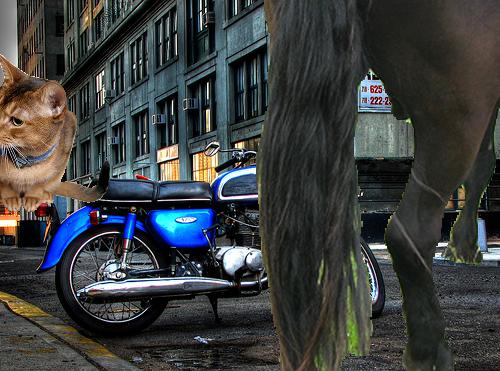}
    	\end{subfigure}%
    	\begin{subfigure}{0.49\linewidth}
    		\includegraphics[height=1.1cm,width=2cm]{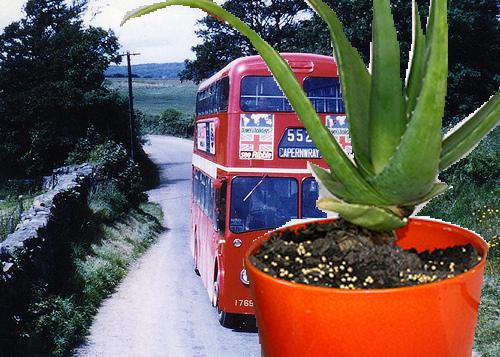}
    	\end{subfigure}%
    	\vspace{.1cm}
    	\\
    	\begin{subfigure}{0.49\linewidth}
    		\includegraphics[height=1.1cm,width=2cm]{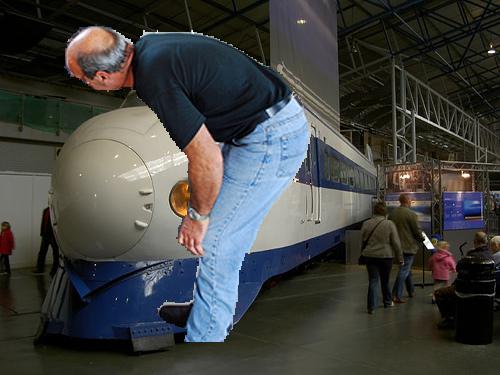}
    	\end{subfigure}%
    	\begin{subfigure}{0.49\linewidth}
    		\includegraphics[height=1.1cm,width=2cm]{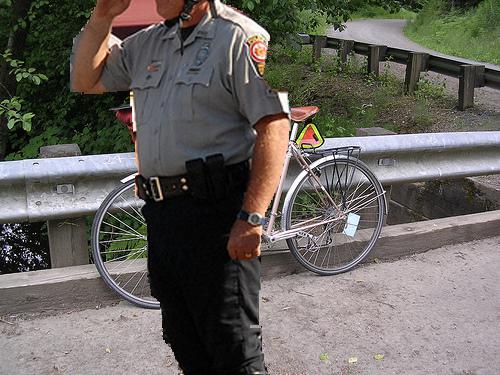}
    	\end{subfigure}%
    	\vspace{.1cm}    	
    	\\
    	\begin{subfigure}{0.49\linewidth}
    		\includegraphics[height=1.1cm,width=2cm]{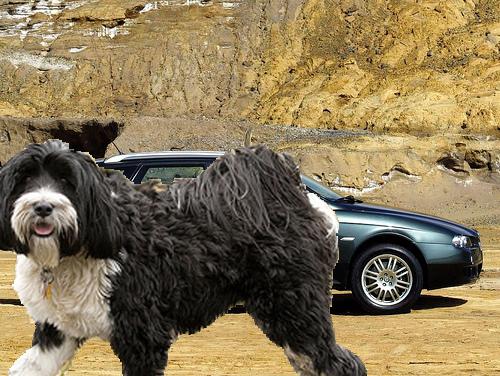}
    	\end{subfigure}%
    	\begin{subfigure}{0.49\linewidth}
    		\includegraphics[height=1.1cm,width=2cm]{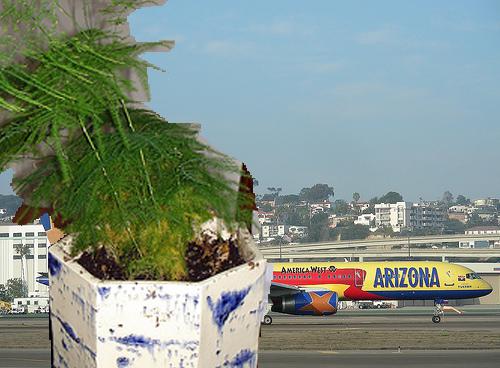}
    	\end{subfigure}%
		\caption{}
		\label{fig:occ-synth-detect}
	\end{subfigure}%
	\begin{subfigure}{0.25\linewidth}
		\centering
    	\begin{subfigure}{0.49\linewidth}
    		\includegraphics[height=1.1cm,width=2cm]{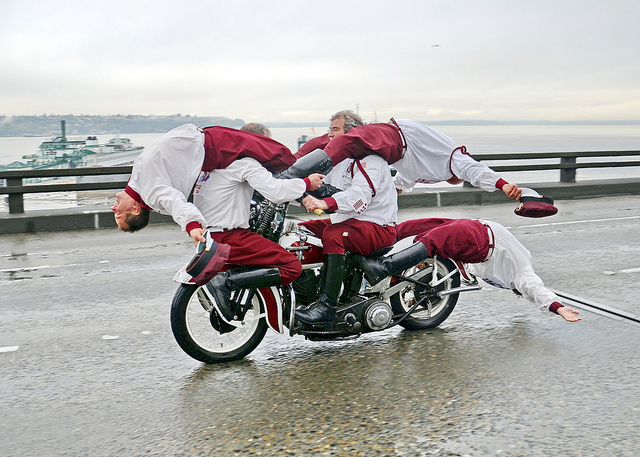}
    	\end{subfigure}%
    	\begin{subfigure}{0.49\linewidth}
    		\includegraphics[height=1.1cm,width=2cm]{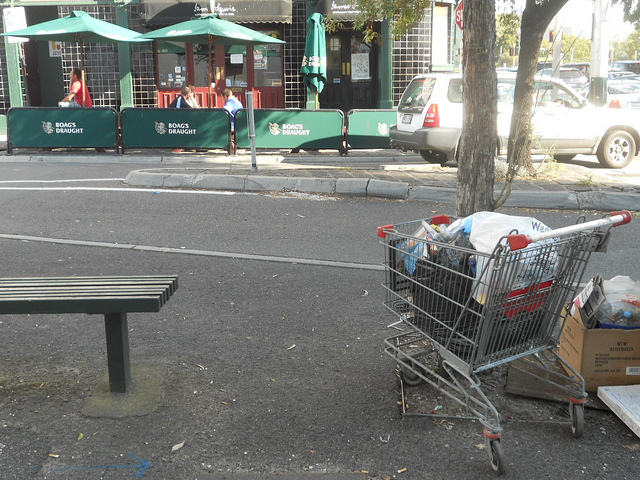}
    	\end{subfigure}%
    	\vspace{.1cm}
    	\\
    	\begin{subfigure}{0.49\linewidth}
    		\includegraphics[height=1.1cm,width=2cm]{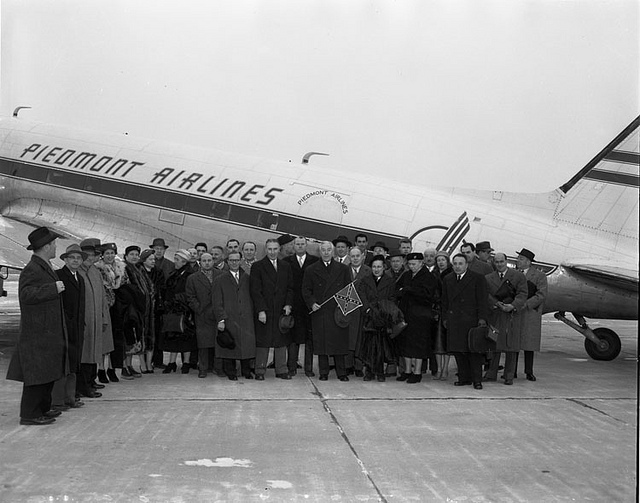}
    	\end{subfigure}%
    	\begin{subfigure}{0.49\linewidth}
    		\includegraphics[height=1.1cm,width=2cm]{img/datasets/detect_real/000000123456.jpg}
    	\end{subfigure}%
    	\vspace{.1cm}    	
    	\\
    	\begin{subfigure}{0.49\linewidth}
    		\includegraphics[height=1.1cm,width=2cm]{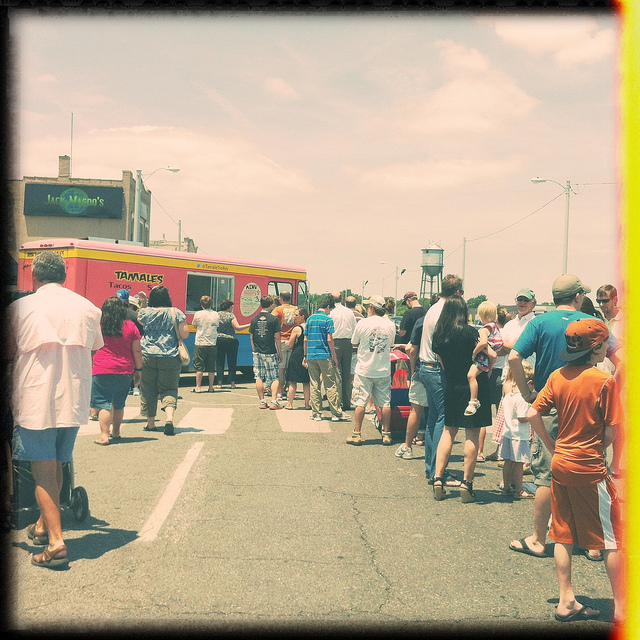}
    	\end{subfigure}%
    	\begin{subfigure}{0.49\linewidth}
    		\includegraphics[height=1.1cm,width=2cm]{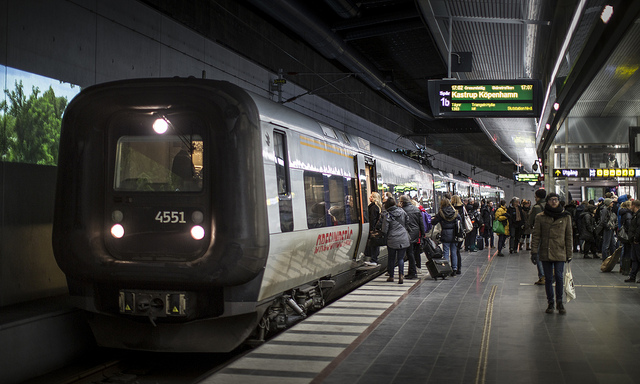}
    	\end{subfigure}%
		\caption{}
		\label{fig:occ-real-detect}
	\end{subfigure}%
    \caption{Example images of partially occluded objects for image classification (a \& b) and object detection (c \& d), with artificial occlusion (a \& c) and real occlusion (b \& d).}
    \label{fig:datasets}
\end{figure*}

\textbf{Localization and Bounding Box Estimation.} 
We denote the normalized response map of the ground truth class as 
$\hat{S}^{c}\in\mathbb{R}^{H\times W}$
and the ground truth annotation as $\bar{S}^c\in\mathbb{R}^{H\times W}$. 
The elements of the response map are computed as:
\begin{equation}
\hat{s}_ {i}^c = \frac{s_ {i, \hat{m}}}{\sum_ {i}s_ {i, \hat{m}}}, \hat{m}=\argmax_m \max_ {i}
p(f_ {i}|\mathcal{A}^m_ {i,y},\chi^m_ {i,y},\Lambda).
\end{equation}
The ground truth map $\bar{S}^c$ is a binary map where the ground truth position $\bar {i}$ is set to $\bar{S}^c(\bar {i})=1$ and all other entries are set to zero. 
The detection loss is then defined as: 
\begin{equation}
\mathcal{L}_{detect}(\hat{S}^c,\bar{S}^c,F,T^c) = 1 -\frac{2\cdot \Sigma_ {i} (\hat{s}^c_ {i} \cdot \bar{s}^c_i)}{\sum_ {i} \hat{s}^c_i + \sum_ {i}  \bar{s}^c_ {i}}.
\end{equation}

%% file: 04_0_exp_class.tex
\section{Experiments}
\label{sec:exp}
We give an overview of the datasets used for evaluation and the training setup in Section \ref{sec:exp-data}. Subsequently, we extensively evaluate our model at image classification and object detection of partially occluded objects in Sections \ref{sec:exp_class} \& \ref{sec:exp_detect}. Finally, we show that CompositionalNets are human interpretable, as their predictions can be understood in terms of occluder localization (Section \ref{sec:exp_occ}), object part detection (Section \ref{sec:exp_dissect}) and object pose estimation (Section \ref{sec:exp_view}).

\subsection{Datasets and Training Setup}
\label{sec:exp-data}
\textit{Datasets for image classification.} We evaluate our model on the \textit{Occluded-P3D+-Vehicles} dataset as proposed in \cite{wang2017detecting} and extended in \cite{kortylewski2019compositional}. The dataset is based on images of vehicles from the PASCAL3D+ dataset \cite{xiang2014beyond} that were synthetically occluded with four different types of occluders: segmented \textit{objects} as well as patches with \textit{constant white color}, \textit{random noise} and \textit{textures} (see Figure \ref{fig:occ-synth-class} for examples). 
The amount of partial occlusion of the object varies in four different levels: $0\%$ (L0), $20$-$40\%$ (L1), $40$-$60\%$ (L2), $60$-$80\%$ (L3).

Furthermore, we introduce a dataset with images of real occlusions which we term \textit{Occluded-COCO-Vehicles}. It contains the same classes as the Occluded-P3D+-Vehicles dataset. 
The dataset was generated by categorizing objects from the MS-COCO \cite{lin2014microsoft} into the four occlusion levels as defined in the Occluded-P3D+-Vehicles dataset. To achieve this, we relate the amount of object that is visible in the MS-COCO images to the one from the Occluded-P3D+-Vehicles based on the segmentation masks that are available in both datasets.
The numbers of test images for each occlusion level are: $2036$ (L0), $768$ (L1), $306$ (L2), $73$ (L3). For training purpose, we define a separate training dataset of $2036$ images from level L0. 
Figure \ref{fig:occ-real-class} illustrates some example images from this dataset.

While the current generation of CompositionalNets has been primarily developed for recognizing a smaller number of vehicle-type objects, we also want to study if CompositionalNets retain their robustness to partial occlusion even when tested at a larger scale with non-vehicle type objects. 
For this, we introduce the \textit{Occluded-ImageNet} dataset. 
In particular, we randomly select different numbers of classes $\{25,50,100\}$ from the ImageNet dataset \cite{deng2009imagenet} to study the influence of the dataset scale on the performance. 
We crop the objects from the training images with available annotation to make our results comparable to the Occluded-P3D+-Vehicles and Occluded-COCO-Vehicles data. 
We randomly split the data into $50$ test images per class and assign the remaining images to the training data. 
This results in $\{12241, 25120, 49263\}$ training and $\{1250,2500,$ $5000\}$ test images for the different sized subsets.
We artificially occlude the test images by using segmented objects from the MS-COCO dataset.
Note that to simulate occlusion, we only use those object classes from MS-COCO that do not overlap with the ImageNet classes.
The amount of partial occlusion of the object varies in four different levels (L0, L1, L2, L3) as in the Occluded-P3D+-Vehicles dataset.

\begin{table*}
	\small
	\centering
	\tabcolsep=0.11cm
	\begin{tabular}{V{2.5}lV{2.5}cV{2.5}c|c|c|cV{2.5}c|c|c|cV{2.5}c|c|c|cV{2.5}cV{2.5}}
		\multicolumn{15}{c}{\textbf{PASCAL3D+ Vehicles Classification under Occlusion}} \\
		\hline
		Occ. Area 					         & \textbf{L0: 0\%} & \multicolumn{4}{cV{2.5}}{\textbf{L1: 20-40\%}} & \multicolumn{4}{cV{2.5}}{\textbf{L2: 40-60\%}} & \multicolumn{4}{cV{2.5}}{\textbf{L3: 60-80\%}} & Mean \\
		\hline
		Occ. Type 	& - & w & n & t & o & w & n & t& o & w & n & t & o & \\    		
		\hline  
		VGG 
		& 99.2&96.9&97.0&96.5&93.8&92.0&90.3&89.9&79.6&67.9&62.1&59.5&62.2&83.6\\
		Resnet50&99.1&96.8&96.8&96.8&91.0&91.6&91.5&91.8&73.4&66.1&69.2&71.4&58.3&84.1\\
		ResNext&\textbf{99.7}&98.7&98.5&98.4&94.5&94.5&93.4&92.3&76.7&68.0&62.8&51.2&55.9&83.4\\
		CoD\cite{kortylewski2019compositional}	&92.1&92.7&92.3&91.7&92.3&87.4&89.5&88.7&90.6&70.2&80.3&76.9&87.1&87.1\\
		TDAPNet \cite{xiao2019tdapnet}
		&99.3&98.4&98.9&98.5&97.4&96.1&97.5&96.6&91.6&82.1&88.1&82.7&79.8&92.8\\		
		\hline	
		CompNet-VGG-p4&97.4&96.7&96.0&95.9&95.5&95.8&94.3&93.8&92.5&86.3&84.4&82.1&\textbf{88.1}&92.2\\
		CompNet-VGG-p5&99.3&98.4&98.6&98.4&96.9&\textbf{98.2}&\textbf{98.3}&97.3&88.1&90.1&\textbf{89.1}&83.0&72.8&93.0\\
		CompNet-Res50-RB4&99.3&98.7&98.7&98.5&96.9&96.9&97.2&96.9&88.7&86.1&83.6&79.4&72.9&91.8\\
		CompNet-RXT-RB3&98.1&95.9&96.3&96.3&94.6&92.1&93.5&92.7&87.4&76.1&80.1&75.2&76.4&88.8\\
		CompNet-RXT-RB4&\textbf{99.7}&\textbf{99.3}&\textbf{99.0}&\textbf{99.2}&\textbf{98.0}&\textbf{98.2}&97.0&\textbf{97.5}&\textbf{93.0}&\textbf{90.3}&83.8&\textbf{84.7}&83.3&\textbf{94.1}\\
	    \hline
	\end{tabular}
	\caption{Classification results for vehicles of PASCAL3D+ with different levels of artificial occlusion (0\%,20-40\%,40-60\%,60-80\% of the object are occluded) and different types of occlusion (w=white boxes, n=noise boxes, t=textured boxes, o=natural objects). CompositionalNets learned from several DCNN backbones (VGG-16, ResNet50,ResNext) outperform related approaches and their non-compositional versions significantly.}
	\label{tab:p3d_class}
\end{table*}

\textit{Datasets for Object Detection.} The object detection datasets are defined in a similar way as the classification data. 
We synthesize an \textit{Occluded-P3D+-Vehicles-Detection} dataset, which contains vehicles at a certain scale and various levels of occlusion. 
The occluders, which include humans, animals and plants, are cropped from the MS-COCO dataset \cite{lin2014microsoft}.
In an effort to accurately depict real-world occlusions, we superimpose the occluders onto the object, such that the occluders are placed not only inside the bounding box of the objects but also on the background. 
We generate the dataset in a total of 9 occlusion levels along two occlusion dimensions: We define three levels of occlusion of the object (FG-L1: 20-40\%, FG-L2:40-60\% and FG-L3:60-80\% of the object area is occluded). Furthermore, we define three levels of occlusion of the context around the object (BG-L1: 0-20\%, BG-L2:20-40\% and BG-L3:40-60\% of the context area is occluded). Example images are shown in Figure \ref{fig:occ-synth-detect}.
In order to evaluate the tested models on real-world occlusions, we test them on the subset of the MS-COCO dataset as defined for classification (Figure \ref{fig:occ-real-detect}).

\textit{Model initialization.} We initialize the vMF kernels $\{\mu_k\}$ and the mixture components $\{\mathcal{A}_y,\chi_y\}$ by maximum likelihood estimation after clustering the training data. 
We compute the mixture assignments using spectral clustering with the hamming distance between vMF kernel activations in the pool4 layer in VGG-16 of all training images. 
The intuition is that objects with a similar viewpoint and 3D structure will have similar vMF activation patterns, and thus should be assigned to the same mixture component. 

\begin{table*}
    \small  
    \centering
	\tabcolsep=0.11cm
	\begin{tabular}{V{2.5}lV{2.5}c|c|c|c|cV{2.5}c|c|c|c|cV{2.5}c|c|c|c|cV{2.5}}
		\multicolumn{16}{c}{\textbf{MS-COCO Vehicles Classification under Occlusion}} \\
		\hline Train Data & \multicolumn{5}{cV{2.5}}{\textbf{PASCAL3D+}}
		                      & \multicolumn{5}{cV{2.5}}{\textbf{MS-COCO}}
		                      & \multicolumn{5}{cV{2.5}}{\textbf{MS-COCO + CutPaste}}\\
		\hline Occ. Area & \textbf{L0} & \textbf{L1} & \textbf{L2}& \textbf{L3}& Avg
		                 & \textbf{L0} & \textbf{L1} & \textbf{L2}& \textbf{L3}& Avg
		                 & \textbf{L0} & \textbf{L1} & \textbf{L2}& \textbf{L3}& Avg\\
		\hline
		VGG &97.8&86.8&79.1&60.3&81.0
		    &99.1&88.7&78.8&63.0&82.4
		    &99.3&92.3&89.9&80.8&90.6\\
		ResNet50        &98.5&89.6&84.9&71.2&86.1
		                &\textbf{99.6}&92.7&86.9&67.1&86.6
		                &\textbf{99.6}&94.1&92.5&84.9&92.8\\
		ResNext         &98.7&90.7&85.9&75.3&87.7
		                &99.5&92.9&87.6&73.9&88.5
		                &\textbf{99.6}&94.5&93.1&89.0&94.1\\
		CoD &91.8&82.7&83.3&76.7&83.6&-&-&-&-&-&-&-&-&-&-\\
		TDAPNet &98.0&88.5&85.0&74.0&86.4
		        &99.4& 88.8& 87.9& 69.9&86.5
		        &98.1 & 89.2 & 90.5 & 79.5&89.3\\
		\hline
		CompNet-VGG-p4 &96.6&91.8&85.6&76.7&87.7
		        &97.7&92.2&86.6&82.2&89.7
		        &98.3&93.8&88.6&84.9&91.4\\
		CompNet-VGG-p5 &98.2&89.1&84.3&78.1&87.5
		        &99.1&92.5&87.3&82.2&90.3
		        &99.4&93.9&90.6&90.4&93.5\\
		CompNet-Res50-RB4      &98.5&92.6&88.9&83.6&90.9
		                &99.2&\textbf{95.2}&91.8&89.0&93.8
		                &99.0&\textbf{95.2}&93.4&89.0&94.2\\
		CompNet-RXT-RB3&97.5&91.7&82.3&73.2&86.2&98.2&93.1&84.1&83.6&89.8&98.7&93.8&87.3&84.9&91.2\\
		CompNet-RXT-RB4    &\textbf{98.8}&\textbf{94.0}&\textbf{93.5}&\textbf{87.7}&\textbf{93.5}
		                &99.0&94.8&\textbf{93.5}&\textbf{91.8}&\textbf{94.8}
		                &99.0&95.0&\textbf{94.1}&\textbf{91.8}&\textbf{95.0}\\
		\hline
	\end{tabular}
	\caption{Classification results for vehicles of MS-COCO with different levels of real occlusion (L0: 0\%,L1: 20-40\%,L2 40-60\%, L3:60-80\% of the object are occluded). 
	The training data consists of images from: PASCAL3D+, MS-COCO as well as data from MS-COCO that was augmented with data augmentation in terms of partial occlusion.
	On average, CompositionalNets outperform their non-compositional versions and related work in all test cases.}
	\label{tab:coco-class}	
\end{table*}

\textit{Training setup for classification.}  We train CompositionalNets from the feature activations of the layer before the classifier for several popular DCNNs: VGG-16 \cite{simonyan2014very}, ResNet50 \cite{he2016deep} and ResNext \cite{xie2017aggregated}. All models were pretrained on ImageNet\cite{deng2009imagenet}. We set the vMF variance to $\sigma_k = 30, \forall k \in \{1,\dots,K\}$. 
We train the parameters of the generative model $\{\{\mu_k\},\{\mathcal{A}_{y},\chi_y\}\}$ using backpropagation and keep the parameters of the backbone $\Omega$ fixed (training $\Omega$ did not have significant effects on the results). 
We learn the parameters of $n=5$ occluder models $\{\beta_1,\dots,\beta_n\}$ in an unsupervised manner as described in Section \ref{sec:vmf} and keep them fixed throughout the experiments. We set the number of mixture components to $M=4$. 
The mixing weights of the loss are set to: $\gamma_1=3.0$, $\gamma_2=3.0$. We train for $50$ epochs using stochastic gradient descent with momentum $r=0.9$ and a learning rate of $lr=0.01$. 
Our reported parameter settings are very general as they are fixed in all our experiments even for different datasets.

\textit{Training setup for object detection.} We optimize the loss described in Equation \ref{eq:loss}, with $\epsilon_1=0.2$ and $\epsilon_2=0.4$. We apply the Adam Optimizer \cite{kingma2014adam} with different learning rates on different parts of CompositionalNets: $lr_{vc}=2 \cdot 10^{-5}$, $lr_{mixture \ model} = lr_{corner \ model} = 5 \cdot 10^{-5}$. The model is trained for a total of 2 epochs with 10600 iteration per epoch. The training takes three hours in total on a machine with 4 NVIDIA TITAN Xp GPUs.

\subsection{Image Classification under Occlusion}
\label{sec:exp_class}
\textit{Occluded-P3D+.} Table \ref{tab:p3d_class} reports the results of VGG-16, ResNet50 and ResNext that were fine-tuned with the respective training data. 
Furthermore, we show the results of dictionary-based compositional models (CoD) \cite{kortylewski2019compositional} and TDAPNet \cite{xiao2019tdapnet}.
We also report CompositionalNets learned from the pool4 and pool5 layer of the VGG-16 network respectively (CompNet-VGG-[p4 \& p5]) and CompositionalNets learned from features of the last residual block (RB4) of ResNet50 (CompNet-Res50) and both the last (RB4) and second last (RB3) residual block of ResNext (CompNet-RXT).
In this setup, all models are trained with non-occluded images ($L0$), while at test time the models are exposed to images with different amount of partial occlusion ($L0$-$L3$). 

Overall, we observe that \textbf{DCNNs do not generalize well to out-of-distribution examples in terms of partial occlusion}. They perform well on the type of data that they were exposed to during training (L0) and generalize to a limited extent away from it (L1). However, their performance collapses when objects are strongly occluded (L3).

In contrast, we observe that CompositionalNets generalize very well even under strong occlusion.
Moreover, using the higher layers of more powerful residual backbones also significantly increases the performance of CompositionalNets.
We also observe that CompNets learned from the RB3 layer of ResNext  perform significantly worse, compared to those learned from RB4 ($-5.3\%$ on average). In contrast, CompNets learned from pool4 of VGG still give comparable results pool5. We conjecture from this observation that the additional convolutional and residual layers in RB4 are of critical importance for the ResNext backbone. 

\begin{table*}
    \small  
    \centering
	\tabcolsep=0.11cm
	\begin{tabular}{V{2.5}lV{2.5}c|c|c|c|cV{2.5}c|c|c|c|cV{2.5}c|c|c|c|cV{2.5}}
		\multicolumn{16}{c}{\textbf{ImageNet Classification under Occlusion}} \\
		\hline Number of classes & \multicolumn{5}{cV{2.5}}{\textbf{25}}
		                      & \multicolumn{5}{cV{2.5}}{\textbf{50}}
		                      & \multicolumn{5}{cV{2.5}}{\textbf{100}}\\
		\hline Occ. Area & \textbf{L0} & \textbf{L1} & \textbf{L2}& \textbf{L3}& Avg
		                 & \textbf{L0} & \textbf{L1} & \textbf{L2}& \textbf{L3}& Avg
		                 & \textbf{L0} & \textbf{L1} & \textbf{L2}& \textbf{L3}& Avg\\
		\hline
        ResNext + CutOut &\textbf{98.6}&69.2&40.8&20.7&57.3&\textbf{97.7}&58.6&28.4&13.5&49.6&\textbf{97.1}&53.0&25.5&10.5&46.5\\
        CompNet-ResNext-RB4-512&98.1&\textbf{70.6}&\textbf{46.9}&\textbf{31.3}&\textbf{61.7}&95.1&60.7&36.5&19.8&53.0&92.9&53.7&30.1&14.7&47.9\\
        CompNet-ResNext-RB4-1024&97.2&69.8&45.9&29.1&60.5&95.4&\textbf{61.3}&\textbf{37.0}&\textbf{21.2}&\textbf{53.7}&94.2&\textbf{56.1}&\textbf{32.6}&\textbf{16.0}&\textbf{49.7}\\
		\hline
	\end{tabular}
	\caption{Classification results for different number of classes from ImageNet and different levels of artificial occlusion (L0: 0\%,L1: 20-40\%,L2 40-60\%, L3:60-80\% of the object are occluded). 
	We compare a standard ResNext model trained with strong data augmentation in terms of CutOut with CompositionalNets that are learned from a ResNext backbone with 512 and 1024 vMF kernels.
	CompositionalNets are more robust under strong occlusion in all experiments and stay competitive in low occlusion scenarios. 
	}
	\label{tab:imagenet-class}	
\end{table*}

\textit{Occluded MS-COCO.}
Table \ref{tab:coco-class} shows classification results under a realistic occlusion scenario by testing on the Occluded-COCO-Vehicles dataset. 
The models in the first part of the Table  (PASCAL3D+) are trained solely on non-occluded images of the PASCAL3D+ data. 
We can observe that from all standard DCNNs VGG-16 transfers the worst to the MS-COCO data, while ResNet50 and ResNext generalize better, in particular to strongly occluded objects. 
ResNext even outperforms the recently proposed TDAPNet that was specifically designed for image classification under occlusion.
\textbf{CompositionalNets consistently outperform non- compositional DCNNs} by a large margin, highlighting the generality of our approach. In particular, CompNet-ResNext-RB4 can generalize very strongly from training on non-occluded PASCAL3D+ data to strongly occluded objects from MS-COCO.

The second part of the table (MS-COCO) shows the classification performance after fine-tuning on the non-occluded training set of the Occluded-COCO-Vehicles dataset. 
A common pattern of all three standard DCNNs is that their performance increases for low occlusion ($L0-2$) while it decreases for stronger occlusion ($L3$). 
In contrast, the performance of the CompositionalNets increases substantially after fine-tuning for all occlusion levels. 

The third part of Table \ref{tab:coco-class} (MS-COCO-CutPaste) shows classification results after training with strong data augmentation in terms of partial occlusion.
In particular, we artificially occlude the non-occluded training images used in the previous experiment with all four types of artificial occluders used in the Occluded-P3D+-Vehicles dataset. This data augmentation increases the dataset by a factor of $5$.
From the classification results, we can observe that data augmentation increases the performance of the DCNNs significantly. VGG-16 still suffers from strong occlusions. However, ResNet50 and particularly ResNext become significantly more robust to occlusion.
The performance of the CompositionalNets learned from VGG-16 increases further when trained with augmented data, whereas the ones learned from ResNet50 and ResNext only benefit slightly, while still outperforming their non-compositional counterparts.
Similar to the results in Table \ref{tab:p3d_class}, CompNet-ResNext-RB3 performs significantly worse compared to CompNet-ResNext-RB4, highlighting the importance of the last layers in the backbone of ResNext. 
Interestingly, CompNet-ResNext-RB3 performs on par to CompNet-ResNext-p4 under real occlusion, whereas it is less robust to the artificial occluders (Table \ref{tab:p3d_class}). 
This suggests that the ResNext features have developed an invariance to partial occlusion from the ImageNet pre-training that does not transfer as well to artificial occlusion. We discuss this in more detail in Section \ref{sec:exp_occ}.

\begin{table*}
	\small
	\centering
	\tabcolsep=0.11cm
	\begin{tabular}{V{2.5}lV{2.5}cV{2.5}c|c|c|cV{2.5}c|c|c|cV{2.5}c|c|c|cV{2.5}cV{2.5}}
		\multicolumn{15}{c}{\textbf{Ablation Study on PASCAL3D+ Vehicles Classification under Occlusion}} \\
		\hline
		Occ. Area 					         & \textbf{L0: 0\%} & \multicolumn{4}{cV{2.5}}{\textbf{L1: 20-40\%}} & \multicolumn{4}{cV{2.5}}{\textbf{L2: 40-60\%}} & \multicolumn{4}{cV{2.5}}{\textbf{L3: 60-80\%}} & Mean \\
		\hline
		Occ. Type 	& - & w & n & t & o & w & n & t& o & w & n & t & o & \\
		\hline  
		CompNet-ResNext initialization& 98.1&95.8&96.0&95.6&91.3&91.7&91.8&91.7&80.8&75.7&77.2&77.1&67.9&87.0\\
        CompNet-ResNext $\gamma_1=0,\gamma_2=0$& 99.5&98.3&97.8&98.0&95.0&94.5&92.9&92.6&81.7&71.0&59.0&59.0&58.3&84.4\\		
        CompNet-ResNext $\gamma_1=0,\gamma_2=3$& 99.6&97.7&96.5&97.4&93.6&95.0&91.4&93.8&86.8&87.0&75.3&80.0&77.6&90.1\\        
		CompNet-ResNext $\gamma_1=3,\gamma_2=0$ &\textbf{99.9}&99.2&98.8&98.9&97.5&97.5&95.6&96.5&92.8&87.8&78.4&81.9&\textbf{83.9}&93.0\\
		CompNet-ResNext $\gamma_1=3,\gamma_2=3$
		&99.7&\textbf{99.3}&\textbf{99.0}&\textbf{99.2}&\textbf{98.0}&\textbf{98.2}&\textbf{97.0}&\textbf{97.5}&\textbf{93.0}&\textbf{90.3}&\textbf{83.8}&\textbf{84.7}&83.3&\textbf{94.1}\\
		\hline
	\end{tabular}
    \caption{Ablation study for the generative regularization of a CompositionalNet learned from the features of the last residual block (RB4) of ResNext. The results highlight the importance of maximum likelihood regularization of both the vMF kernels and the mixture models.}
    \label{tab:p3d_ablation}
\end{table*}

\textit{Occluded ImageNet.} 
Table \ref{tab:imagenet-class} shows classification results for larger scale experiments with non-vehicle objects on the Occluded-ImageNet dataset.
We compare a standard ResNext model trained with strong data augmentation using CutOut \cite{devries2017improved} with CompositionalNets learned from a ResNext backbone that were trained with non-augmented images only. We also test CompositionalNets with $512$ and $1024$ vMF kernels.
We observe that CompositionalNets are more robust under strong occlusion in all experiments and stay competitive in low occlusion scenarios. 
The CompositionalNet performance decreases slightly on non-occluded images in the 100 class experiment (ResNext+CutOut: 97.1; CompNet-ResNext-1024: 94.2). However, CompositionalNets perform better under any amount of partial occlusion even for this large set of non-vehicle objects.

Furthermore, we observe that the number of vMF kernels does not have a critical influence on the performance. 
This highlights the advantage of the compositional representation which enables an efficient sharing of the vMF kernels among the different classes.
Nevertheless, we can observe that increasing the number of vMF kernels has a positive effect on the performance for higher number classes, because the model can represent the objects more accurately when more parts are available. 

Overall, we can observe that CompositionalNets are already capable to generalize to large amounts of non-vehicle classes, while retaining their robustness to partial occlusion. 
This is a very promising result considering that our model was mainly tested at the robust analysis of a smaller set of vehicle objects. 
Therefore, we expect improvements by investigating how CompositionalNets could better model articulated objects such as animals. 
 
\textit{Summary of classification experiments.} All classification experiments clearly highlight the robustness of CompositionalNets at classifying partially occluded objects, while also being highly discriminative when objects are not occluded. 
Overall, CompositionalNets learned from several popular DCNNs show a significantly improved generalization performance compared to non-compositional DCNNs under artificial as well as real occlusion.
Notably, CompNet-ResNext trained from non-occluded PASCAL3D+ data performs essentially on par with the best DCNN (ResNext) trained from strongly augmented MS-COCO images. 
This highlights the data efficiency and strong generalization ability of CompositionalNets.
Furthermore, we observe that CompositionalNets have a lot of potential for large-scale classification tasks with non-vehicle type objects.

\begin{table*}
\tabcolsep=0.11cm
\begin{tabular}{V{2.5}lV{2.5}cV{2.5}cccV{2.5}cccV{2.5}ccc V{2.5}cV{2.5}}
\multicolumn{12}{c}{\textbf{PASCAL3D+ Vehicles Detection under Occlusion}} \\
\hline
                      Occ. Area Foreground& FG L0 & \multicolumn{3}{c}{FG L1} & \multicolumn{3}{V{2.5}c}{FG L2} & \multicolumn{3}{V{2.5}cV{2.5}}{FG L3} & \multicolumn{1}{cV{2.5}}{Mean} \\
                      \hline
Occ. Area Background                & BG L0 & BG L1   & BG L2  & BG L3  & BG L1   & BG L2  & BG L3  & BG L1   & BG L2  & BG L3 & -- \\
\hline
Faster R-CNN           &{\bf 98.0}     &88.8       &85.8       &83.6       &72.9       &66.0        &60.7       &46.3      &36.1       &27.0       &66.5        \\
Faster R-CNN with reg. &97.4     &{\bf 89.5}       &86.3       &89.2       &76.7       &70.6        &67.8       &54.2      &45.0       &37.5       &71.1        \\
\hline
CompNet-VGG-p4-RPN $\omega=0.5$ & 74.2     & 68.2    & 67.6    & 67.2   & 61.4     & 60.3     & 59.6   & 46.2    & 48.0    & 46.9   & 60.0 \\
CompNet-VGG-p4-RPN $\omega=0$ & 73.1    & 67.0    & 66.3    & 66.1   & 59.4    & 60.6    & 58.6   & 47.9    & 49.9    & 46.5   & 59.6 \\
\hline

CompNet-VGG-p4 $\omega=0.5$ & 91.7	& 85.8	& 86.5	& 86.5	& 78	& 77.2	& 77.9	& 61.8	& 61.2	& 59.8	& 76.6 \\
CompNet-VGG-p4 $\omega=0.2$ & 92.6  & 87.9    & 88.5    & {\bf 88.6}   & 82.2    & {\bf 82.2}    & {\bf 81.1}   & 71.5    & {\bf 69.9}    & {\bf 68.2}   & 81.3 \\
CompNet-VGG-p4 $\omega=0$ & 94.0    & 89.2    & {\bf 89.0}    & 88.4   & {\bf 82.5}    & 81.6    & 80.7   & {\bf 72.0}    & 69.8    & 66.8   & {\bf 81.4} \\
\hline
CompNet-ResNext-RB3 $\omega=0.2$ & 94.6 & 85.5 & 85.3 & 85.4 & 76.4 & 74.7  & 74.7 & 62.4 & 60.7 & 58.0 & 75.8 \\
\hline
\end{tabular}
\caption{Detection results on the Occluded-P3D+-Vehicles-Detection dataset under different levels of occlusions. All models were trained on non-occluded images of the PASCAL3D+ dataset except Faster R-CNN with reg., which was trained with CutOut. The results are measured by correct AP(\%) @IoU0.5, which means only correct classified images with $IoU>0.5$ of the predicted bounding box are treated as true-positive. Notice with $\omega=0.5$ the context-aware model reduces to a CompositionalNet as proposed in Section \ref{sec:compnet}.}
\label{tab:p3d}
\end{table*}

\subsubsection{Ablation Study}
In Table \ref{tab:p3d_ablation} we show the results of an ablation study using a CompositionalNet learned from a ResNext backbone. 
After initialization, the CompositionalNet achieves an average performance of $87\%$ and hence already outperforms a standard ResNext architecture by $3.6\%$ on average.
Note that the CompositionalNet at this point is not as discriminative at $L0$ and performs on par at $L1$, while significantly outperforming ResNext at $L2$ and $L3$.

When the parameters of the vMF kernels and mixture components are not regularized ($\gamma_1=0, \gamma_2=0$) the overall performance decreases.
In particular, the CompositionalNet becomes more discriminative for the type of data it has observed at training time ($L0$) and cannot generalize well to stronger occlusion. 
Hence, it behaves as one would expect from a standard DCNN.

Regularizing only the mixture components to maximize the likelihood of the vMF kernel activations ($\gamma_1=0, \gamma_2=3$) increases the performance in all experiments. 
In particular, the end-to-end training makes CompositionalNets highly discriminative for the within-distribution-data ($L0$) while preserving strong generalization performance for out-of-distribution data ($L1-L3$).
Similarly, regularizing only the vMF kernels to maximize the likelihood of the ResNext features ($\gamma_1=3, \gamma_2=0$) also increases performance for all experiments significantly for non-occluded as well as strongly occluded data.
The best performance is achieved with a joint regularization of the vMF kernels and mixture components ($\gamma_1=3,\gamma_2=3$).

%% file: 04_1_exp_detect.tex
\begin{table}
\tabcolsep=0.11cm
\begin{tabular}{V{2.5}lV{2.5}cccccV{2.5}}
\multicolumn{6}{c}{\textbf{MS-COCO Vehicles Detection under Occlusion}} \\
\hline
Occ. Area & L0 & L1 & L2 & L3 & Avg\\
       \hline
Faster R-CNN                        & 77.2	&59.0 	&40.8	&24.6&50.4 \\
Faster R-CNN with reg.              & 80.7	&63.3	&45.0 	&33.3&55.6 \\
Faster R-CNN with occ.              & 82.5	&66.0	&50.7	&45.6&61.2 \\
       \hline
CompNet-VGG-p4-RPN       & 60.0 & 49.7 & 45.4 & 38.6&48.4 \\
       \hline
       
CompNet-VGG-p4 $\omega$=$0.5$     & 81.6 & 70.8 & 51.7 & 40.4&61.1 \\
CompNet-VGG-p4 $\omega$=$0.2$      & 86.8 & {\bf 77.8} & 65.4 & {\bf 59.6}&\textbf{72.4} \\
CompNet-VGG-p4 $\omega$=$0$       & {\bf 89.4} & 76.2 & 61.1 & 54.4&70.3\\
\hline
CompNet-RXT-RB3 $\omega$=$0.2$ & 85.7 & 72.5 & {\bf 65.9} & {\bf 59.6}&70.9\\
\hline
\end{tabular}
\caption{Detection results on the Occluded-COCO-Vehicles-Detection dataset, measured by AP(\%) @IoU0.5. All models are trained on non-occluded images of the PASCAL3D+ dataset, except Faster R-CNN with reg. is trained with cutout and Faster R-CNN with occ. is trained with images that were artificially occluded using segmented objects. CompNet-VGG-p4-RPN has been evaluated with $\omega=0$. Note how the CompositionalNets are significantly more robust to partial occlusion compared to Faster R-CNN.}
\label{tab:coco}
\end{table}
\begin{figure*}
    \centering
	\begin{subfigure}{0.5\linewidth}
		\includegraphics[height=4.2cm]{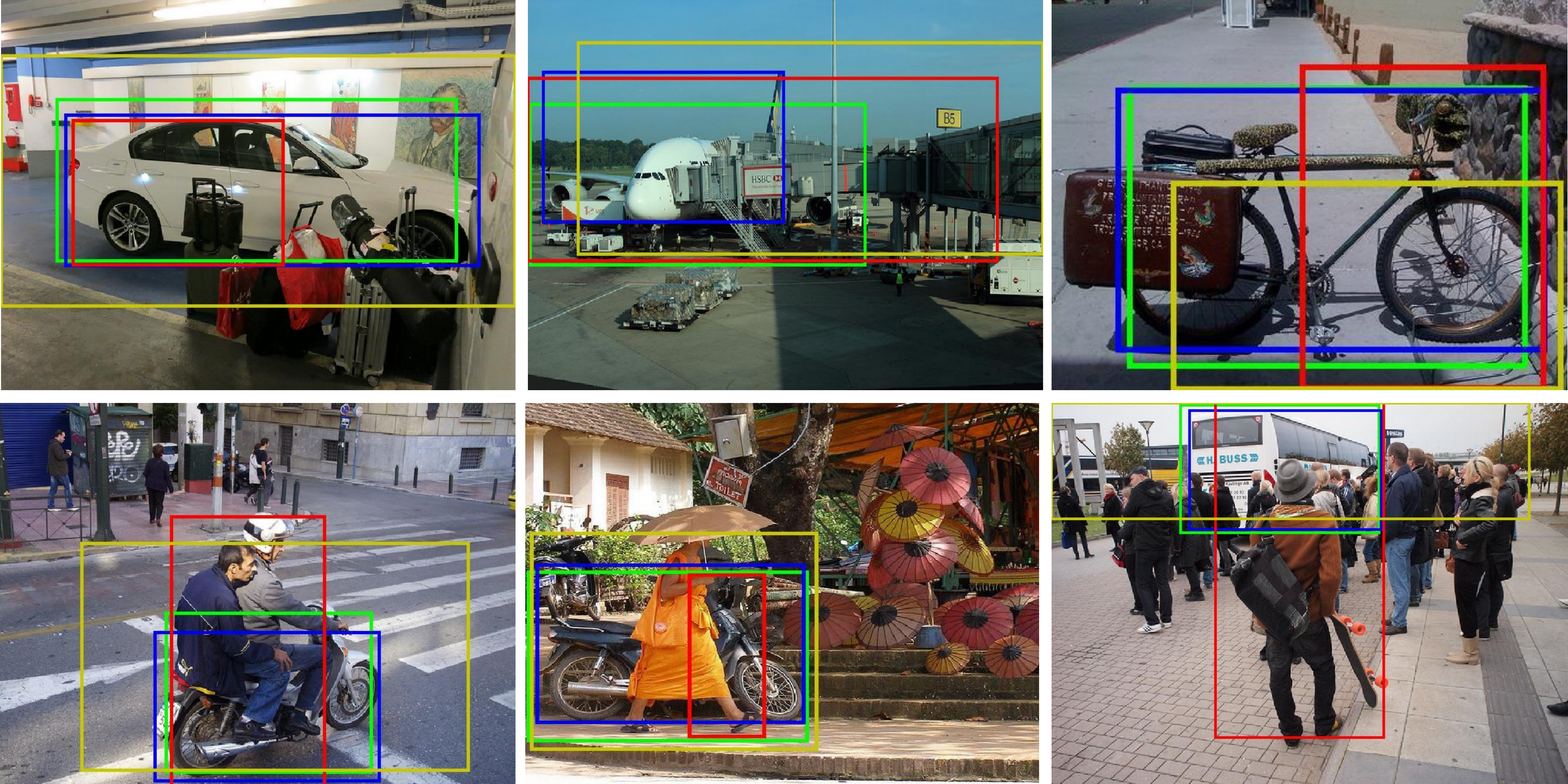}
 	\end{subfigure}%
 	\begin{subfigure}{0.5\linewidth}
		\includegraphics[height=4.2cm]{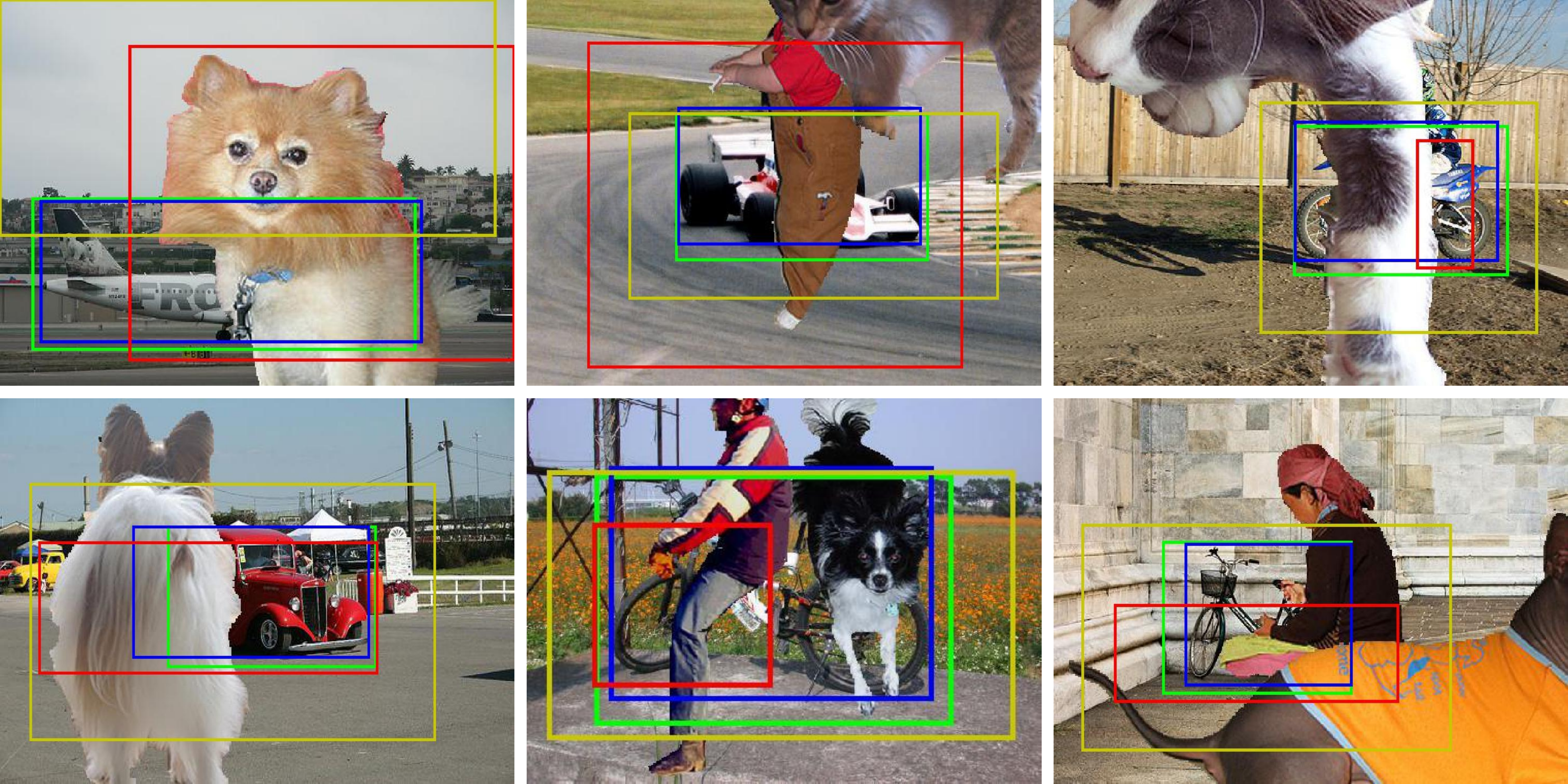}
 	\end{subfigure}%

    \caption{Qualitative detection results on images with synthetic and real occlusion. Blue box: Ground truth. Red box: Bounding box by Faster R-CNN. Yellow Box: RPN+CompositionalNet. Green box: CompositionalNet + robust bounding-box voting.}
    \label{fig:exp_detect_quali}
\end{figure*}

\subsection{Object Detection under Occlusion}
\label{sec:exp_detect}

In this Section, we evaluate CompositionalNets at the task of object detection. 
In the experiments, we use a context-aware CompNet-VGG16-pool4 and a CompNet-ResNext-RB3 because of their higher resolution compared to the other types of backbones tested in image classification. 

\textit{Occluded-P3D+.} 
In Table \ref{tab:p3d}, we report the results of Faster-RCNN and CompositionalNets on the Occluded-P3D+-Vehicles-Detection dataset.
The models are trained on the images from the original PASCAL3D+ dataset with non-occluded objects. Qualitative detection results are illustrated in Figure \ref{fig:exp_detect_quali}.

We observe that \textbf{Faster R-CNN fails to detect strongly occluded objects reliably}, while it performs well under low levels of occlusion. 
When trained with strong data augmentation in terms of partial occlusion using CutOut \cite{devries2017improved}, the detection performance increases under strong occlusion. 
However, the model still suffers from a $59.9\%$ drop in performance on strong occlusion, compared to the non-occlusion setup.
We suspect that the inaccurate prediction is because of two major factors. 1) The Region Proposal Network (RPN) in the Faster R-CNN is not able to predict accurate proposals of objects that are heavily occluded. 
2) The VGG-16 classifier cannot successfully classify valid object regions under heavy occlusion.

We proceed to investigate the performance of the region proposals on occluded images. 
We conduct this experiment by replacing the VGG-16 classifier in the Faster R-CNN with a CompositionalNet classifier that is learned from the pool4 layer of VGG, which is expected to be more robust to occlusion. 
Based on the results in Table 1, we observe two phenomena. 1) In high levels of occlusion, the performance is better than Faster R-CNN. Thus, the CompositionalNet generalizes to heavy occlusions better than the VGG-16 classifier. 2) In low levels of occlusion, the performance is worse than Faster R-CNN. 

\textit{The importance of spatial alignment in CompositionalNets.} Recall that in the classification experiments, we observed that CompositionalNets are robust to occlusion because they can roughly localize occluders and subsequently focus on the non-occluded object parts for classification. 
They can localize occluders because the different components in the mixture models explicitly represent the \textit{spatial distribution} of features of an object in a certain pose. To be successful, the localization process requires the features of the object in the image to be roughly aligned with the mixture models. 
Therefore, CompositionalNets require bounding box proposals in which the center of the object is roughly aligned to the center of the bounding box, independent of whether the object is occluded or not. 
In this sense, CompositionalNets are rather high precision models which require spatial alignment between image and model. 
This is in contrast to standard deep networks, which are observed to not use spatial information extensively and behave more like bag-of-words type models  \cite{brendel2019approximating}. 
The results in Section \ref{sec:exp_class} and Section \ref{sec:exp_occ} show that the spatial distribution of object parts in an image is very important for computer vision models, because it enables the rough localization of occluders and thus a robust classification and detection under occlusion.  
One problem with the RPN proposals is that they mostly cover the visible part of the object only and often do not align the object center to the center of the bounding box. 

\textit{Effect of robust bounding box voting.} Our approach of accurately estimating the corners of the bounding box substantially improves the performance of the CompositionalNet, in comparison with the RPN. This further validates our conclusion that the CompositionalNet classifier requires precise proposals to classify objects correctly with partial occlusions.

\textit{Effect of context-aware representation.} 
We observe that models with a reduced influence of the context ($\omega=0.2, \omega=0.0$) outperform models with equal weight on context and object representation ($\omega=0.5$). Hence, \textbf{context-aware CompositionalNets are more robust to partial occlusion than unaware models}. Furthermore, the performance of all models follows a similar trend over all three levels of foreground occlusions: the performance decreases as the level of background occlusion increases from BG-L1 to BG-L3. This further confirms our understanding of the effects of the context as a valuable source of information in object detection.

\textit{Occluded-MS-COCO.} 
As show in Table \ref{tab:coco} and Figure \ref{fig:exp_detect_quali}, the context-aware CompositionalNet with robust bounding box voting outperforms Faster R-CNN and CompNet+RPN significantly.
Furthermore, the quantitative results clearly show the benefit of the context-awareness ($\omega=0.2$) over unaware CompositionalNets ($\omega=0.5$). 
While fully deactivating the context ($\omega=0$) slightly decreases the performance, controlling the prior of the context model to $\omega=0.2$ reaches a sweet spot where the context is helpful but does not have an overwhelming influence as the in the original CompositionalNet. 
Similar as observed in the classification experiments, CompNet-RXT-RB3 performs significantly worse compared to its pool4 variant for artificial occluders (Table \ref{tab:p3d}), whereas it performs similarly under real occlusion (Table \ref{tab:coco}).
We will discuss this phenomenon in more detail in Section \ref{sec:exp_occ}.

%% file: 04_2_exp_interpret.tex
\begin{figure*}
    \centering
	\begin{subfigure}{0.33\linewidth}
		\centering
		\includegraphics[width=1.1\linewidth]{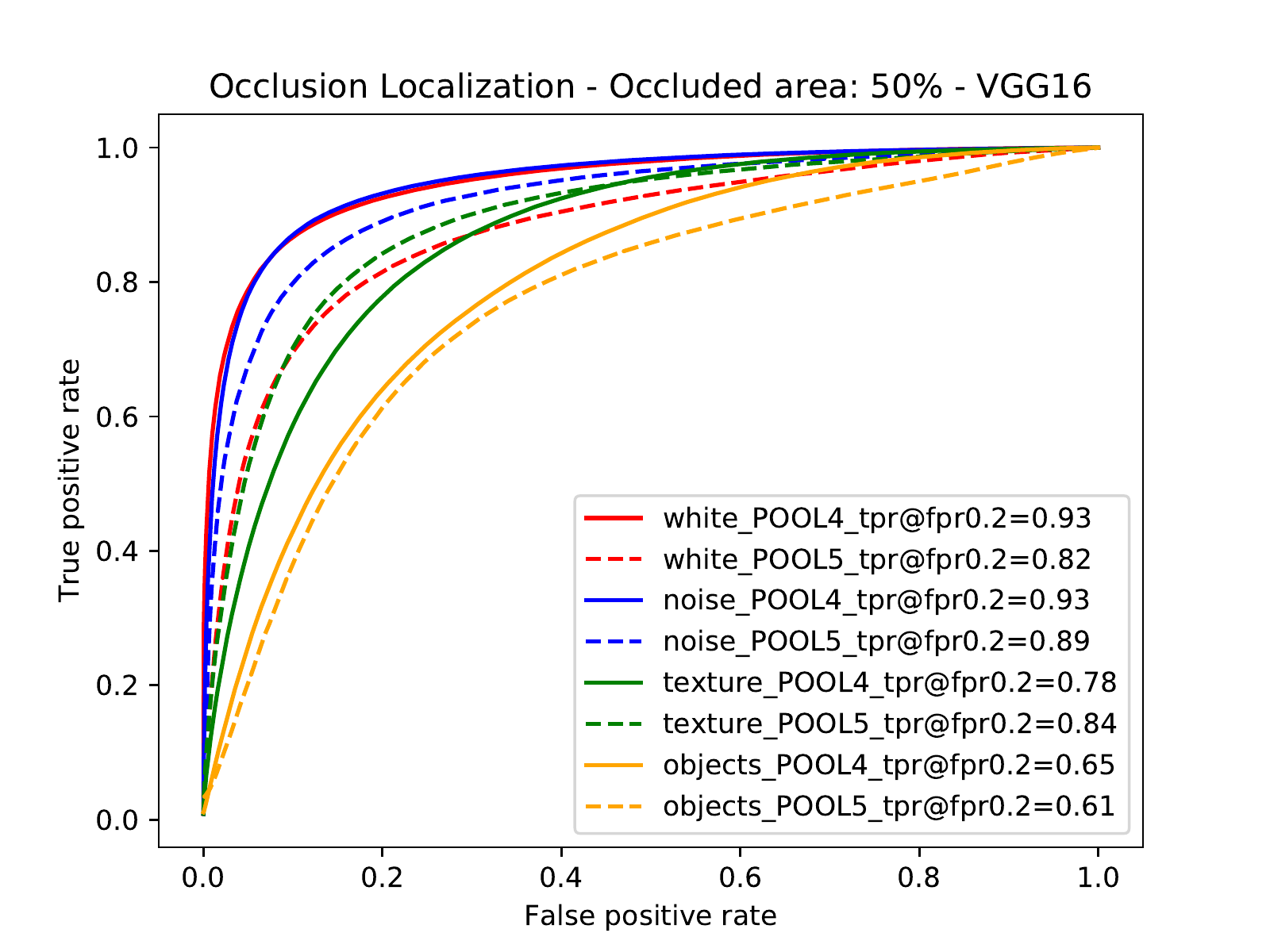}
		\caption{}
		\label{fig:roc_vgg}
	\end{subfigure}%
	\begin{subfigure}{0.33\linewidth}
		\centering
		\includegraphics[width=1.1\linewidth]{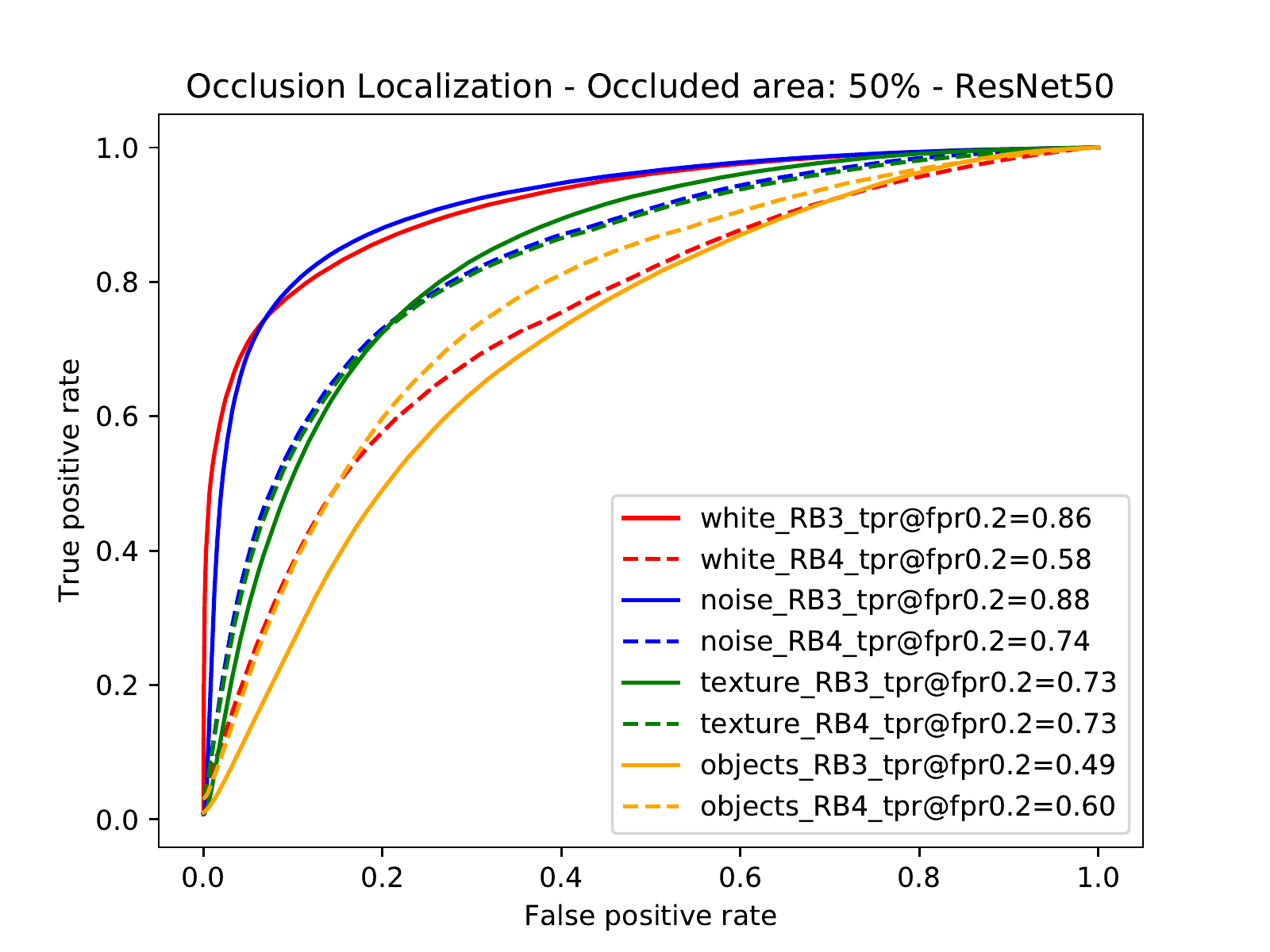}
		\caption{}
		\label{fig:roc_res50}
	\end{subfigure}%
	\begin{subfigure}{0.33\linewidth}
		\centering
		\includegraphics[width=1.1\linewidth]{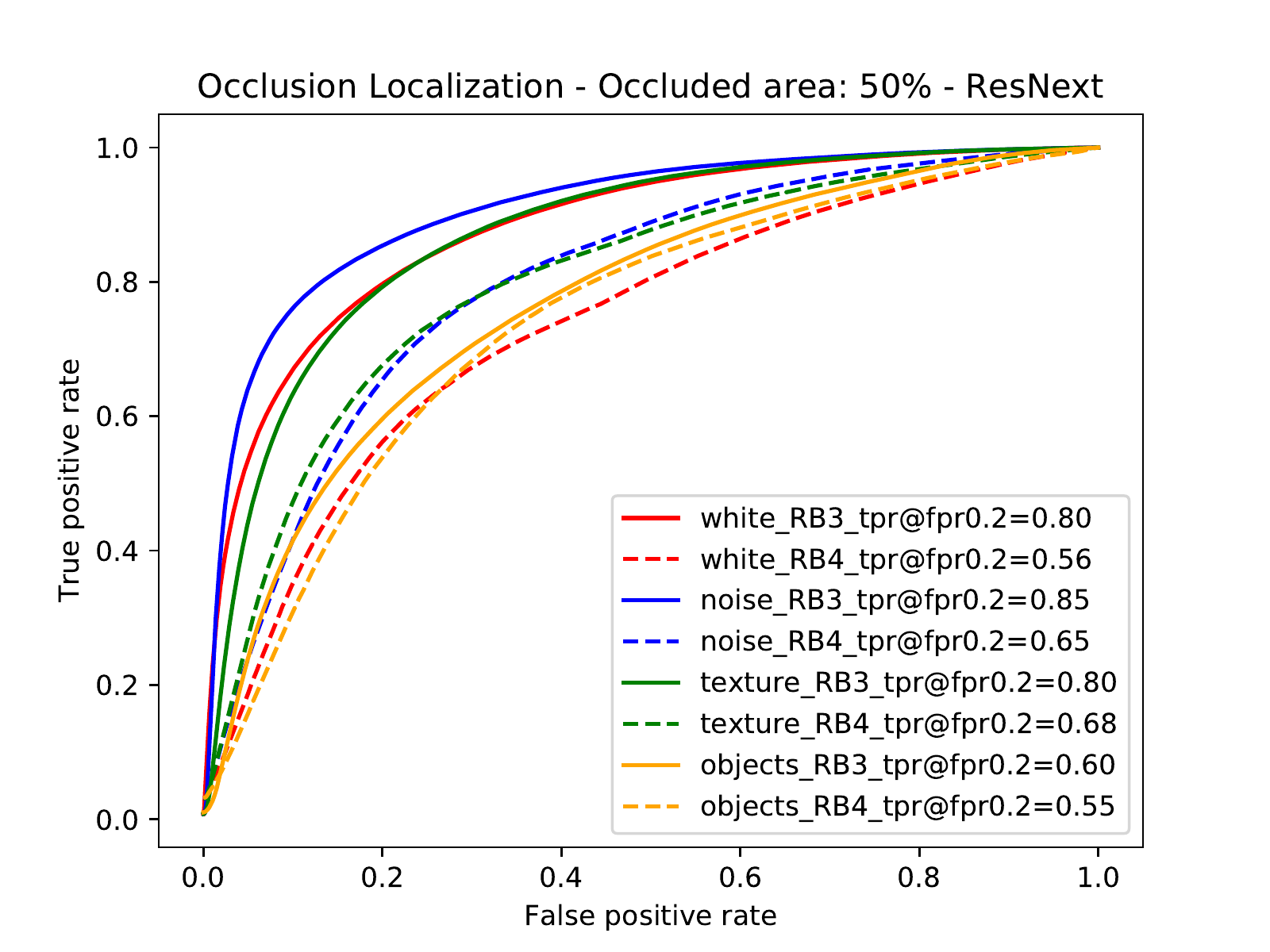}
		\caption{}
		\label{fig:roc_resnext}
	\end{subfigure}%
	\caption{ROC curves measuring occlusion localization scores in image classification with CompositionalNets learned from different DCNN backbones: (a) pool4 and pool5 layer of VGG16, and features after the residual block 3 (RB3) and residual block 4 (RB4) of (b) ResNet50 and (c) ResNext. The objects in the test data are on average $50\%$ occluded. Note how all models can localize occluders well. The CompNets learned from VGG16 significantly outperform the backbones with residual connections.}
	\label{fig:occ_loc_class}
\end{figure*}

\subsection{Model Interpretability} 
\label{sec:exp_explain}

While it is important that computer vision systems can robustly generalize to out-of-distribution examples in terms of partial occlusion, in real-world applications it is equally important that their prediction result is human interpretable. 
In this section, we show that CompositionalNets are highly interpretable models. 
We demonstrate that they can localize occluders accurately in image classification and object detection (Section \ref{sec:exp_occ}), while being trained with class-level supervision only.
Furthermore, we show that the predictions of CompositionalNets can be understood in terms of detecting object parts (Section \ref{sec:exp_dissect}) and estimating the objects' viewpoint (Section \ref{sec:exp_view}).

\subsubsection{Occluder Localization}
\label{sec:exp_occ}
A successful localization of occluders increases the robustness of a model to partial occlusion and also enables a human observer to better understand a models' underlying reasoning process.
In the following, we test the ability of CompositionalNets at occluder localization.
We compute the occlusion score as the log-likelihood ratio between the occluder model and the object model: $\log p(f_p|z^m_p\texttt{=}1)/p(f_p|z^m_p\texttt{=}0)$, where $m=\argmax_m p(F|\theta^m_y)$ is the mixture component that explains the feature activations of the DCNN backbone the best.

\begin{figure*}
    \begin{subfigure}{\linewidth}
       \centering
        \includegraphics[height=3cm]{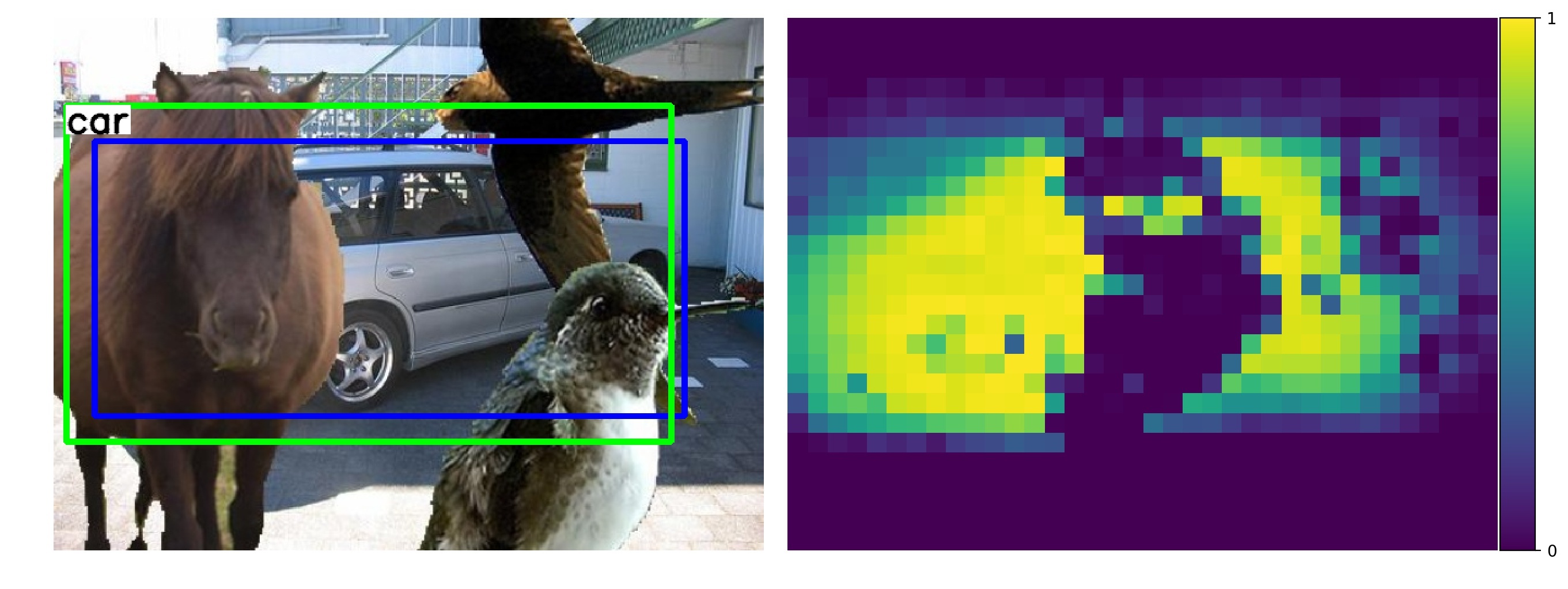}
        \includegraphics[height=3cm]{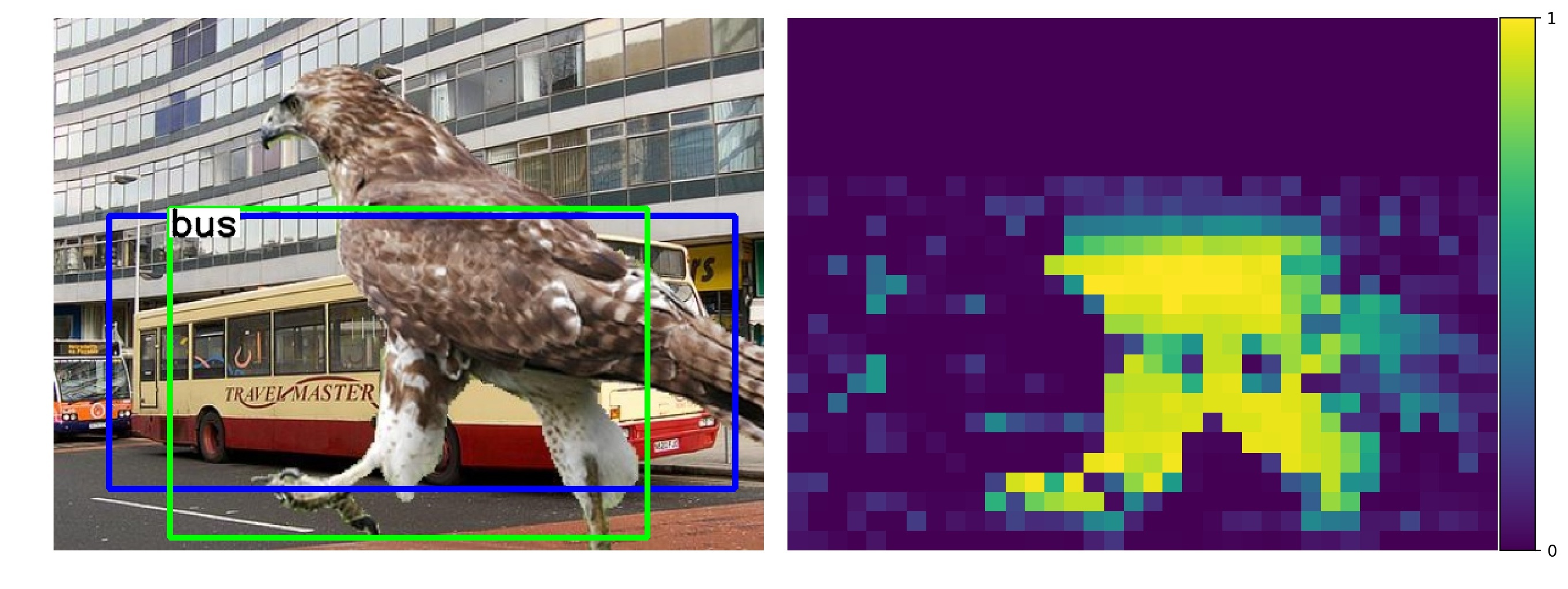}
    \end{subfigure}
    \\
    \begin{subfigure}{\linewidth}
       \centering
        \includegraphics[height=3cm]{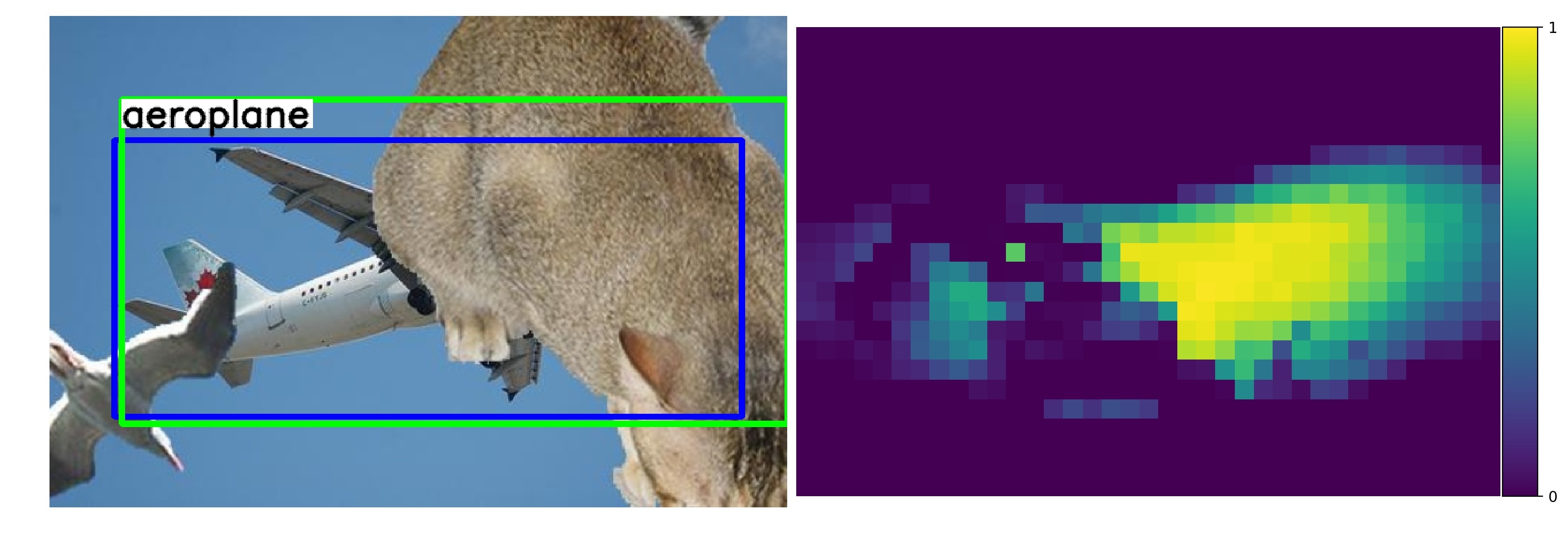}
        \includegraphics[height=3cm]{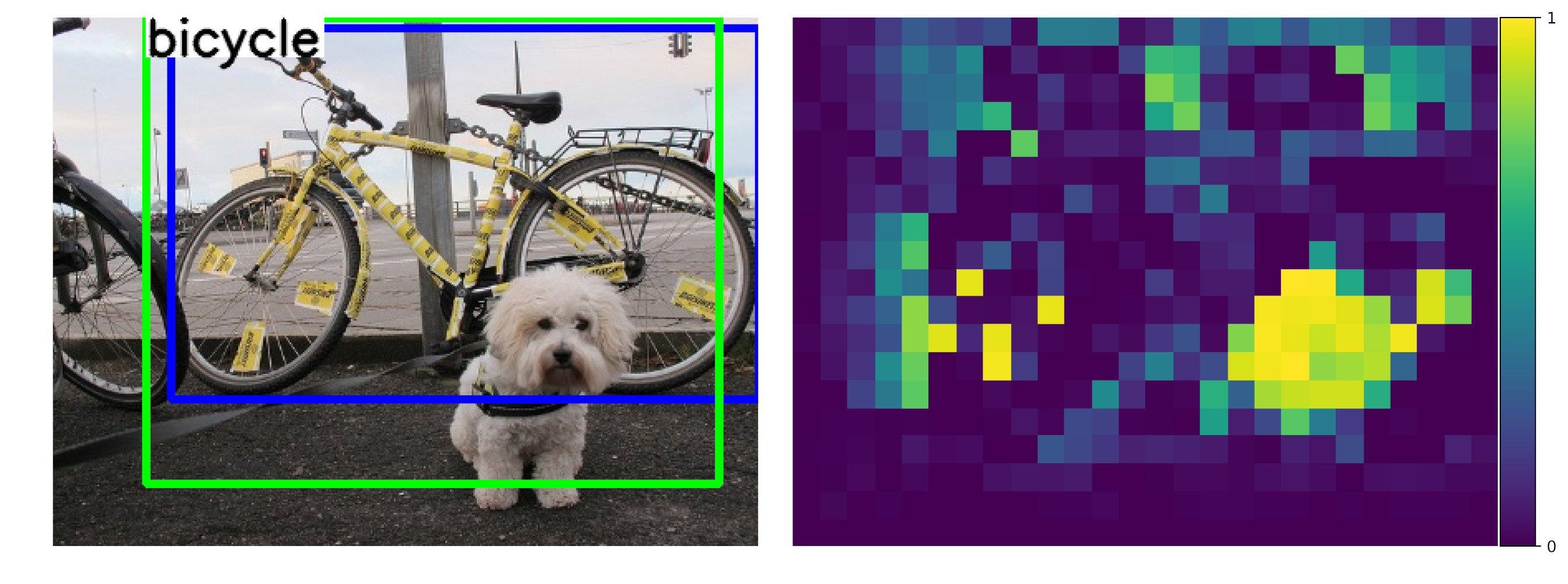}
    \end{subfigure}
    \\
    \begin{subfigure}{\linewidth}
       \centering
        \includegraphics[height=2.8cm]{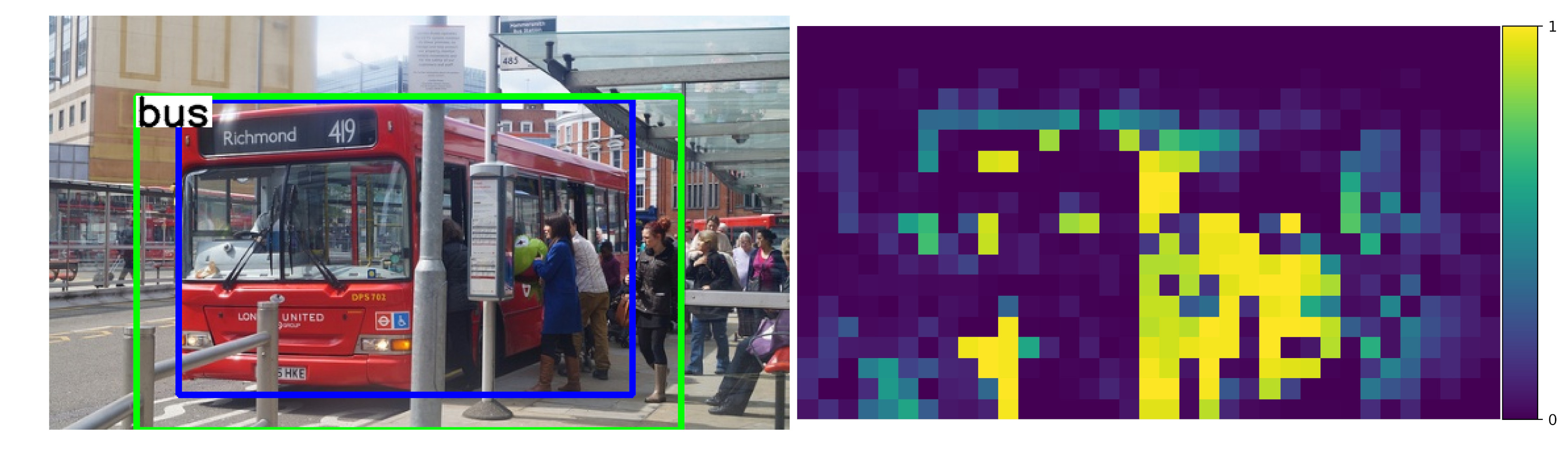}
        \includegraphics[height=2.8cm]{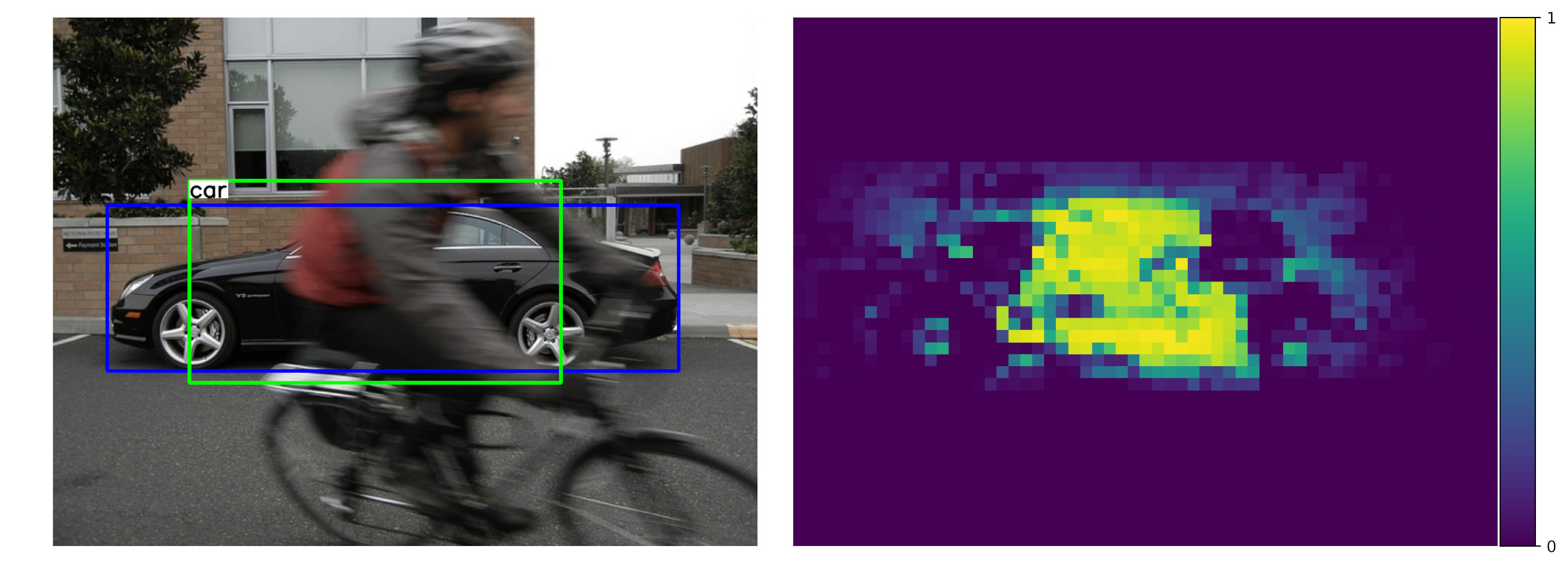}
    \end{subfigure}
    \\
    \begin{subfigure}{\linewidth}
       \centering
        \includegraphics[height=2.8cm]{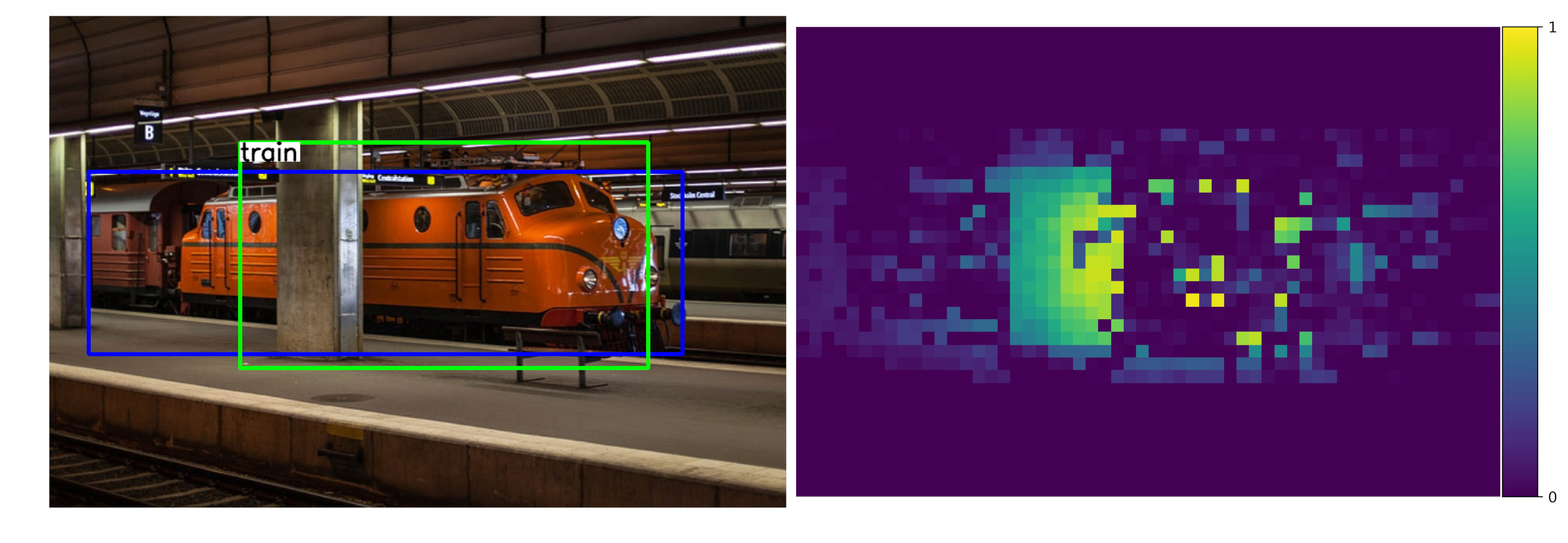}
        \includegraphics[height=2.8cm]{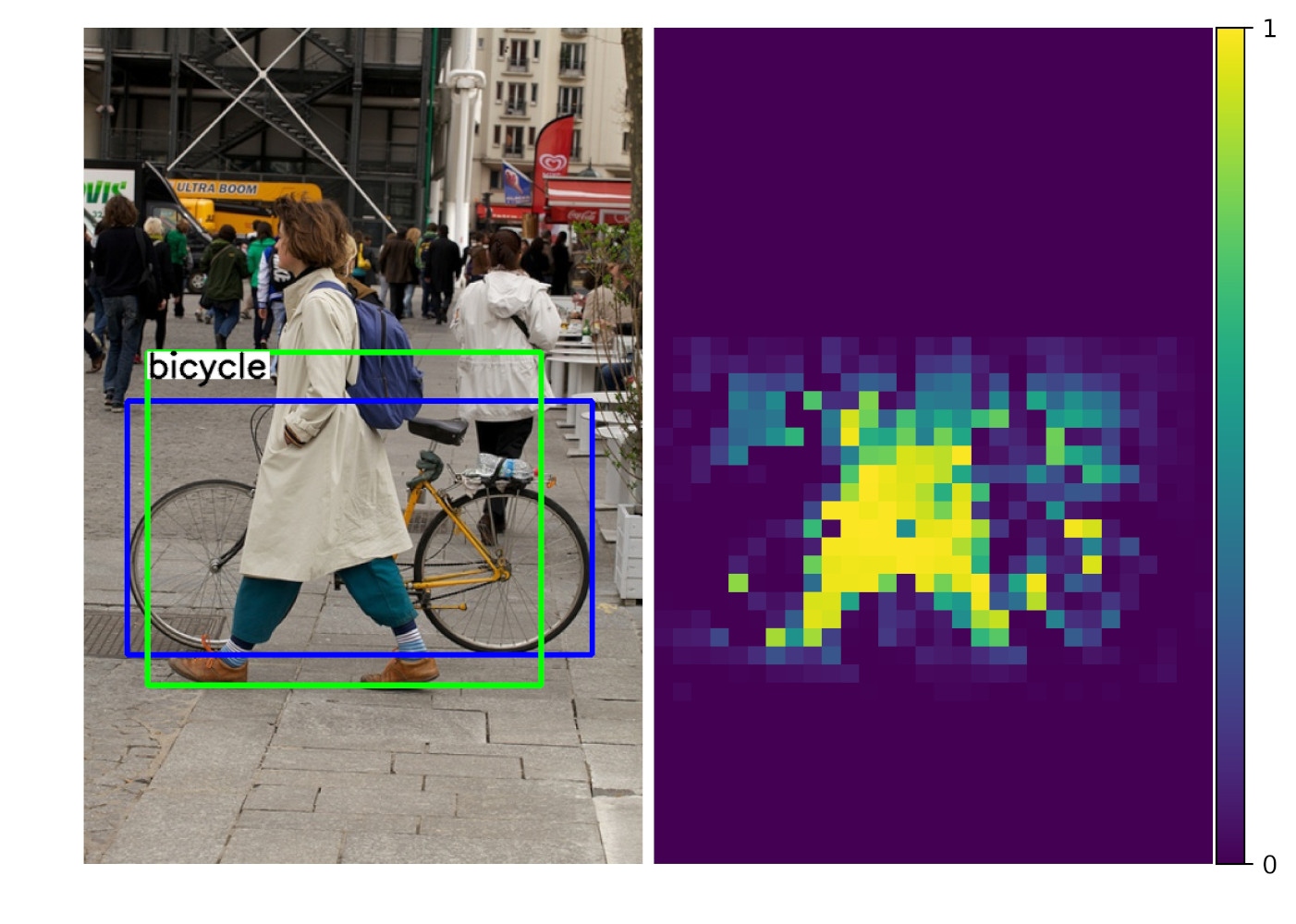}
        \includegraphics[height=2.8cm]{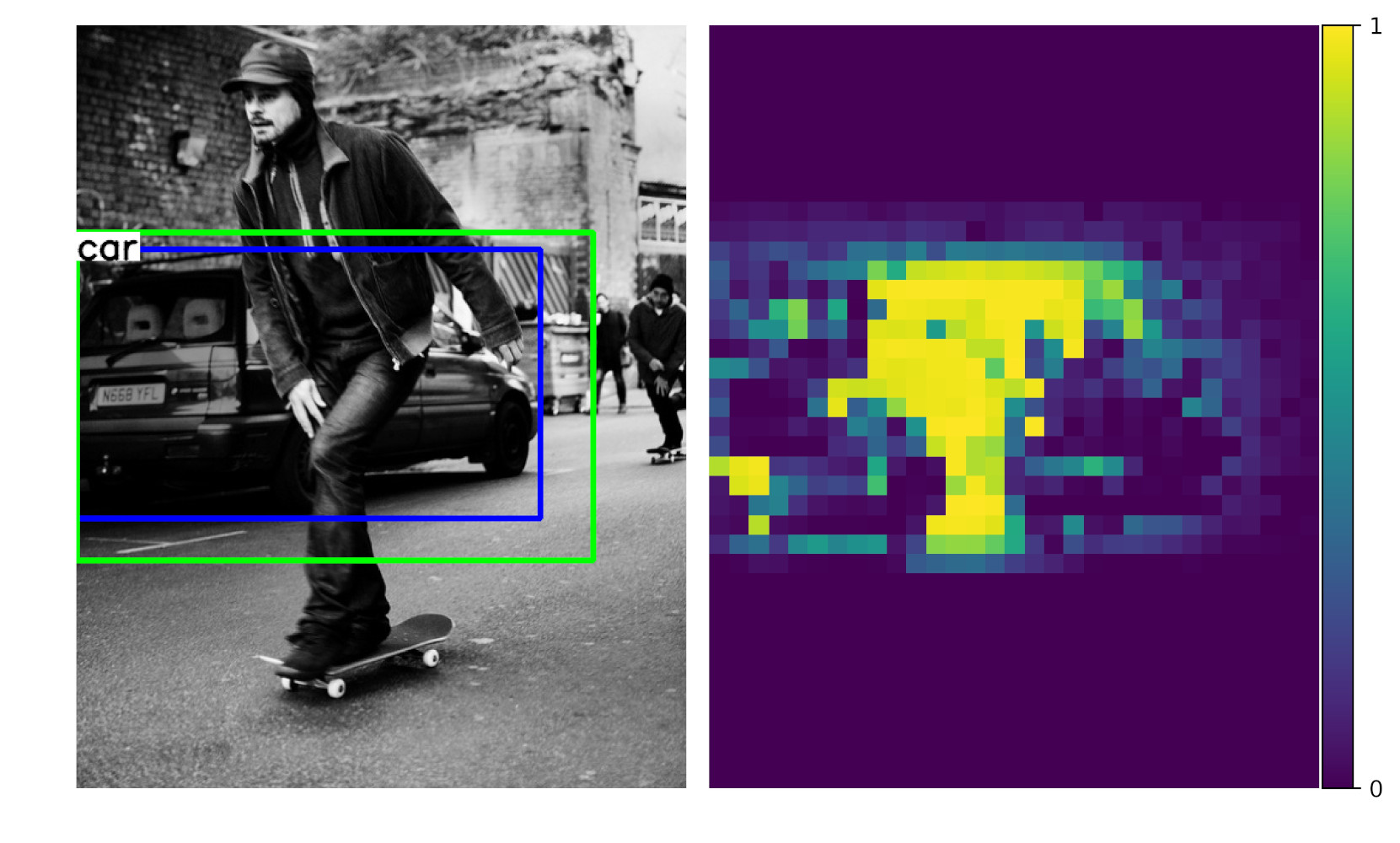}
        \caption{}
    \end{subfigure}
	\vspace{.5cm}
	\\
	\begin{subfigure}{\linewidth}
	        \begin{subfigure}{\linewidth}
	            \centering
        		\includegraphics[height=1.1cm]{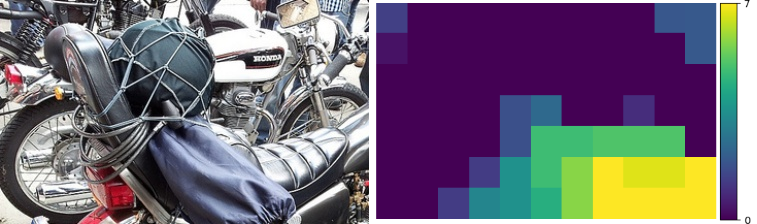}
                \includegraphics[height=1.1cm]{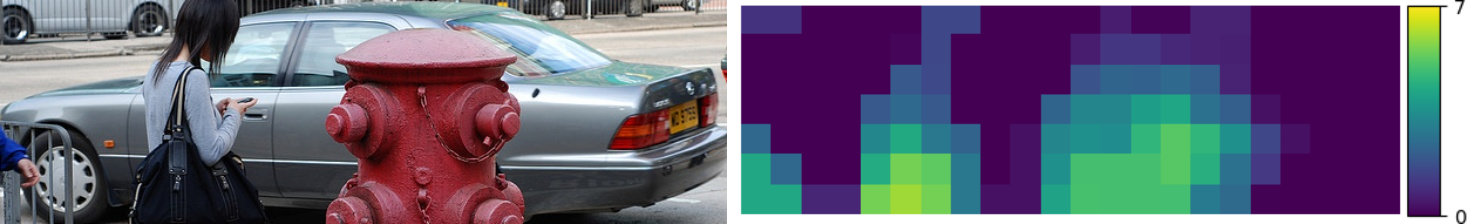}
                \includegraphics[height=1.1cm]{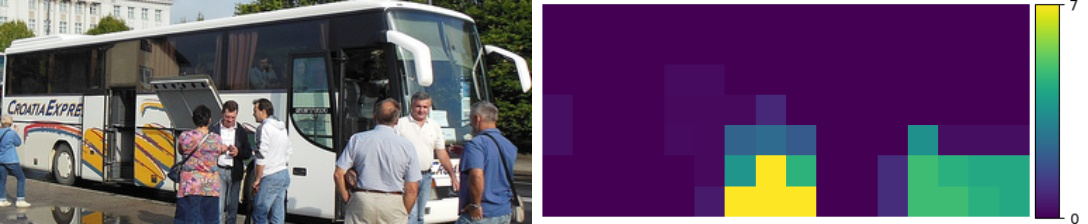}
            \end{subfigure}
            \vspace{.1cm}
            \\
	        \begin{subfigure}{\linewidth}
	            \centering
        		\includegraphics[height=1.29cm]{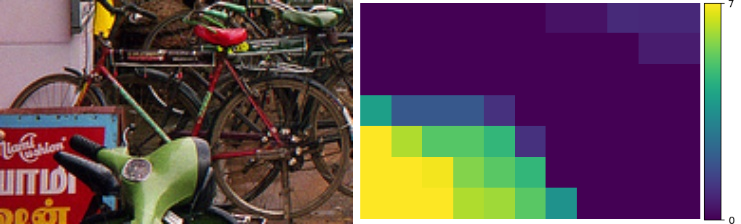}
                \includegraphics[height=1.29cm]{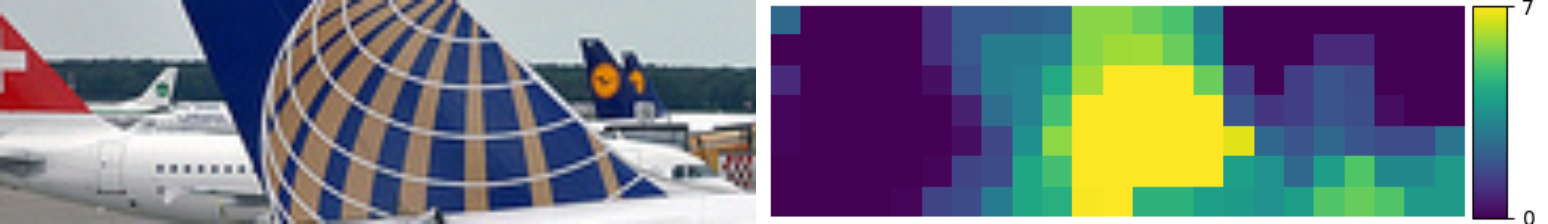}
                \includegraphics[height=1.29cm]{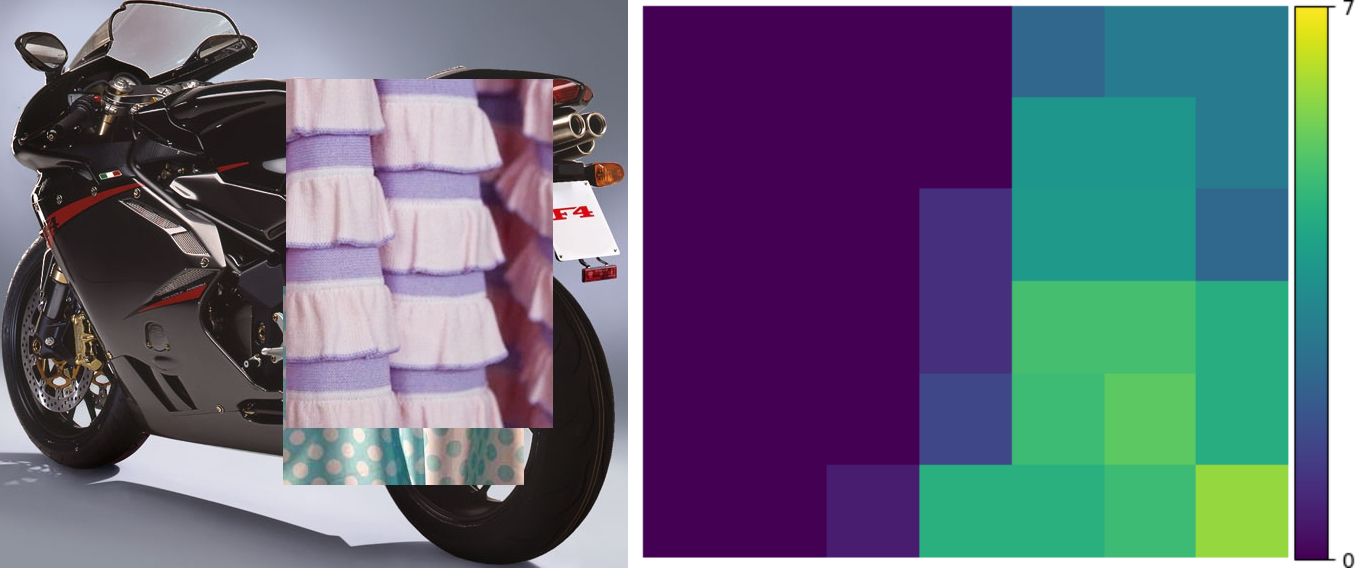}
            \end{subfigure}      
            \vspace{.1cm}
            \\
	        \begin{subfigure}{\linewidth}
	            \centering
        		\includegraphics[height=1.32cm]{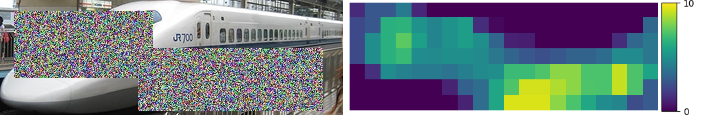}
                \includegraphics[height=1.32cm]{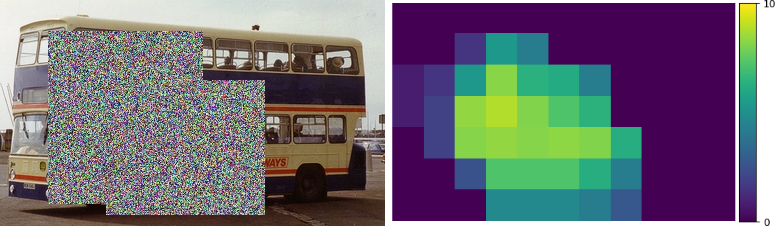}
                \includegraphics[height=1.32cm]{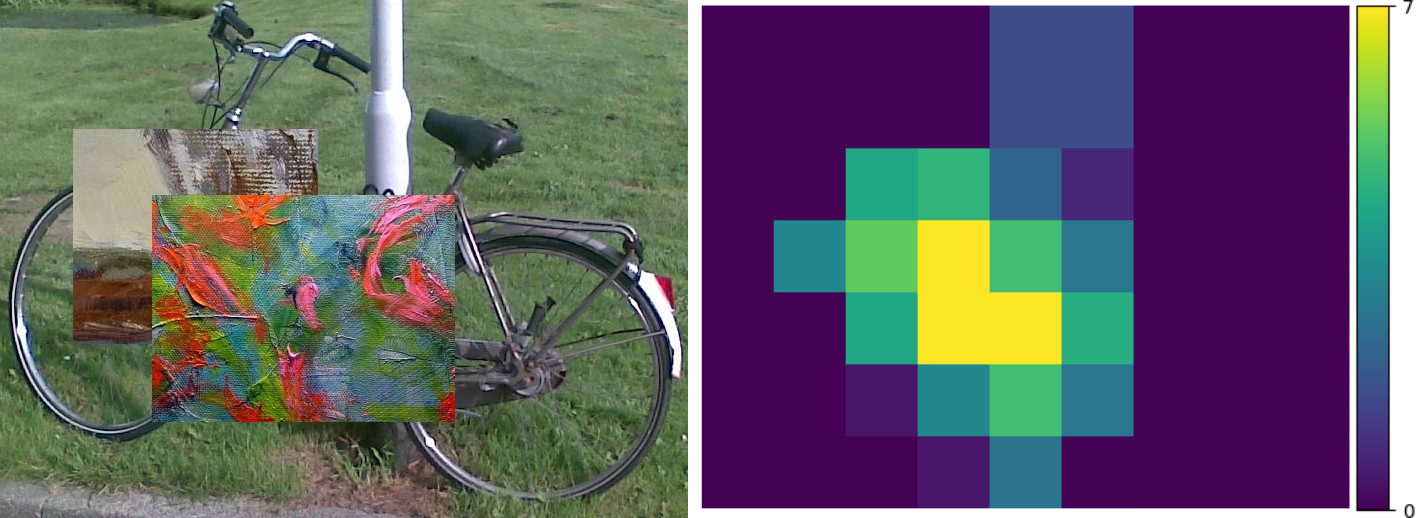}
            \end{subfigure}                 
            \caption{}
            \label{exp_occ_loc_detect_class}
	\end{subfigure}
	\caption{Qualitative results for occluder localization in: (a) Object detection with a context-aware CompNet-VGG16-pool4. (b) Image classification with a CompNet-VGG16-pool5. 
	CompositionalNets can localize occluders accurately for different objects under real and artificial occlusions, despite the high variability of the occluders in terms of shape and appearance. Note that occluder localization is performed independently per pixel in the feature maps.
	}
	\label{fig:exp_occ_loc_detect_qual}
\end{figure*}

\textit{Occluder localization in image classification.} 
We study occluder localization quantitatively (Figure \ref{fig:occ_loc_class}) for correctly classified images from the Occluded-P3D+-Vehicles dataset using the ground truth segmentation masks of the occluders and the objects. 
We show qualitative results in Figure \ref{fig:exp_occ_loc_detect_qual}b.
The evaluation is done on the occlusion level $L2$ , hence a pixel on the object will be occluded with $50\%$ chance. 
We evaluate CompositionalNets learned from different DCNN backbones (VGG-16, ResNet50, ResNext) from the last and second-last layer before the classifier. 
All models were trained for classifying non-occluded vehicles of the PASCAL3D+ dataset (classification performance can be seen in Table \ref{tab:p3d_class}).

In general, we can observe from the ROC curves in Figure \ref{fig:occ_loc_class} that \textbf{CompositionalNets can localize occluders accurately}, although there are differences in terms of their performance.
Furthermore, we observe that it is more difficult for the models to localize natural object occluders compared to box-shaped occluders, probably because of the fine-grained irregular shape of objects. 

\textit{Insights into robustness to partial occlusion in neural networks.} When comparing our experimental results at image classification and occluder localization in more detail, we make four interesting observations:
\begin{enumerate}
    \item CompositionalNets learned from the lower layers of a backbone (pool4 and RB3) can consistently localize occluders more accurately when compared to models learned from higher layers of the same backbone.
    \item The CompositionalNets learned from the VGG backbone (Figure \ref{fig:roc_vgg}) are more successful at localizing occluders, compared to models learned from layers with a similar resolution of ResNet50 and ResNext (Figure \ref{fig:roc_res50} \& \ref{fig:roc_resnext}). 
    \item The performance of CompositionalNets with residual backbones is higher on images with real occlusion, compared to those with artificial object occluders. In contrast, we observe the opposite for CompositionalNets learned from the VGG backbone.
    \item The ability to localize occluders more accurately does not directly translate into a superior performance at classifying partially occluded objects (Table \ref{tab:p3d_class}). For example all three high-level models perform similarly at localizing ``object`` occluders, nevertheless CompNet-ResNext-RB4 performs more than $10\%$ better at classifying images with these types of occluders at level $L3$ compared to the other models. This phenomenon can also be observed for the ``white box`` occluders, which CompNet-ResNext cannot localize as accurately as CompNet-VGG16-pool5, while their performance is on par at level $L3$.
\end{enumerate}

In general, \textbf{neural networks can exploit two complementary approaches to achieve robustness to occlusion}, depending on the backbone architecture:
(A1) When the backbone is powerful enough, the features can become robust to occlusion, in a similar way as they are robust to illumination or viewpoint changes. Hence, by using a powerful feature extractor, residual models can achieve a very high classification performance even when using a rather simple classifier (global average pooling and one fully-connected layer. 
(A2) When the backbone cannot learn features that are robust to occlusion, then a more complex classifier is required, such as the multiple fully-connected layers in the VGG network.

Based on this intuition, our experimental results for occluder localization and object recognition lead us to the following conjecture: 
Using powerful residual backbones, CompositionalNets achieve high classification performance, because of the highly discriminative features. However, they cannot localize the occluders as accurately, because the robustness to occlusion makes it difficult to distinguish between occluded and non-occluded features. In contrast, CompositionalNets based on the VGG backbone can localize the occluders well, because the features are not robust to occlusion. However, those CompositionalNets do not achieve the highest classification performance because the features are less discriminative compared to those of the residual backbones. 
Furthermore, the lower layers in neural networks typically exhibit less invariance. Therefore, CompositionalNets learned from those layer can consistently achieve better localization performance compared to those learned from higher layers of the same backbone.
Finally, CompositionalNets with residual backbones rely more on invariance then on occluder localization. They perform better on real data, because the they rely on invariant features that were learned during ImageNet pre-training. This invariance does do not generalize well to the artificial occluders, therefore their performance is lower compared to the real occlusion scenario. 
In contrast, CompositionalNets learned from VGG rely less on invariance to occlusion and more on occluder localization. As the artificial occluders are easier to localize, their performance is higher on artificially generated occlusions compared to the real data.

In general, replacing the standard classifiers with compositional models increases the classification performance under occlusion for all models and also enables them to localize occluders. However, the ability to localize occluders and the final classification performance are at odds with each other, depending on the extent to which the features are invariant to occlusion. 

\begin{figure}
    \centering
    \includegraphics[width=.85\linewidth]{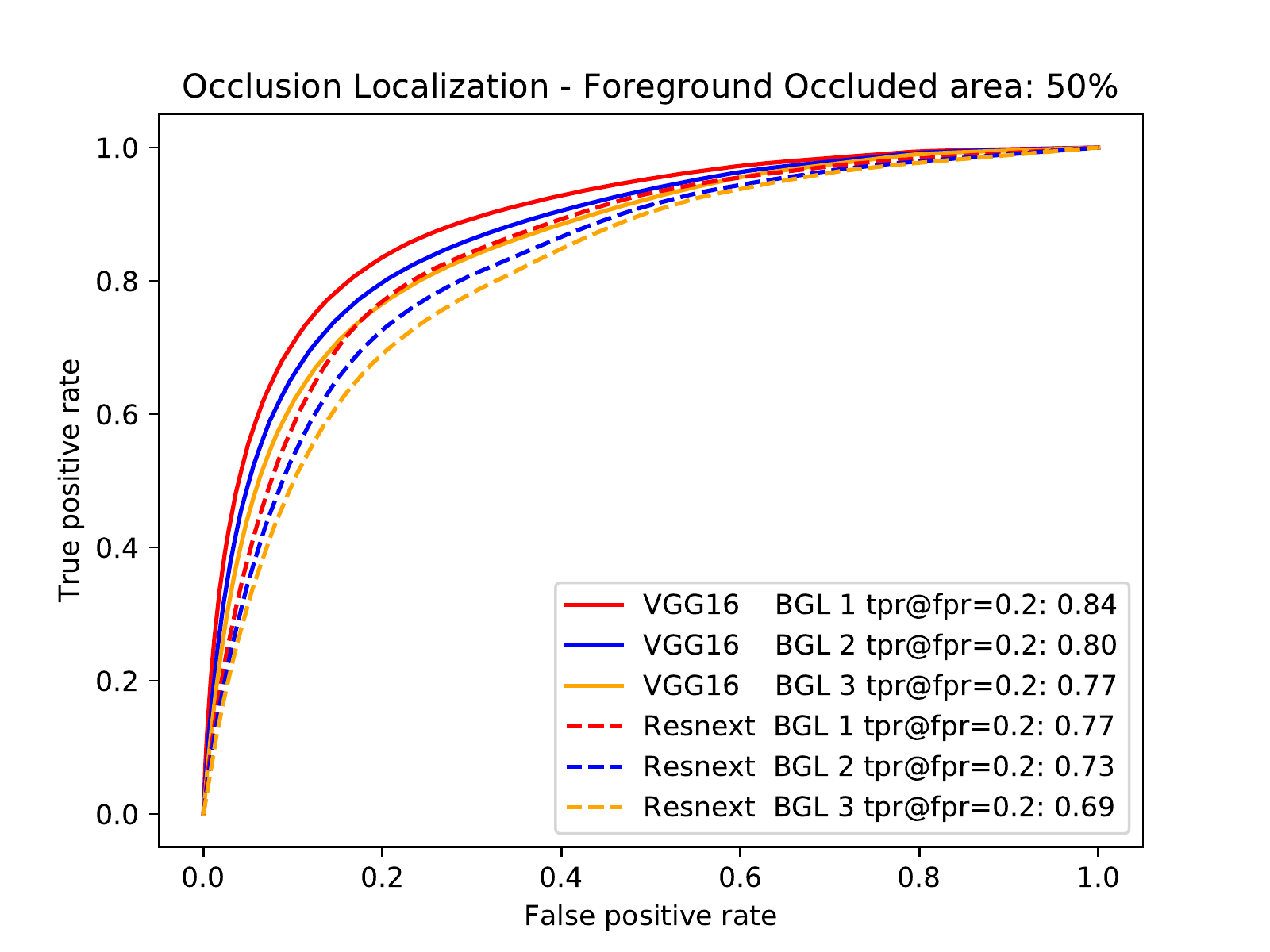}
	\caption{ROC curves measuring occlusion localization scores in object detection with context-aware CompositionalNets learned from pool4 of VGG (solid lines) and RB3 of ResNext using $\omega=0.2$. We the object is on average $50\%$ occluded at each level of background occlusion (colored lines). Note that context-aware CompositionalNets can predict the occluded regions of the objects accurately at object detection.}
	\label{fig:occ_loc_detect}
\end{figure}

\textit{Occluder localization in object detection.} 
Figure \ref{fig:occ_loc_detect} illustrates the ROC curve of a context-aware CompNet-VGG16-pool4 and CompNet-RXT-RB3 for successfully detected objects. 
We can observe that they can predict occluded regions accurately. 
Furthermore, the performance increases compared to the classification experiments as the context-awareness reduces false-positive detections in the background regions.
In Figure \ref{fig:exp_occ_loc_detect_qual}a, we show qualitative results of a context-aware CompNet-VGG16-pool4 at occluder localization in object detection. 
We illustrate results for artificially occluded objects and real occlusions from the MS-COCO dataset, in which the CompositionalNet could successfully locate the objects.
Overall, the model can locate occluders with high accuracy, despite their large variability in terms of appearance and shape. 
Note how the shape of the occluders is outlined accurately, although the localization is done for each pixel in the feature map independently.
In summary, we observe that the occluder localization results for object detection are consistent with those for image classification, in that they confirm the ability of CompositionalNets at localizing occluders accurately.

\begin{figure*}
    \centering
    \includegraphics[width=0.8\linewidth]{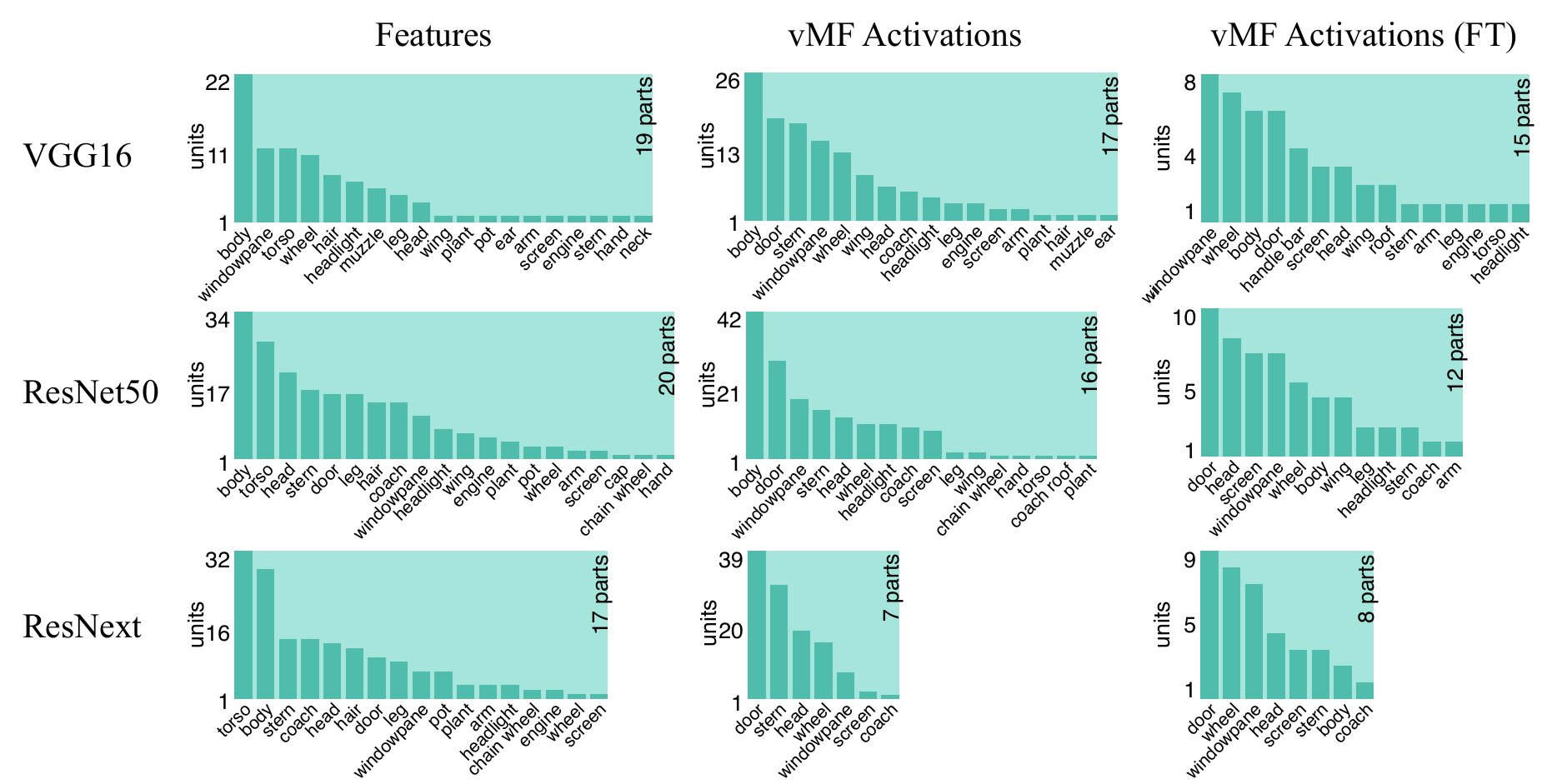}
    \caption{Network dissection results for different backbone architectures and different activation layers on the Broden dataset ``Part`` category. The horizontal axis of the bar-plots represents the parts with above-threshold correlation score, and the vertical axis represents the number of hidden units that are correlated with the specific part. 
    Note that the vMF kernel activations are more focused on the vehicle related parts, while after fine-tuning the number of correlated units for each parts is reduced, hence avoiding redundancies in the representation.}
    \label{fig:netdissect}
\end{figure*}

\subsubsection{Interpretation of vMF Kernels}
\label{sec:exp_dissect}

\begin{figure*}
    \centering
	\begin{subfigure}{0.249\linewidth}
		\centering
		\includegraphics[width=.95\linewidth]{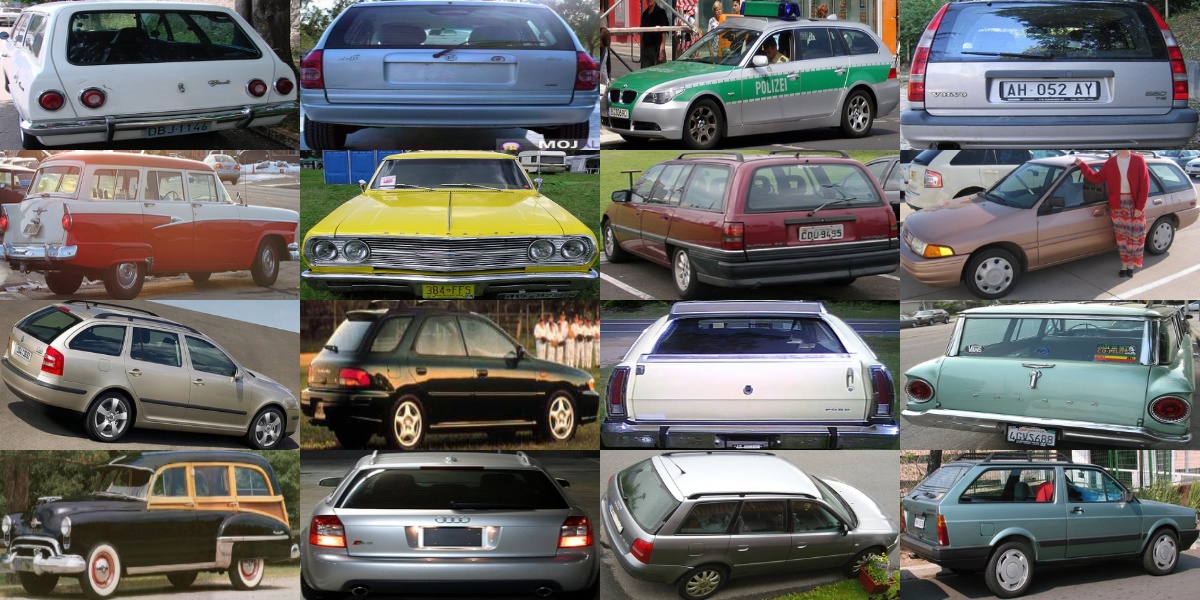}
	\end{subfigure}%
	\begin{subfigure}{0.249\linewidth}
		\centering
		\includegraphics[width=.95\linewidth]{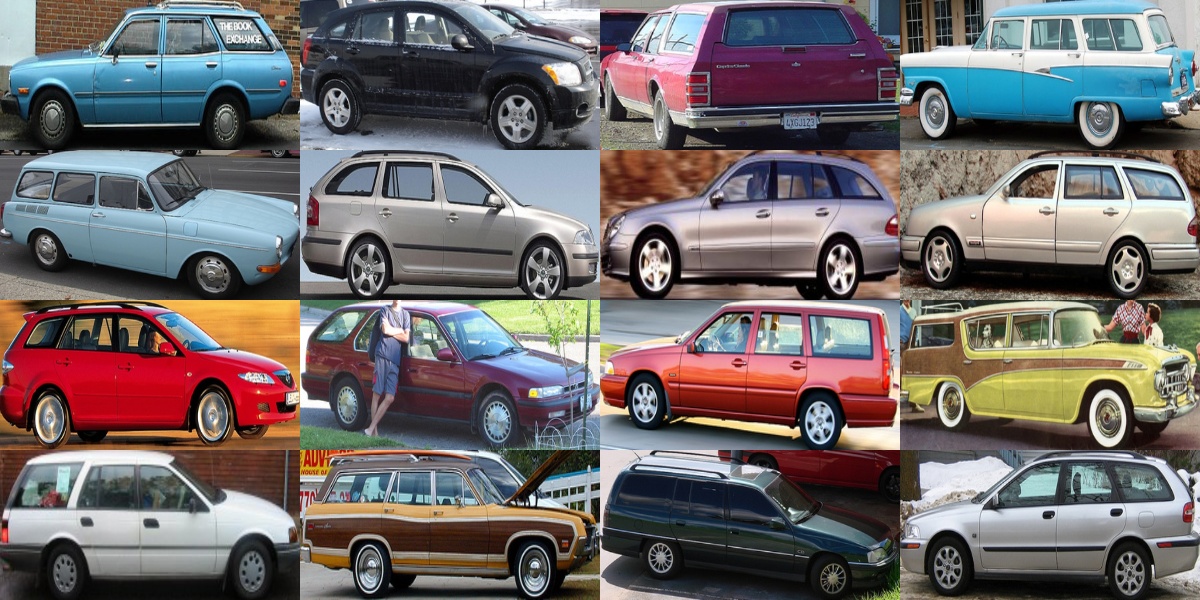}
	\end{subfigure}%
	\begin{subfigure}{0.249\linewidth}
		\centering
		\includegraphics[width=.95\linewidth]{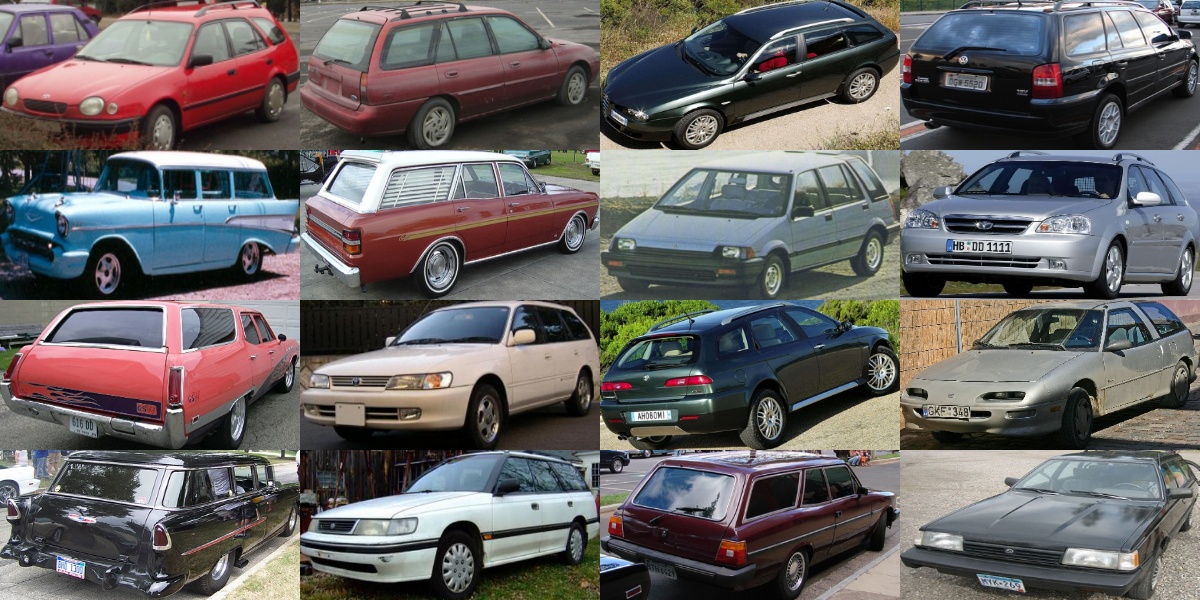}
	\end{subfigure}%
	\begin{subfigure}{0.249\linewidth}
		\centering
		\includegraphics[width=.95\linewidth]{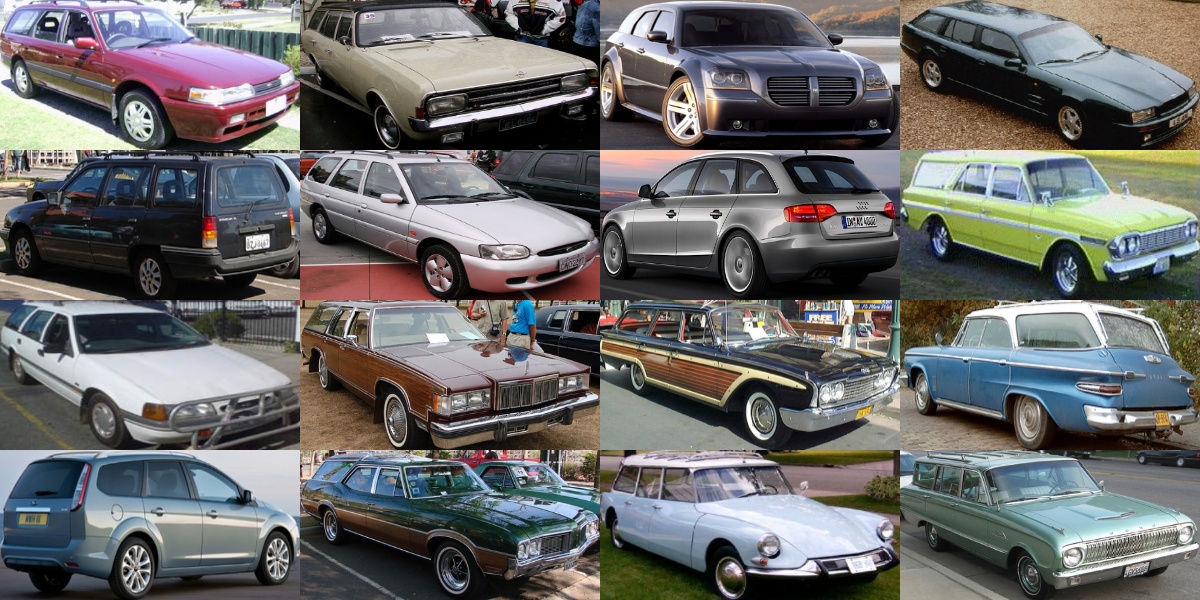}
	\end{subfigure}%

	\begin{subfigure}{0.249\linewidth}
		\centering
		\includegraphics[width=.75\linewidth]{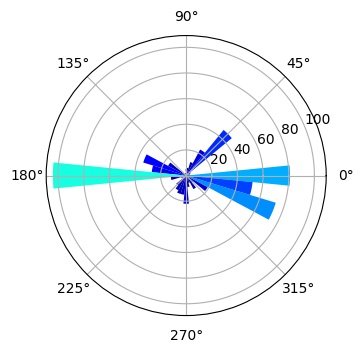}
	\end{subfigure}%
	\begin{subfigure}{0.249\linewidth}
		\centering
		\includegraphics[width=.75\linewidth]{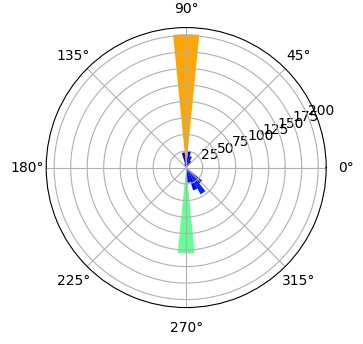}
	\end{subfigure}%
	\begin{subfigure}{0.249\linewidth}
		\centering
		\includegraphics[width=.75\linewidth]{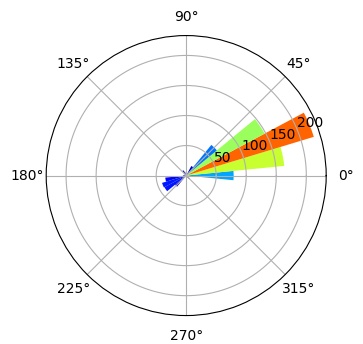}
	\end{subfigure}%
	\begin{subfigure}{0.249\linewidth}
		\centering
		\includegraphics[width=.75\linewidth]{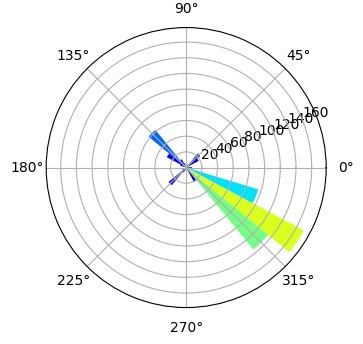}
	\end{subfigure}%

	\begin{subfigure}{0.249\linewidth}
		\centering
		\includegraphics[width=.95\linewidth]{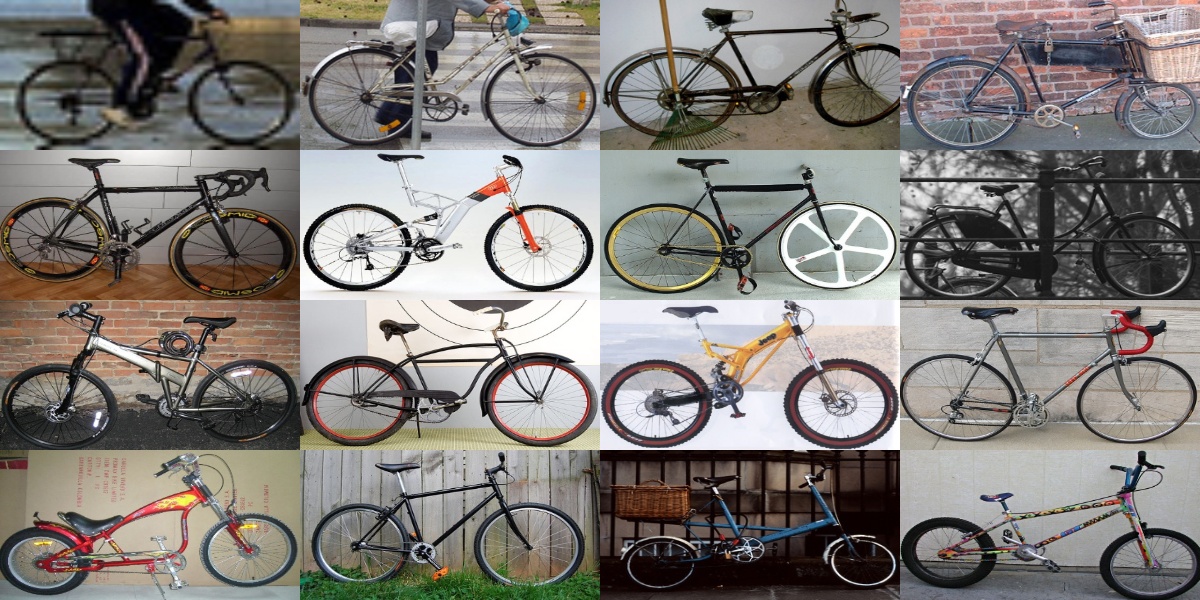}
	\end{subfigure}%
	\begin{subfigure}{0.249\linewidth}
		\centering
		\includegraphics[width=.95\linewidth]{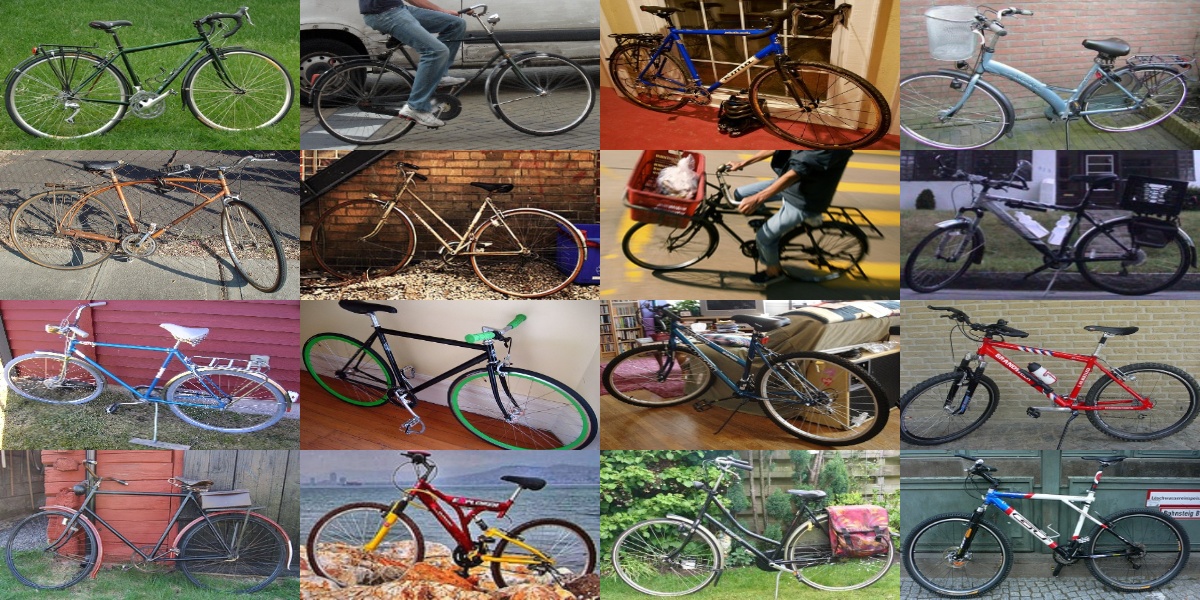}
	\end{subfigure}%
	\begin{subfigure}{0.249\linewidth}
		\centering
		\includegraphics[width=.95\linewidth]{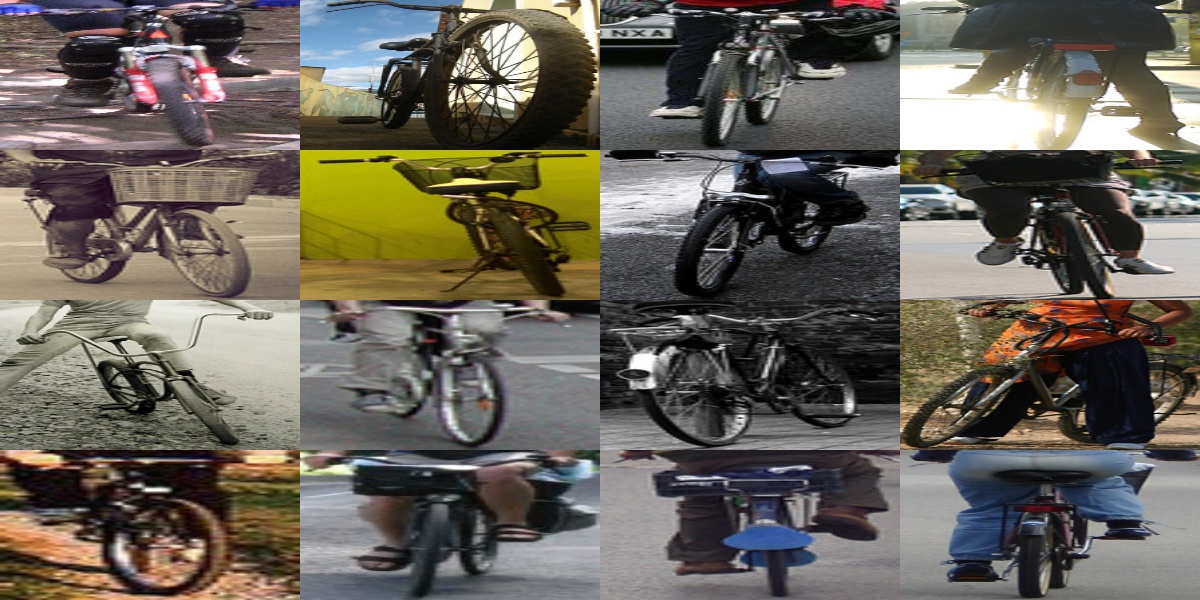}
	\end{subfigure}%
	\begin{subfigure}{0.249\linewidth}
		\centering
		\includegraphics[width=.95\linewidth]{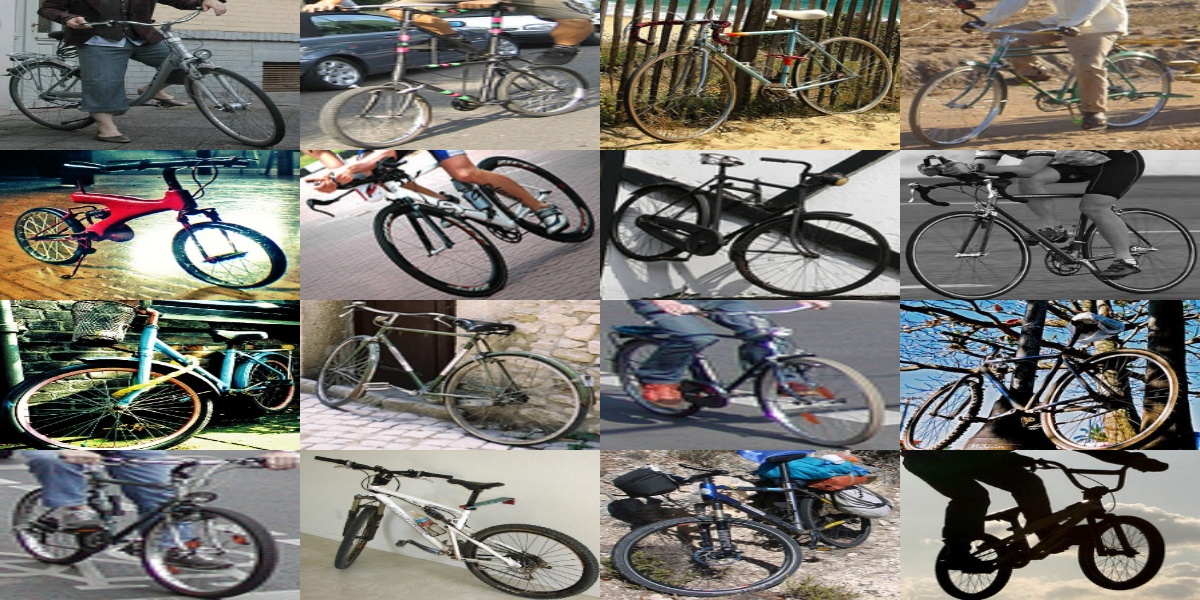}
	\end{subfigure}%

	\begin{subfigure}{0.249\linewidth}
		\centering
		\includegraphics[width=.75\linewidth]{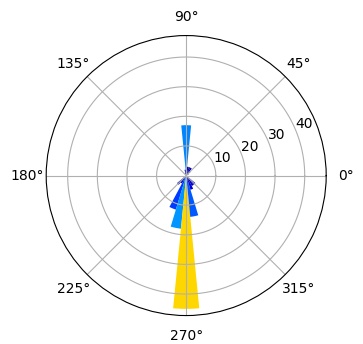}
	\end{subfigure}%
	\begin{subfigure}{0.249\linewidth}
		\centering
		\includegraphics[width=.75\linewidth]{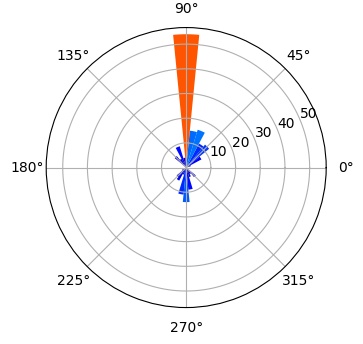}
	\end{subfigure}%
	\begin{subfigure}{0.249\linewidth}
		\centering
		\includegraphics[width=.75\linewidth]{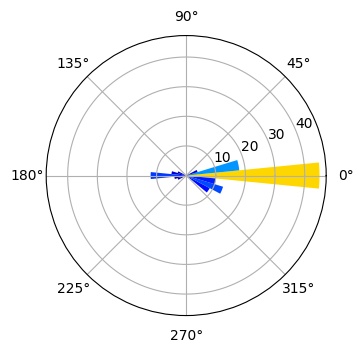}
	\end{subfigure}%
	\begin{subfigure}{0.249\linewidth}
		\centering
		\includegraphics[width=.75\linewidth]{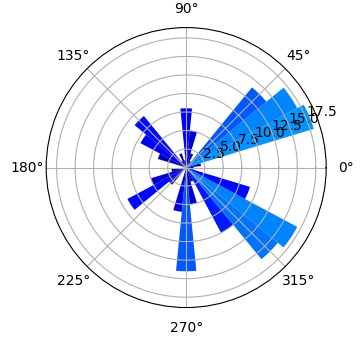}
	\end{subfigure}%

	\begin{subfigure}{0.249\linewidth}
		\centering
		\includegraphics[width=.95\linewidth]{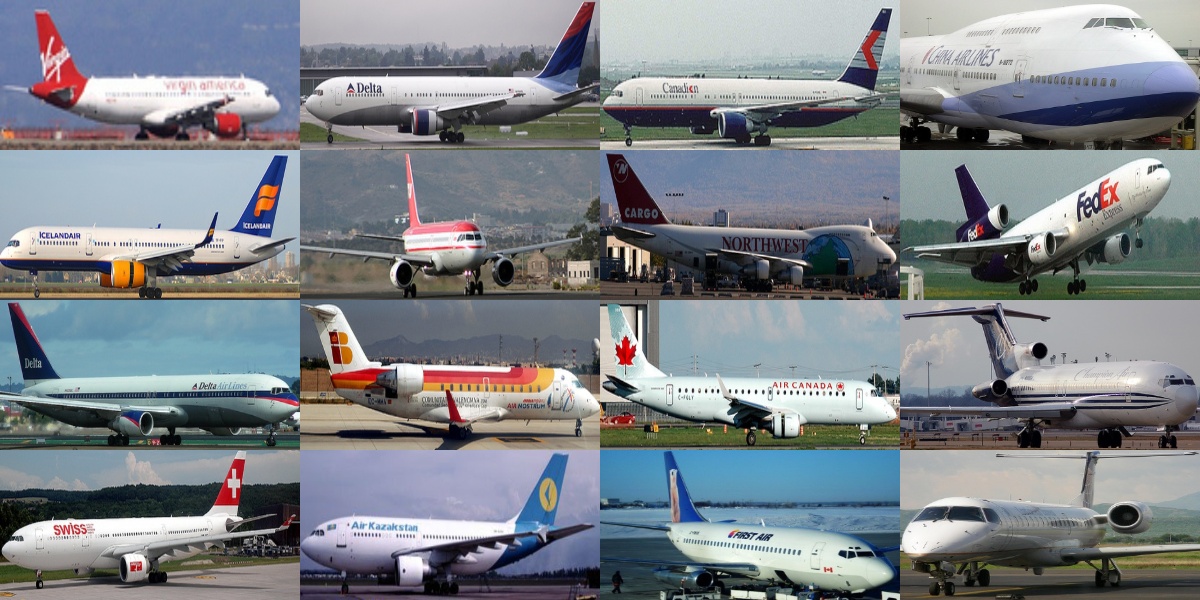}
	\end{subfigure}%
	\begin{subfigure}{0.249\linewidth}
		\centering
		\includegraphics[width=.95\linewidth]{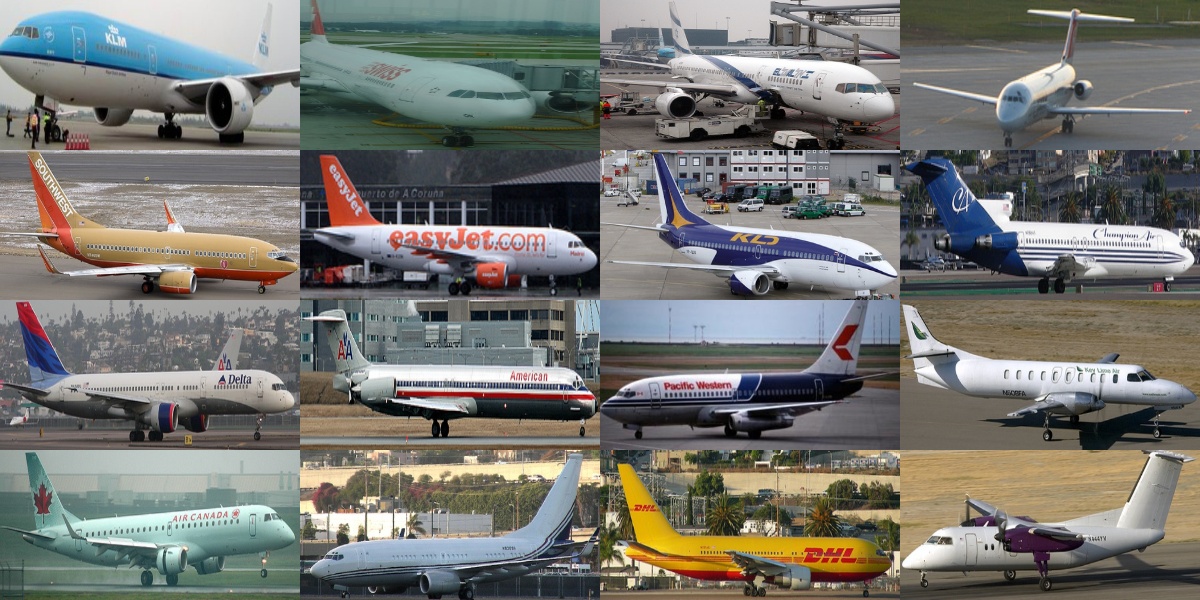}
	\end{subfigure}%
	\begin{subfigure}{0.249\linewidth}
		\centering
		\includegraphics[width=.95\linewidth]{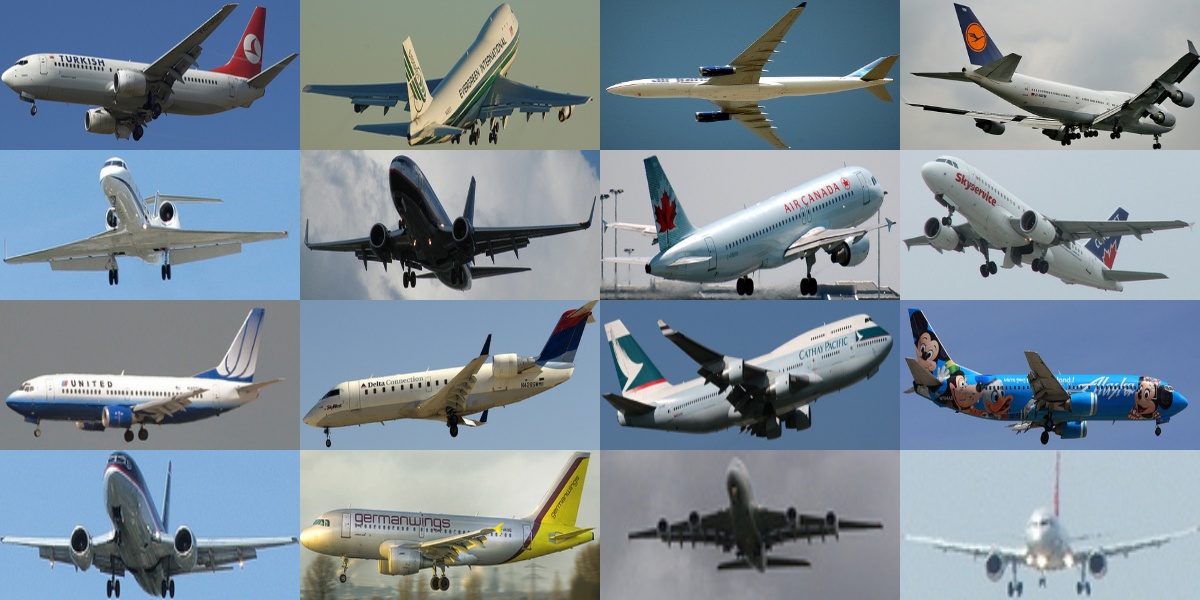}
	\end{subfigure}%
	\begin{subfigure}{0.249\linewidth}
		\centering
		\includegraphics[width=.95\linewidth]{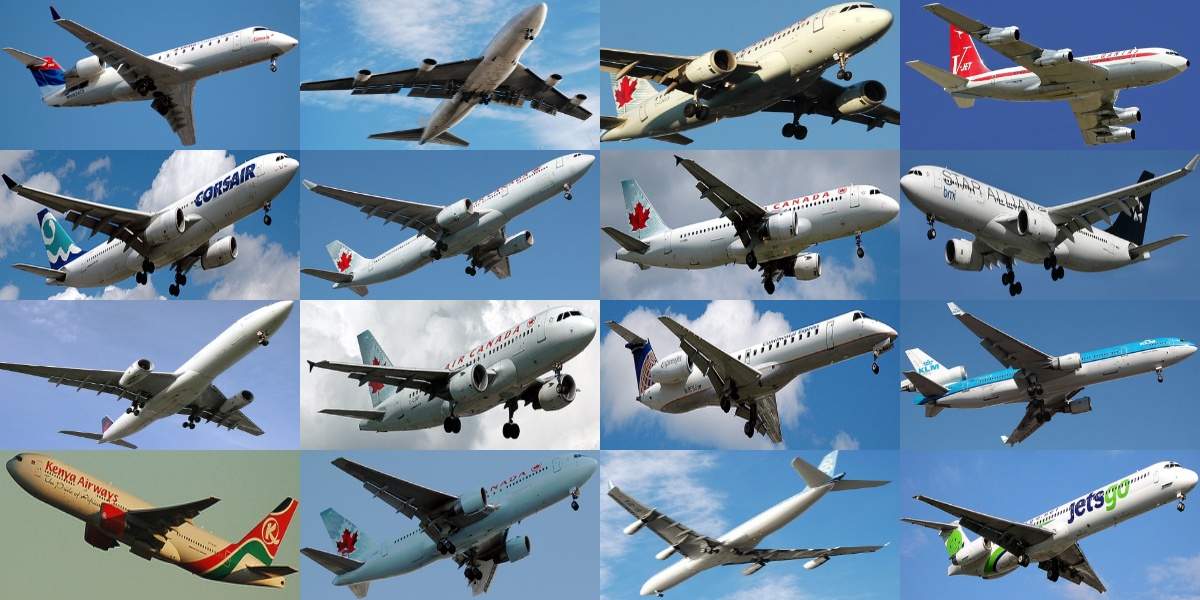}
	\end{subfigure}%

	\begin{subfigure}{0.249\linewidth}
		\centering
		\includegraphics[width=.75\linewidth]{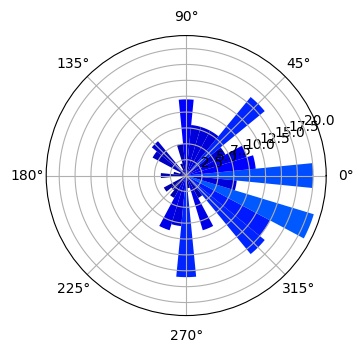}
	\end{subfigure}%
	\begin{subfigure}{0.249\linewidth}
		\centering
		\includegraphics[width=.75\linewidth]{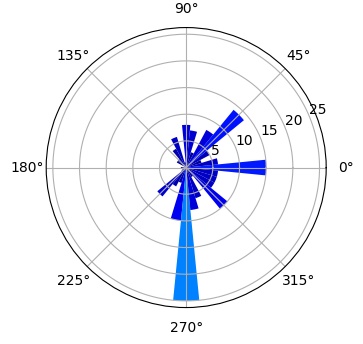}
	\end{subfigure}%
	\begin{subfigure}{0.249\linewidth}
		\centering
		\includegraphics[width=.75\linewidth]{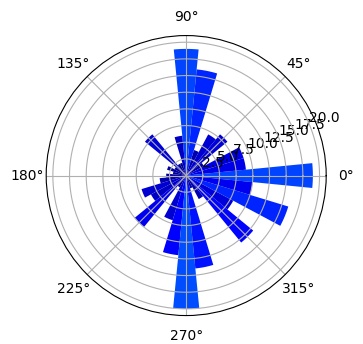}
	\end{subfigure}%
	\begin{subfigure}{0.249\linewidth}
		\centering
		\includegraphics[width=.75\linewidth]{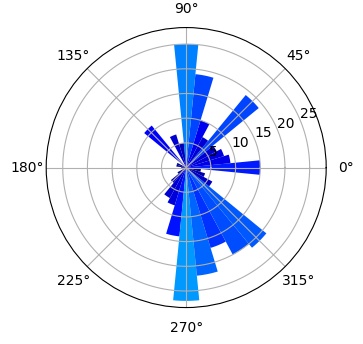}
	\end{subfigure}%
	
	\begin{subfigure}{0.249\linewidth}
		\centering
		\includegraphics[width=.95\linewidth]{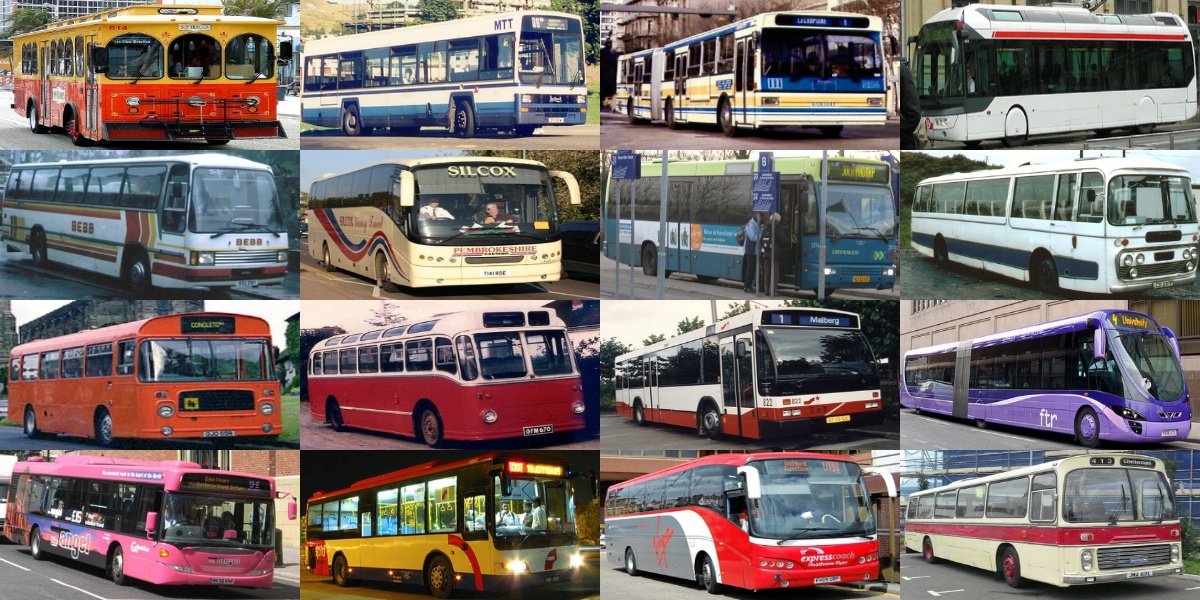}
	\end{subfigure}%
	\begin{subfigure}{0.249\linewidth}
		\centering
		\includegraphics[width=.95\linewidth]{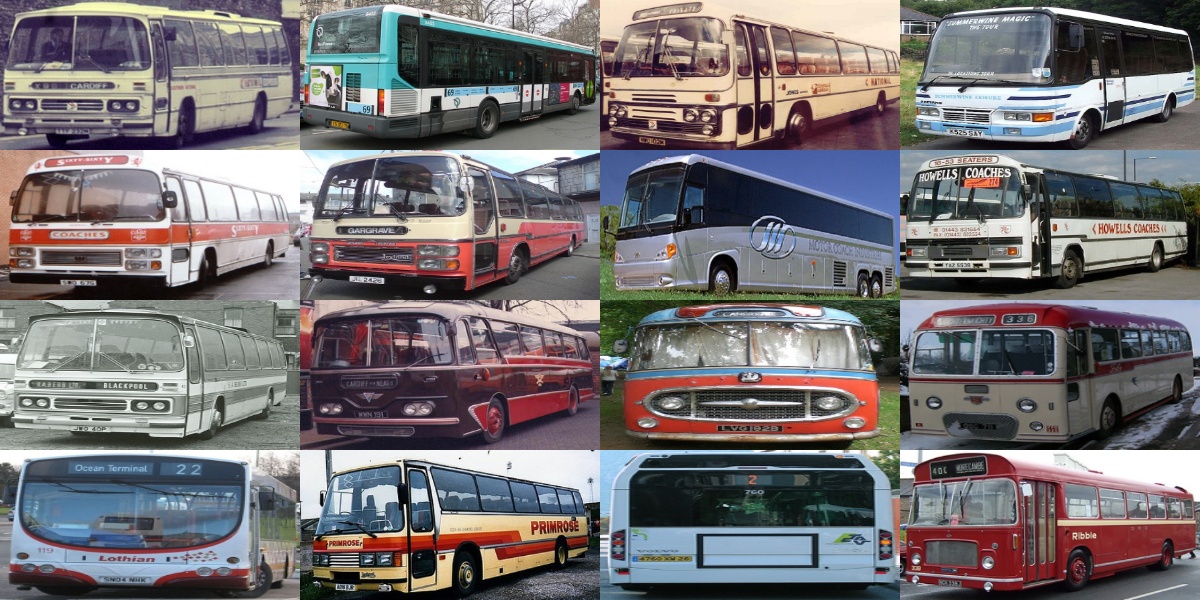}
	\end{subfigure}%
	\begin{subfigure}{0.249\linewidth}
		\centering
		\includegraphics[width=.95\linewidth]{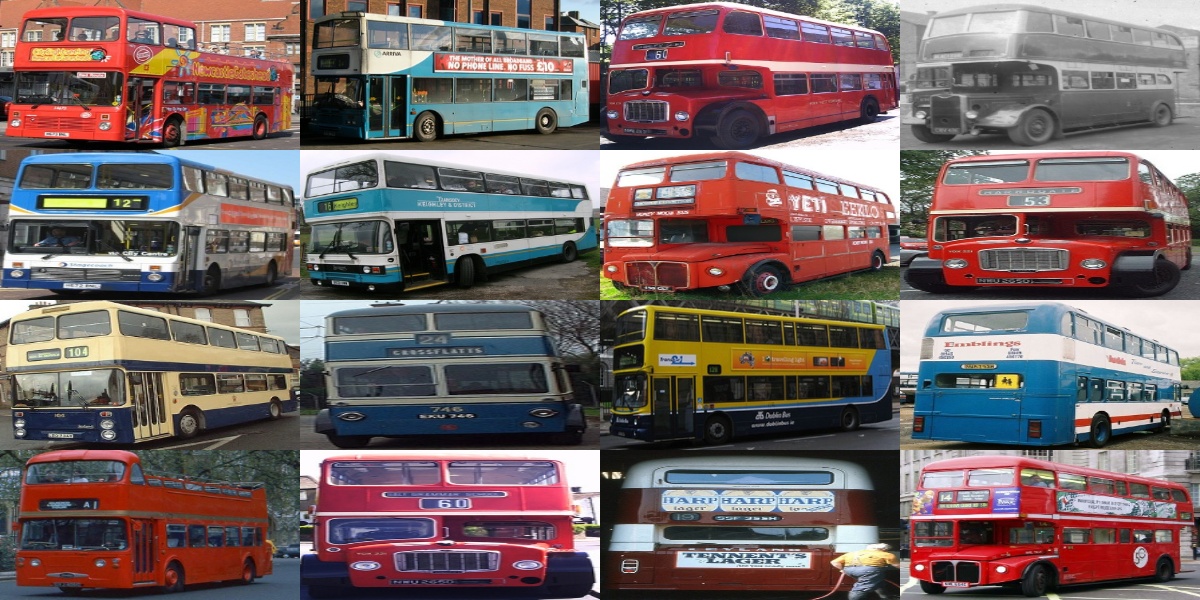}
	\end{subfigure}%
	\begin{subfigure}{0.249\linewidth}
		\centering
		\includegraphics[width=.95\linewidth]{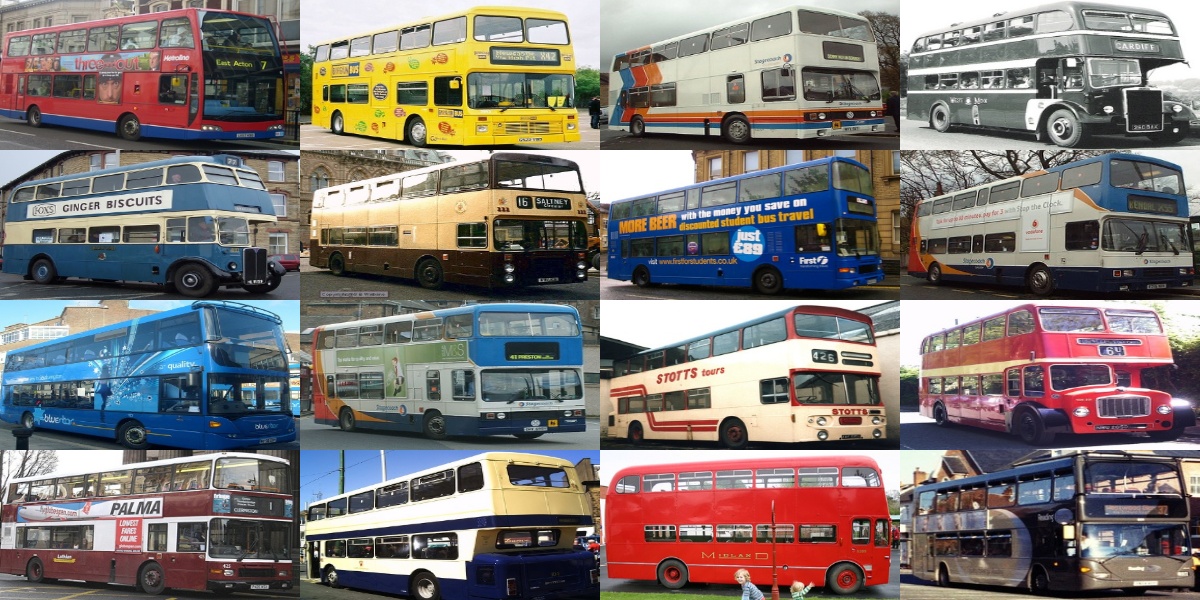}
	\end{subfigure}%

	\begin{subfigure}{0.249\linewidth}
		\centering
		\includegraphics[width=.75\linewidth]{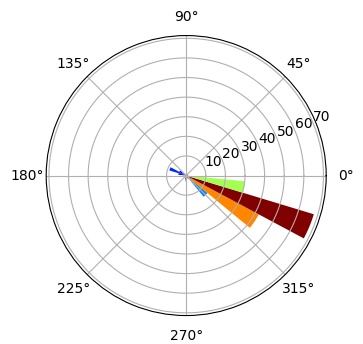}
	\end{subfigure}%
	\begin{subfigure}{0.249\linewidth}
		\centering
		\includegraphics[width=.75\linewidth]{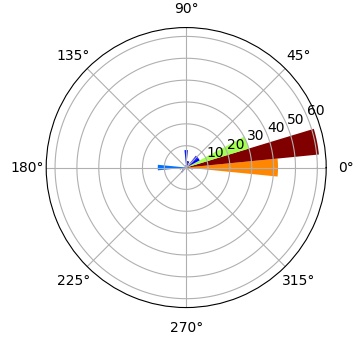}
	\end{subfigure}%
	\begin{subfigure}{0.249\linewidth}
		\centering
		\includegraphics[width=.75\linewidth]{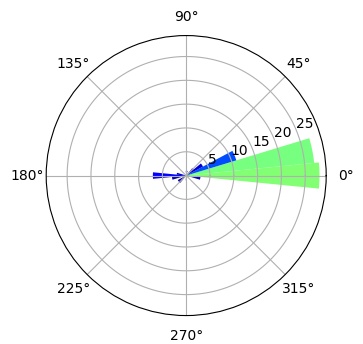}
	\end{subfigure}%
	\begin{subfigure}{0.249\linewidth}
		\centering
		\includegraphics[width=.75\linewidth]{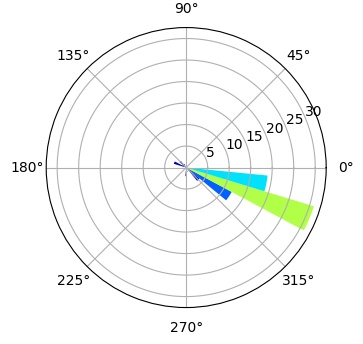}
	\end{subfigure}%
	\caption{Visualization of mixture components $p(F|\theta^m_y)$ for $M=4$ components learned from the \texttt{pool5} layer of a VGG-16 networks and corresponding azimuth pose distribution (below each visualization). Note how images with different 3D viewpoint are approximately separated into different components.}
    \label{fig:exp_view}
\end{figure*}

We further investigate the interpretability of our CompositionalNets using network dissection as proposed by \cite{bau2017network}. In short, network dissection looks at the top activation of the hidden units and correlates them with a large range of human labeled visual concepts in the Broden dataset. Most of the concepts are annotated as segmentation mask with input resolution and the activation maps are up-scaled to the same size to calculate the intersection over union (IoU) scores. By setting a threshold for the best matched score, Network Dissection studies the latent representations of various layers in a network. Since CompositionalNets are unifying part-based compositional models with deep neural networks, intuitively we expect to see the hidden units in CompositionalNets to be more correlated with visual concepts of parts, since the models evaluated are trained for classifying vehicles in PASCAL3D+, we expect to see more vehicle related parts emerge.

For the experiment, we adopt the code from the authors of \cite{bau2017network} and use most default settings, except that we change the testing categories into ``part'' only. This means that during the hidden unit and visual concepts correlation test, only classes from the part category are involved. Note that these classes include both vehicle parts (e.g., windowpane, wheel, stern, etc.) and non-vehicle ones (e.g., hair, torso, muzzle, etc.).

As aforementioned, we expect to see the hidden units in CompositionalNets to be highly correlated with vehicle related parts. More specifically, we are interested in studying (1) what part concepts are correlated with the units before and after vMF kernels, and (2) how the end-to-end training affects these correlations. In Figure \ref{fig:netdissect}, we examine three different backbone architectures, one in each row, and three different types of hidden units, one in each column. ``Features'' are the hidden units from the layer before the vMF kernel, which is pool5 for VGG16 and the residual block four (RB4) layer for ResNet50 and ResNext, all with weights initialized from ImageNet pretrained models. ``vMF Activations'' are the units right after the vMF kernel, where the kernel weights are initialized by clustering as described in Section \ref{sec:compnet}. ``vMF Activations (FT)'' shows the same units after the end-to-end fine-tuning. In the barplots, the horizontal axis lists the parts with scores above the threshold, while the vertical axis shows the number of hidden units that are correlated with a specific part.

Comparing the ``Features'' versus the ``vMF Activations'', we observe that the vMF kernels indeed help the hidden units to be more concentrated on the vehicle related parts. The non-vehicle parts, such as ``pot'' and ``cap'', are removed while more units become correlated with vehicle ones, like ``door'' and ``stern''. 
However, this might also introduce a lot of redundancy in the representation and hence a waste of computational resources.
When comparing ``vMF Activations'' with ``vMF Activations (FT)'', we find that not only some non-vehicle parts are further removed, the number of units correlated to each vehicle part is also reduced. 
This indicates the training may help the vMF kernels to recognize more diverse part representations and reduce the redundancy in the representation.

It is worth mentioning that the annotation of parts in the Broden dataset is coarse and not specific for different object classes, e.g., windowpane is shared for car, airplane, and house, etc.. Therefore, it may not match the internal representations learned by the CompositionalNets.
Nevertheless, the results support our conjecture observed in Figure \ref{fig:vc} that the vMF kernels work as part detectors and CompositionalNets learn part representations without supervision.

\subsubsection{Interpretation of Mixture Components}
\label{sec:exp_view}

In an effort to better understand the inner workings of CompositionalNets we study what the individual mixture components have learned to represent during training. 
We study a CompNet-VGG16-pool5 with $M=4$ mixture components, but the following analysis gives very similar results for other CompositionalNet architectures.

Each mixture component is by design of the training process specific to a particular object class.
We have already shown that the vMF kernels are responsive to individual object parts (see Section \ref{sec:exp_dissect} and Figure \ref{fig:vc}).
During training the mixture components are learned by clustering of the spatial activation patterns of vMF kernel activations.
As the spatial distribution of parts of an object varies significantly with changes in the object pose, the mixture components become specific to certain viewpoints of the objects.
We illustrate this property of CompositionalNets in Figure \ref{fig:exp_view} by showing the training images with highest likelihood in each mixture. 
Furthermore, we illustrate the histogram of poses for all training images in each mixture components as bar plots (using the pose annotations in the PASCAL3D+ dataset). 
As vehicles in natural images mostly vary in terms of their azimuth angle, we restrict the plots to this angle only. 
Each bar plot shows the distributions of azimuth angles in the range of $[0^\circ-360^\circ]$ degrees.
The length of the bars is normalized such that the longest bar indicates the most frequent azimuth angle within each mixture. 
The color of the bar is selected from the colormap ``jet`` and normalized such that the maximal value (dark red) is equivalent to more than $10\%$ of the total number of training images for a particular class.
We can observe from Figure \ref{fig:exp_view} that for the classes ``car``, ``bicycle`` and ``bus`` the mixture components are very specific to objects in a certain azimuth angle. 
Note that this happens despite a significant variability in the objects' appearance and backgrounds.

In contrast, for the class ``airplane`` the pose distribution is less viewpoint specific.
We think that the reason is that airplanes naturally vary in terms of several pose angles, in contrast to vehicles on the street which vary mostly in terms of their azimuth angle w.r.t. the camera.
As the number of mixture components is fixed a-priori it is difficult for the model to become specific to a certain pose during maximum likelihood learning from the data.
In future work, it would therefore be useful to explore unsupervised strategies for determining the number of mixture components per class based on the training data.

%% file: 05_conclusion.tex
\section{Conclusion}
In this work, we studied the problem of generalizing beyond the training data in terms of partial occlusion. 
We showed that current approaches to computer vision based on deep learning fail to generalize to out-of-distribution examples in terms of partial occlusion.
In an effort to overcome this fundamental limitation 
we made several important contributions:
\begin{itemize}
    \item We introduced \textbf{Compositional Convolutional Neural Networks} - a deep model that unifies compositional part-based representations and deep convolutional neural networks (DCNNs). 
    In particular we replace the fully connected classification head of DCNNs with a differentiable generative compositional model.
    \item We demonstrated that CompositionalNets built from a variety of popular DCNN architectures (VGG16, ResNet50 and ResNext) have a significantly increased ability to \textbf{robustly classify out-of-distribution data} in terms of partial occlusion compared to their non-compositional counterparts.
    \item We found that a robust detection of partially occluded objects requires a separation of the representation of the objects' context from that of the object itself. 
    We proposed \textbf{context-aware CompositionalNets} that are learned from image segmentations obtained using bounding box annotations and showed that context-awareness increases robustness to partial occlusion in object detection.
    \item We found that \textbf{CompositionalNets can exploit two complementary approaches to achieve robustness to partial occlusion}: learning features that are invariant to occlusion, or localizing and discarding occluders during classification. We showed that CompositionalNets that combine both approaches achieve the highest robustness to partial occlusion.
\end{itemize}

Furthermore, we showed that CompositionalNets learned from \textit{class-label supervision only} develop a number of intriguing properties in terms of model interpretability:
\begin{itemize}
    \item We showed that \textbf{CompositionalNets can localize occluders accurately} and that the DCNN backbone has a significant influence on this ability. 
    \item Qualitative and quantitative results show that \textbf{CompositionalNets learn meaningful par representations}. This enables them to recognize objects based on the spatial configuration of a few visible parts.
    \item Qualitative and quantitative results show that \textbf{the mixture components in CompositionalNets are viewpoint specific}.
    \item In summary, the \textbf{predictions of CompositionalNets are highly interpretable} in terms of where the model thinks the object is occluded and where the model perceives the individual object parts as well as the objects viewpoint.
\end{itemize}

Our experimental results also hint at important future research directions. We observed that a good occluder localization is add odds with classification performance, because classification benefits from features that are invariant to occlusion, whereas occluder localization requires features to be sensitive to occlusion. 
We believe that it is important to resolve this trade-off wit new types of models that achieve high classification performance while also being able to localize occluders accurately. 
Furthermore, we observed that defining the number of mixture components a-priori fails to account for the fact that different objects can have significantly different pose variability. 
We think unsupervised approaches for determining the number of mixture components are needed to enable the adaptive adjustment of the complexity of the object representation in CompositionalNets.